\documentclass[twoside,11pt]{article}

\usepackage{jmlr2e}
\usepackage{amsmath}
\usepackage{algorithm}
\usepackage{algorithmic}
\usepackage{subfigure}
\usepackage{graphicx}
\usepackage{bbm}
\usepackage{tikz}
\usetikzlibrary{arrows}
\usetikzlibrary{graphs}
\usepackage{url}

\newcommand{\mb}{\mathbf}
\newcommand{\mc}{\mathcal}
\newcommand{\bs}{\boldsymbol}
\newcommand{\ie}{\textit{i.e.}}
\newcommand{\eg}{\textit{e.g.}}

\ShortHeadings{IFM LAB TUTORIAL SERIES \# 9, COPYRIGHT \copyright IFM LAB}{JIAWEI ZHANG, IFM LAB DIRECTOR}
\firstpageno{1}

\begin{document}

\title{Robot Kinematics: Motion, Kinematics and Dynamics}

\author{\name Jiawei Zhang \email jiawei@ifmlab.org \\
	\addr{Founder and Director}\\
       {Information Fusion and Mining Laboratory}\\
       (First Version: November 2022; Revision: November 2022.)}

\maketitle

\begin{abstract}

This is a follow-up tutorial article of \cite{RobotBasics_2022}. For better understanding of the topics covered in this articles, we recommend the readers to first read our previous tutorial article \cite{RobotBasics_2022} on robot basics. Specifically, in this article, we will cover some more advanced topics on robot kinematics, including \textit{robot motion}, \textit{forward kinematics}, \textit{inverse kinematics}, and \textit{robot dynamics}. For the topics, terminologies and notations introduced in the previous article\cite{RobotBasics_2022}, we will use them directly without re-introducing them again in this article. Also similar to the previous article, math and formulas will also be heavily used in this article as well (hope the readers are well prepared for the upcoming math bomb). After reading this article, readers should be able to have a deeper understanding about how robot motion, kinematics and dynamics. As to some more advanced topics about robot control, we will introduce them in the following tutorial articles for readers instead.

\end{abstract}


\begin{keywords}
Robotics; Motion; Forward Kinematics; Inverse Kinematics; Dynamics
\end{keywords}

\tableofcontents

\section{Introduction}\label{sec:introduction}

According to the forecast report from UN (United Nations) \cite{un_ageing}, the world population aging will become one of the most challenging global problem of the 21st century. In the upcoming decades, the working age population growth of the major developing countries (e.g., China, India) will gradually slow down and even start to decrease together with the major developed countries (e.g., UK, France, Germany, Japan and USA). To fulfill the ``tremendous gap'' between the supply and demand of working labor forces, various robots and automated machines has been (and will continue to be) developed and employed for massive production work globally. This is an irreversible trend for the 21st century. To accomplish such an objective, educating and training researchers and practitioner on robotics is imperative and critical at present.

\subsection{This Article}

This is a follow-up tutorial article of \cite{RobotBasics_2022}. In the previous tutorial article, we have introduced the basic knowledge about \textit{robot representation}, \textit{robot rotation}, \textit{position and orientation transformation} and \textit{velocity transformation} already. In this follow-up article, we will introduce several advanced topics about robot kinematics for readers, which include \textit{robot motion}, \textit{forward kinematics}, \textit{inverse kinematics}, and \textit{robot dynamics}. 

\vspace{5pt}
\noindent \textbf{A Reminder}: Similar to \cite{RobotBasics_2022}, this article will also be very math-heavy. When you read the equations in this article, it will be great for you to figure out their physical meanings in mind as well. The math equations will deliver the same information to you just like the textual descriptions. We prefer to use math equations, since they can deliver information to readers in a more precise way.
\vspace{5pt}

For some more advanced topics about robotics, like \textit{robot control}, \textit{trajectory generation}, \textit{motion planning}, \textit{zero moment point}, \textit{biped walking}, \textit{robot manipulation} and \textit{robot simulation}, we plan to introduce them in the follow-up tutorial articles. If the readers are interested in robotics, we also have several textbooks recommended for you to read as well, like \cite{10.5555/3100040, 10.5555/3165183, 10.5555/2490533}.

\subsection{Basic Notations}\label{subsec:basic_notation}

In the sequel of this article, we will use the lower case letters (e.g., $x$) to represent scalars, lower case bold letters (e.g., $\mb{x}$) to denote column vectors, bold-face upper case letters (e.g., $\mb{X}$) to denote matrices. Given a vector $\mb{x}$, its length is denoted as $\left\| \mb{x} \right\|$. Given a matrix $\mb{X}$, we denote $\mb{X}(i,:)$ and $\mb{X}(:,j)$ as its $i_{th}$ row and $j_{th}$ column, respectively. The ($i_{th}$, $j_{th}$) entry of matrix $\mb{X}$ can be denoted as either $\mb{X}(i,j)$ or $\mb{X}_{i,j}$, which will be used interchangeably. We use $\mb{X}^\top$ and $\mb{x}^\top$ to represent the transpose of matrix $\mb{X}$ and vector $\mb{x}$. The cross product of vectors $\mb{x}$ and $\mb{y}$ is represented as $\mb{x} \times \mb{y}$. A coordinate system is denoted as $\Sigma$, and the vector $\mb{x}$ in coordinate system $\Sigma$ can also be specified as $\mb{x}^{[\Sigma]}$. For a scalar $x$, vector $\mb{x}$ and matrix $\mb{X}$, we can also represent their first-order derivatives as $\dot{x}$, $\dot{\mb{x}}$ and $\dot{\mb{X}}$, and second-order derivatives as $\ddot{x}$, $\ddot{\mb{x}}$ and $\ddot{\mb{X}}$.


\section{Robot Motion}\label{sec:robot_motion}

Before we talk about the \textit{forward kinematics} and \textit{inverse kinematics}, we would like to discuss about the \textit{rigid-body robot motion} first in this section. We will revisit the \textit{homogeneous transformation matrix} defined in the previous tutorial article and discuss about some of its properties in this section as well.

\subsection{Homogeneous Transformation Matrix}

In the previous tutorial article, most of the materials we have introduced are about the robot \textit{rotation matrix} and its transformation impact on changing the \textit{position}, \textit{orientation}, \textit{linear velocity} and \textit{angular velocity} of the robot arm end point. We have also briefly introduced the \textit{homogeneous transformation matrix} for readers but we didn't discuss much about its property or usage in robot motion actually, which will be covered in this section.

\subsubsection{Special Orthogonal Group and Special Euclidean Group}

The set of \textit{rotation matrix} we introduce in the previous article can also be clearly represented as the \textit{special orthogonal group} defined as follows.
\begin{definition}
(\textbf{Special Orthogonal Group}): The \textit{special orthogonal group} $SO(3)$, also known as the group of rotation matrices in $\mathbbm{R}^3$, denotes the set of all $3 \times 3$ matrices $\mb{R} \in \mathbbm{R}^{3 \times 3}$ that satisfy (1) $\mb{R}^\top \mb{R} = \mb{I}$ and (2) $\det \mb{R} = 1$.
\end{definition}
In the above definition, the first constraint $\mb{R}^\top \mb{R} = \mb{I}$ denotes the \textit{rotation matrix} $\mb{R}$ is \textit{orthogonal} and each column vector is a unit vector. Meanwhile, the second constraint $\det \mb{R} = 1$ indicates that the rotation is a right-handed frame.

\vspace{5pt}

Meanwhile, based on the rotation matrix, we can also represent the set of \textit{homogeneous transformation matrix} as the \textit{special Euclidean group} as follows.
\begin{definition}
(\textbf{Special Euclidean Group}): The \textit{special Euclidean group} $SE(3)$, also known as the \textit{homogeneous transformation matrix} in $\mathbbm{R}^3$, denotes the set of all $4 \times 4$ matrices $\mb{T} \in \mathbbm{R}^{4 \times 4}$ in the following form
\begin{equation}
\mb{T} = \begin{bmatrix}
\mb{R} & \mb{p}\\
\mb{0} & 1 \\
\end{bmatrix} = \begin{bmatrix}
\mb{R}_{1,1} & \mb{R}_{1,2} & \mb{R}_{1,3} & \mb{p}_1\\
\mb{R}_{2,1} & \mb{R}_{2,2} & \mb{R}_{2,3} & \mb{p}_2\\
\mb{R}_{3,1} & \mb{R}_{3,2} & \mb{R}_{3,3} & \mb{p}_3\\
0 & 0 & 0 & 1\\
\end{bmatrix},
\end{equation}
where the matrix $\mb{R} = \begin{bmatrix}
\mb{R}_{1,1} & \mb{R}_{1,2} & \mb{R}_{1,3}\\
\mb{R}_{2,1} & \mb{R}_{2,2} & \mb{R}_{2,3}\\
\mb{R}_{3,1} & \mb{R}_{3,2} & \mb{R}_{3,3}\\
\end{bmatrix} \in SO(3)$ is a rotation matrix and $\mb{p} = \begin{bmatrix}
\mb{p}_1\\
\mb{p}_3\\
\mb{p}_3\\
\end{bmatrix} \in \mathbbm{R}^3$ is a column vector denoting the robot arm position.
\end{definition}

In the above definitions, the numerical number $3$ in the notations $SO(3)$ and $SE(3)$ indicates that these matrices are defined in the $\mathbbm{R}^3$ space, and it doesn't denote the dimensions of the matrices. Like for a homogeneous transformation matrix $\mb{T} \in SE(3)$, it actually has a dimension of $4 \times 4$.

\subsubsection{Chain Rule on Homogeneous Transformation Matrix}

The \textit{homogeneous transformation matrix} defined across coordinate systems follows the chain rule.

\begin{figure}[t]
    \centering
    \includegraphics[width=0.9\textwidth]{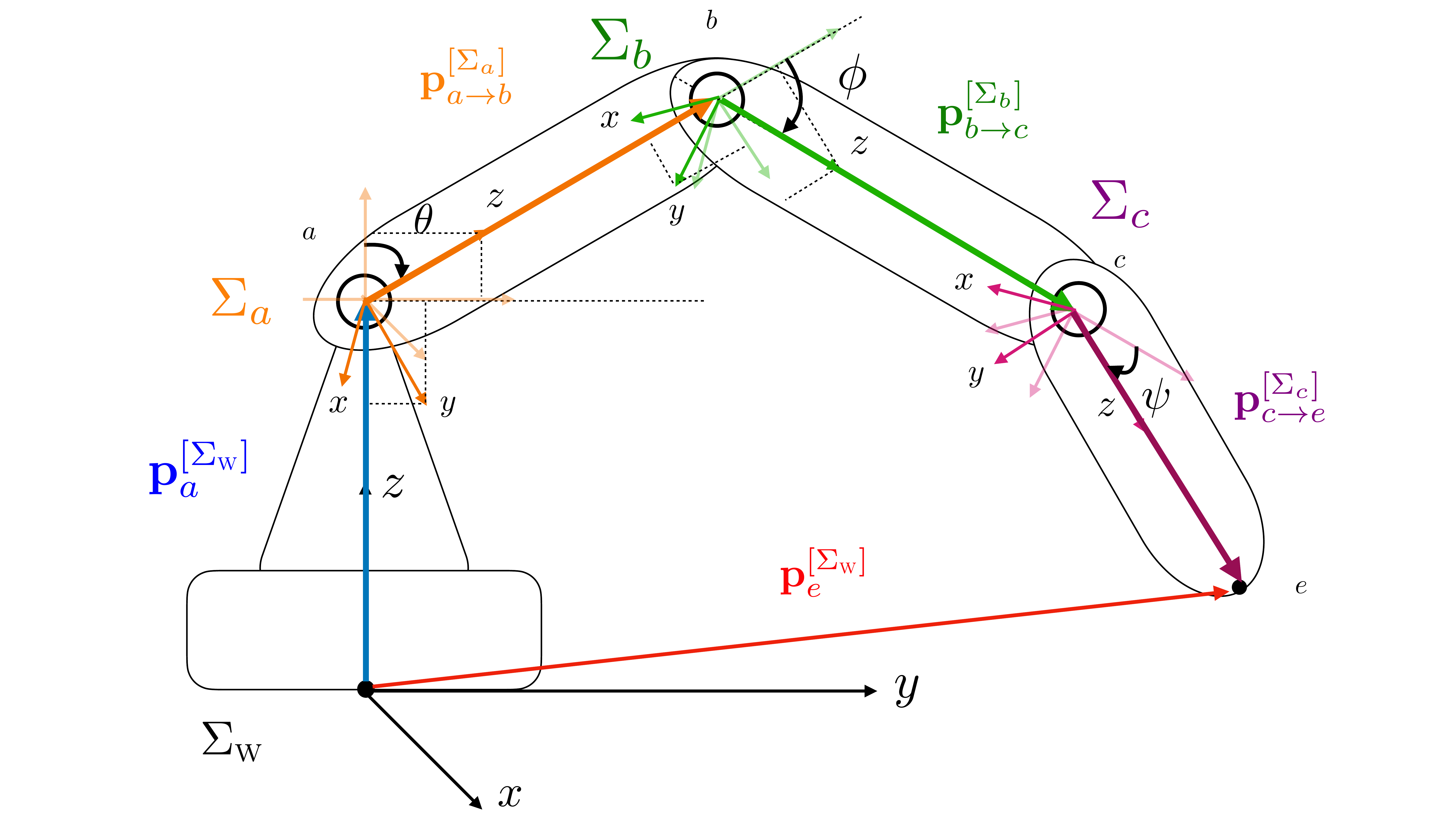}
    \caption{An Example of Rigid-Body Robot Structure (involving 1 base link and 3 movable links connected via joints $a$, $b$ and $c$). The arm rotation angles are denoted as $\theta$, $\phi$ and $\psi$, respectively.}
    \label{fig:robot_arm}
\end{figure}

\begin{example}
Here, we will take a 4-link robot arm shown in Figure~\ref{fig:robot_arm} as an example to illustrate the \textit{forward kinematics} process. Specifically, the robot arm involves 1 base link from the origin $o$ of the world coordinate $\Sigma_{\textsc{w}}$ to joint $a$, and 3 movable links connected via joints $a$, $b$ and $c$. Attached to each joint, we define a local coordinate with origin attached to them, which are denoted as $\Sigma_a$, $\Sigma_b$ and $\Sigma_c$, respectively. Within the local coordinate systems, we can represent the robot arm links as vectors $\mb{p}_{a \to b}^{[\Sigma_a]}$, $\mb{p}_{b \to c}^{[\Sigma_b]}$, and $\mb{p}_{c \to e}^{[\Sigma_c]}$, respectively.

According to the previous tutorial article \cite{RobotBasics_2022}, based on the robot arm rotation angles, we can define the \textit{homogeneous transformation matrix} between the coordinate systems as follows:
\begin{align}
\mb{T}^{[\Sigma_a \to \Sigma_{\textsc{w}}]} &= \begin{bmatrix}
\mb{R}^{[\Sigma_a \to \Sigma_{\textsc{w}}]} & \mb{p}_a^{[\Sigma_{\textsc{w}}]} \\
\mb{0} & 1\\
\end{bmatrix} = \begin{bmatrix}
\exp^{\widehat{\bs{\omega}}_a^{[\Sigma_{\textsc{w}}]}} & \mb{p}_a^{[\Sigma_{\textsc{w}}]} \\
\mb{0} & 1\\
\end{bmatrix},
\end{align}
where the rotation matrix $\mb{R}^{[\Sigma_a \to \Sigma_{\textsc{w}}]} = \exp^{\widehat{\bs{\omega}}_a^{[\Sigma_{\textsc{w}}]}}$ is obtained from the previous tutorial article \cite{RobotBasics_2022}. Notation $\widehat{\bs{\omega}}_a^{[\Sigma_{\textsc{w}}]}$ projects the \textit{angular velocity} vector ${\bs{\omega}}_a^{[\Sigma_{\textsc{w}}]}$ to the matrix representation. As introduced before, the \textit{angular velocity} vector can also be represented as ${\bs{\omega}}_a^{[\Sigma_{\textsc{w}}]} = \mb{e}_a^{[\Sigma_{\textsc{w}}]} \dot{q}_a$, where $\mb{e}_a^{[\Sigma_{\textsc{w}}]}$ denotes the axis unit vector of the joint $a$ and scalar $\dot{q}_a$ denotes the \textit{angular velocity} with the unit rad/s.

Similar to the \textit{homogeneous transition matrix} $\mb{T}^{[\Sigma_a \to \Sigma_{\textsc{w}}]}$ introduced above, we can also define the \textit{homogeneous transition matrices} between other local coordinate systems, including
\begin{equation}
\mb{T}^{[\Sigma_b \to \Sigma_{a}]} = \begin{bmatrix}
\mb{R}^{[\Sigma_b \to \Sigma_{a}]} & \mb{p}_{a \to b}^{[\Sigma_{a}]} \\
\mb{0} & 1\\
\end{bmatrix} \text{, and } \mb{T}^{[\Sigma_c \to \Sigma_b]} = \begin{bmatrix}
\mb{R}^{[\Sigma_c \to \Sigma_b]} & \mb{p}_{b \to c}^{[\Sigma_b]} \\
\mb{0} & 1\\
\end{bmatrix}.\\
\end{equation}

Via the \textit{chain rule}, we can obtain the \textit{homogeneous transition matrix} $\mb{T}^{[\Sigma_c \to \Sigma_{\textsc{w}}]}$, which can be denoted as follows:
\begin{equation}
\mb{T}^{[\Sigma_c \to \Sigma_{\textsc{w}}]} = \mb{T}^{[\Sigma_a \to \Sigma_{\textsc{w}}]} \mb{T}^{[\Sigma_b \to \Sigma_{a}]} \mb{T}^{[\Sigma_c \to \Sigma_b]}.
\end{equation}
\end{example}

\subsubsection{Robot Arm Link Projection across Coordinate Systems}

Via the \textit{homogeneous transformation matrix}, we can project a robot arm from a local coordinate to another local/world coordinate system.

\begin{example}
Still based on the example illustrated in Figure~\ref{fig:robot_arm}, with the end point's current position vector within the local coordinate $\Sigma_c$, we can calculate its end point's position within the world coordinate as
\begin{equation}
\begin{bmatrix}
\mb{p}_e^{[\Sigma_\textsc{w}]}\\
1
\end{bmatrix} = \mb{T}^{[\Sigma_c \to \Sigma_{\textsc{w}}]}
\begin{bmatrix}
\mb{p}_{c \to e}^{[\Sigma_c]}\\
1
\end{bmatrix}.
\end{equation}
\end{example}

In the above example, vector $\mb{p}_{c \to e}^{[\Sigma_c]}$ denotes the third movable robot arm link pointing from the joint $c$ to the end point $e$ in the local coordinate system $\Sigma_c$. Via the \textit{homogeneous transformation matrix} $\mb{T}^{[\Sigma_c \to \Sigma_{\textsc{w}}]}$, we can project it to the world coordinate system and the result vector $\mb{p}_e^{[\Sigma_\textsc{w}]}$ pointing from the origin $o$ of the world coordinate system $\Sigma_{\textsc{w}}$ to the end point $e$ in $\Sigma_{\textsc{w}}$.

\subsection{Robot Twist Velocity}

At the last section of the previous tutorial article \cite{RobotBasics_2022}, we have introduced the transformation of the robot arm end point's \textit{linear velocity} and \textit{angular velocity} during the robot arm rotation. The previous analytic representations of the \textit{linear velocity} and \textit{angular velocity} are derived separately from each other. In this part, we will combine the robot \textit{linear velocity} and \textit{angular velocity} together, which is also named as the \textit{twist} of robots.

\subsubsection{Twist Velocity in World Coordinate System}\label{subsubsec:twist_world}

\begin{definition}
(\textbf{Twist}): Formally, given the \textit{angular velocity} $\bs{\omega}$ and \textit{linear velocity} $\mb{v}$ of a point in robot body, we can combine $\bs{\omega}$ and $\mb{v}$ into one vector, which defines the \textit{twist velocity vector} of the robot body:
\begin{equation}
\bs{\nu} = \begin{bmatrix}
\bs{\omega}\\
\mb{v}\\
\end{bmatrix} \in \mathbbm{R}^6.
\end{equation}
\end{definition}

In the previous article, we introduce the operators $\land: \bs{\omega} \to \dot{\mb{R}}\mb{R}^\top$ (or $\land: \bs{\omega} \to \dot{\mb{R}}\mb{R}^{-1}$) and $\lor: \dot{\mb{R}}\mb{R}^\top \to \bs{\omega}$ (or $\lor: \dot{\mb{R}}\mb{R}^{-1} \to \bs{\omega}$) to create the relationships between the \textit{angular velocity vector} $\omega$ and the \textit{rotation matrix} $\mb{R}$. Here, we can introduce similar operators to the \textit{twist vector} $\bs{\nu}$ and the \textit{homogeneous transformation matrix} $\mb{T}$.

\begin{definition}
(\textbf{Operator Definition}): In this paper, we introduce ab operator $\land$ to project the \textit{twist velocity} vector $\bs{\nu} = \begin{bmatrix}
\bs{\omega}\\
\mb{v}\\
\end{bmatrix}$ to its matrix representation as follows:
\begin{equation}
\widehat{\bs{\nu}} = \begin{bmatrix}
\widehat{\bs{\omega}} & \mb{v}\\
\mb{0} & 0\\
\end{bmatrix}.
\end{equation}
\end{definition}

The above matrix representation of $\widehat{\bs{\nu}}$ is closely correlated with the \textit{homogeneous transformation matrix} $\mb{T}$ actually, and their relationship will be analyzed as follows.

\begin{example}
To clearly explain the physical meanings of the terms to be derived below, we can take the \textit{homogeneous transformation matrix} $\mb{T}^{[\Sigma_a \to \Sigma_{\textsc{w}}]} = \begin{bmatrix}
\mb{R}^{[\Sigma_a \to \Sigma_{\textsc{w}}]} & \mb{p}_a^{[\Sigma_{\textsc{w}}]} \\
\mb{0} & 1\\
\end{bmatrix}$ as an example. 
Let's first calculate the product of matrices $\dot{\mb{T}}^{[\Sigma_a \to \Sigma_{\textsc{w}}]}(\mb{T}^{[\Sigma_a \to \Sigma_{\textsc{w}}]})^{-1}$, which can be represented as follows:
\begin{align}
&\dot{\mb{T}}^{[\Sigma_a \to \Sigma_{\textsc{w}}]} \left(\mb{T}^{[\Sigma_a \to \Sigma_{\textsc{w}}]}\right)^{-1} \\
&= \begin{bmatrix}
\dot{\mb{R}}^{[\Sigma_a \to \Sigma_{\textsc{w}}]} & \dot{\mb{p}}_a^{[\Sigma_{\textsc{w}}]}\\
\mb{0} & 0\\
\end{bmatrix}
\begin{bmatrix}
(\mb{R}^{[\Sigma_a \to \Sigma_{\textsc{w}}]})^{-1} &- (\mb{R}^{[\Sigma_a \to \Sigma_{\textsc{w}}]})^{-1} \mb{p}_a^{[\Sigma_{\textsc{w}}]}\\
\mb{0} & 1\\
\end{bmatrix} \\
&= \begin{bmatrix}
\dot{\mb{R}}^{[\Sigma_a \to \Sigma_{\textsc{w}}]} ({\mb{R}}^{[\Sigma_a \to \Sigma_{\textsc{w}}]})^{-1} & \dot{\mb{p}}_a^{[\Sigma_{\textsc{w}}]}-\dot{\mb{R}}^{[\Sigma_a \to \Sigma_{\textsc{w}}]} ({\mb{R}}^{[\Sigma_a \to \Sigma_{\textsc{w}}]})^{-1} \mb{p}_a^{[\Sigma_{\textsc{w}}]}\\
\mb{0} & 0\\
\end{bmatrix}.
\end{align}
Considering that $\dot{\mb{R}}^{[\Sigma_a \to \Sigma_{\textsc{w}}]} \left({\mb{R}}^{[\Sigma_a \to \Sigma_{\textsc{w}}]} \right)^{-1} = \dot{\mb{R}}^{[\Sigma_a \to \Sigma_{\textsc{w}}]} \left({\mb{R}}^{[\Sigma_a \to \Sigma_{\textsc{w}}]}\right)^\top = \widehat{\bs{\omega}}_a^{[\Sigma_{\textsc{w}}]}$ and $\widehat{\bs{\omega}}_a^{[\Sigma_{\textsc{w}}]} \mb{p}_a^{[\Sigma_{\textsc{w}}]} = {\bs{\omega}}_a^{[\Sigma_{\textsc{w}}]} \times \mb{p}_a^{[\Sigma_{\textsc{w}}]}$, the above representation of matrix $\dot{\mb{T}}^{[\Sigma_a \to \Sigma_{\textsc{w}}]}(\mb{T}^{[\Sigma_a \to \Sigma_{\textsc{w}}]})^{-1}$ can be represented with 
\begin{equation}
\dot{\mb{T}}^{[\Sigma_a \to \Sigma_{\textsc{w}}]} \left(\mb{T}^{[\Sigma_a \to \Sigma_{\textsc{w}}]} \right)^{-1} = 
\begin{bmatrix}
\widehat{\bs{\omega}}_a^{[\Sigma_{\textsc{w}}]} & \dot{\mb{p}}_a^{[\Sigma_{\textsc{w}}]}-{\bs{\omega}}_a^{[\Sigma_{\textsc{w}}]} \times \mb{p}_a^{[\Sigma_{\textsc{w}}]} \\
\mb{0} & 0 \\
\end{bmatrix}.
\end{equation}
The physical meaning of $\widehat{\bs{\omega}}_a^{[\Sigma_{\textsc{w}}]} = \dot{\mb{R}}^{[\Sigma_a \to \Sigma_{\textsc{w}}]} \left({\mb{R}}^{[\Sigma_a \to \Sigma_{\textsc{w}}]} \right)^{-1}$ is very straight-forward and we have introduced the operator in the previous article already. As to the term $\dot{\mb{p}}_a^{[\Sigma_{\textsc{w}}]}-{\bs{\omega}}_a^{[\Sigma_{\textsc{w}}]} \times \mb{p}_a^{[\Sigma_{\textsc{w}}]}$, it actually denotes the \textit{linear velocity} of joint $a$ within the world coordinate system $\Sigma_{\textsc{w}}$ with the velocity caused by the rotation being excluded, {\ie}, ${\bs{\omega}}_a^{[\Sigma_{\textsc{w}}]} \times \mb{p}_a^{[\Sigma_{\textsc{w}}]}$. Such a \textit{linear velocity} can be caused by the movement of the robot arm base, which introduce extra velocity for all points in the robot arm.

For simplicity, we can denote $\mb{v}_a^{[\Sigma_{\textsc{w}}]} = \dot{\mb{p}}_a^{[\Sigma_{\textsc{w}}]}-{\bs{\omega}}_a^{[\Sigma_{\textsc{w}}]} \times \mb{p}_a^{[\Sigma_{\textsc{w}}]}$, so we can simlify rewrite the \textit{homogeneous transformation matrix} product as follows:
\begin{equation}
\dot{\mb{T}}^{[\Sigma_a \to \Sigma_{\textsc{w}}]} \left(\mb{T}^{[\Sigma_a \to \Sigma_{\textsc{w}}]} \right)^{-1} = \begin{bmatrix}
\widehat{\bs{\omega}}_a^{[\Sigma_{\textsc{w}}]} & \mb{v}_a^{[\Sigma_{\textsc{w}}]}\\
\mb{0} & 0\\
\end{bmatrix}.
\end{equation}
\end{example}

Based on the above analysis, we observe that with in the world coordinate system $\Sigma_{\textsc{w}}$, we can apply the $\land$ operator to the \textit{twist velocity} $\bs{\nu}_a^{[\Sigma_{\textsc{w}}]}$, the result matrix is identical as the corresponding \textit{homogeneous matrix} product $\dot{\mb{T}}^{[\Sigma_a \to \Sigma_{\textsc{w}}]} \left(\mb{T}^{[\Sigma_a \to \Sigma_{\textsc{w}}]} \right)^{-1}$ actually:
\begin{equation}\label{equ:twist_world_coordinate_system}
({\bs{\nu}}_a^{[\Sigma_{\textsc{w}}]})^{\land} = \widehat{\bs{\nu}}_a^{[\Sigma_{\textsc{w}}]} = \begin{bmatrix}
\widehat{\bs{\omega}}_a^{[\Sigma_{\textsc{w}}]} & \mb{v}_a^{[\Sigma_{\textsc{w}}]}\\
\mb{0} & 0\\
\end{bmatrix} = \dot{\mb{T}}^{[\Sigma_a \to \Sigma_{\textsc{w}}]} \left(\mb{T}^{[\Sigma_a \to \Sigma_{\textsc{w}}]} \right)^{-1}.
\end{equation}

\subsubsection{Twist in Local Coordinate System}\label{subsubsec:twist_local}

In the above analysis, we multiply $\left(\mb{T}^{[\Sigma_a \to \Sigma_{\textsc{w}}]}\right)^{-1}$ on the right side of matrix $\dot{\mb{T}}^{[\Sigma_a \to \Sigma_{\textsc{w}}]}$. Some readers may also wonder if we multiple it to the left side will affect the result or not? In this part, we will answer this question.

Similar to the above analysis, we can first calculate the left multiplication of $\left(\mb{T}^{[\Sigma_a \to \Sigma_{\textsc{w}}]}\right)^{-1}$ to $\dot{\mb{T}}^{[\Sigma_a \to \Sigma_{\textsc{w}}]}$ as follows:
\begin{align}
&\left(\mb{T}^{[\Sigma_a \to \Sigma_{\textsc{w}}]}\right)^{-1} \dot{\mb{T}}^{[\Sigma_a \to \Sigma_{\textsc{w}}]} \\
&= \begin{bmatrix}
(\mb{R}^{[\Sigma_a \to \Sigma_{\textsc{w}}]})^{-1} &- (\mb{R}^{[\Sigma_a \to \Sigma_{\textsc{w}}]})^{-1} \mb{p}_a^{[\Sigma_{\textsc{w}}]}\\
\mb{0} & 1\\
\end{bmatrix} \begin{bmatrix}
\dot{\mb{R}}^{[\Sigma_a \to \Sigma_{\textsc{w}}]} & \dot{\mb{p}}_a^{[\Sigma_{\textsc{w}}]}\\
\mb{0} & 0\\
\end{bmatrix} \\
&= \begin{bmatrix}
(\mb{R}^{[\Sigma_a \to \Sigma_{\textsc{w}}]})^{-1} \dot{\mb{R}}^{[\Sigma_a \to \Sigma_{\textsc{w}}]} & (\mb{R}^{[\Sigma_a \to \Sigma_{\textsc{w}}]})^{-1} \dot{\mb{p}}_a^{[\Sigma_{\textsc{w}}]} \\
\mb{0} & 0\\
\end{bmatrix}\\
&= \begin{bmatrix}
(\mb{R}^{[\Sigma_a \to \Sigma_{\textsc{w}}]})^\top \dot{\mb{R}}^{[\Sigma_a \to \Sigma_{\textsc{w}}]} & (\mb{R}^{[\Sigma_a \to \Sigma_{\textsc{w}}]})^\top \dot{\mb{p}}_a^{[\Sigma_{\textsc{w}}]} \\
\mb{0} & 0\\
\end{bmatrix}.
\end{align}

In the above representation, the physical meaning of term $(\mb{R}^{[\Sigma_a \to \Sigma_{\textsc{w}}]})^\top \dot{\mb{p}}_a^{[\Sigma_{\textsc{w}}]} = \mb{v}_a^{[\Sigma_a]}$ should be trivial for the readers to figure out. As to $(\mb{R}^{[\Sigma_a \to \Sigma_{\textsc{w}}]})^\top \dot{\mb{R}}^{[\Sigma_a \to \Sigma_{\textsc{w}}]}$, as illustrated as follows, it actually denotes $\widehat{\bs{\omega}}_a^{[\Sigma_a]}$.

\begin{align}
\widehat{\bs{\omega}}_a^{[\Sigma_a]} &= \left( (\mb{R}^{[\Sigma_a \to \Sigma_{\textsc{w}}]})^{-1} \bs{\omega}_a^{\Sigma_{\textsc{w}}} \right)^{\land} \\
&= \left( (\mb{R}^{[\Sigma_a \to \Sigma_{\textsc{w}}]})^\top \bs{\omega}_a^{\Sigma_{\textsc{w}}} \right)^{\land} \\
&= (\mb{R}^{[\Sigma_a \to \Sigma_{\textsc{w}}]})^\top \widehat{\bs{\omega}}_a^{\Sigma_{\textsc{w}}} (\mb{R}^{[\Sigma_a \to \Sigma_{\textsc{w}}]})\\
&= (\mb{R}^{[\Sigma_a \to \Sigma_{\textsc{w}}]})^\top \left( \dot{\mb{R}}^{[\Sigma_a \to \Sigma_{\textsc{w}}]} (\mb{R}^{[\Sigma_a \to \Sigma_{\textsc{w}}]})^\top \right) \mb{R}^{[\Sigma_a \to \Sigma_{\textsc{w}}]} \\
&= (\mb{R}^{[\Sigma_a \to \Sigma_{\textsc{w}}]})^\top  \dot{\mb{R}}^{[\Sigma_a \to \Sigma_{\textsc{w}}]}  \left( (\mb{R}^{[\Sigma_a \to \Sigma_{\textsc{w}}]})^\top \mb{R}^{[\Sigma_a \to \Sigma_{\textsc{w}}]} \right) \\
&= (\mb{R}^{[\Sigma_a \to \Sigma_{\textsc{w}}]})^\top  \dot{\mb{R}}^{[\Sigma_a \to \Sigma_{\textsc{w}}]}.
\end{align}

Therefore, we observe that the left multiplication of the homogeneous transformation matrix terms will be denoted as
\begin{equation}\label{equ:twist_local_coordinate_system}
\left(\mb{T}^{[\Sigma_a \to \Sigma_{\textsc{w}}]}\right)^{-1} \dot{\mb{T}}^{[\Sigma_a \to \Sigma_{\textsc{w}}]} = 
\begin{bmatrix}
\widehat{\bs{\omega}}_a^{[\Sigma_a]} & \mb{v}_a^{[\Sigma_a]}\\
\mb{0} & 0\\
\end{bmatrix} = \begin{bmatrix}
{\bs{\omega}}_a^{[\Sigma_a]}\\
\mb{v}_a^{[\Sigma_a]}\\
\end{bmatrix}^{\land} = \widehat{\bs{\nu}}_a^{[\Sigma_a]}.
\end{equation}
In other words, the left matrix multiplication $\left(\mb{T}^{[\Sigma_a \to \Sigma_{\textsc{w}}]}\right)^{-1} \dot{\mb{T}}^{[\Sigma_a \to \Sigma_{\textsc{w}}]}$ will be equivalent to the result we can get by applying the $\land$ operator to the \textit{twist velocity} ${\bs{\nu}}_a^{[\Sigma_a]}$ at joint $a$ within the local coordinate system $\Sigma_a$.

\subsection{Twist Velocity Transformation across Coordinate Systems}

In the previous subsection, we introduce the \textit{twist velocity} vector for readers, and also illustrate the \textit{twist velocity} vector representations within different coordinate systems subject to the new $\land$ operator. In this part, we will further discuss about the relationships between the \textit{twist velocity vectors} $\widehat{\bs{\nu}}_a^{[\Sigma_a]}$ and $\widehat{\bs{\nu}}_a^{[\Sigma_\textsc{w}]}$, as well as between ${\bs{\nu}}_a^{[\Sigma_a]}$ and ${\bs{\nu}}_a^{[\Sigma_\textsc{w}]}$, across coordinate systems.

\subsubsection{Relationship Between $\widehat{\bs{\nu}}_a^{[\Sigma_a]}$ And $\widehat{\bs{\nu}}_a^{[\Sigma_\textsc{w}]}$}\label{subsubsec:twist_hat_relationship}

Let's first talk about the relationship between $\widehat{\bs{\nu}}_a^{[\Sigma_a]}$ and $\widehat{\bs{\nu}}_a^{[\Sigma_\textsc{w}]}$ in different coordinate systems. Based on the detailed derivations in the previous two subsections, we know that
\begin{equation}\label{equ:twist_relationship}
\widehat{\bs{\nu}}_a^{[\Sigma_a]} = \left(\mb{T}^{[\Sigma_a \to \Sigma_{\textsc{w}}]}\right)^{-1} \dot{\mb{T}}^{[\Sigma_a \to \Sigma_{\textsc{w}}]}.
\end{equation}

Meanwhile, since 
\begin{equation}
\widehat{\bs{\nu}}_a^{[\Sigma_{\textsc{w}}]} = \dot{\mb{T}}^{[\Sigma_a \to \Sigma_{\textsc{w}}]} \left(\mb{T}^{[\Sigma_a \to \Sigma_{\textsc{w}}]} \right)^{-1},
\end{equation} 
by multiplying $\mb{T}^{[\Sigma_a \to \Sigma_{\textsc{w}}]}$ to both sides of the above equation, we can get
\begin{equation}
\widehat{\bs{\nu}}_a^{[\Sigma_{\textsc{w}}]} \mb{T}^{[\Sigma_a \to \Sigma_{\textsc{w}}]} = \dot{\mb{T}}^{[\Sigma_a \to \Sigma_{\textsc{w}}]}.
\end{equation} 

By replacing the representation of $\dot{\mb{T}}^{[\Sigma_a \to \Sigma_{\textsc{w}}]}$ in the above equation into Equation~\ref{equ:twist_relationship}, we can illustrate the relationship between $\widehat{\bs{\nu}}_a^{[\Sigma_{\textsc{w}}]}$ and $\widehat{\bs{\nu}}_a^{[\Sigma_a]}$ as follows:
\begin{align}
\widehat{\bs{\nu}}_a^{[\Sigma_a]} 
&= \left(\mb{T}^{[\Sigma_a \to \Sigma_{\textsc{w}}]}\right)^{-1} \dot{\mb{T}}^{[\Sigma_a \to \Sigma_{\textsc{w}}]}\\
&= \left(\mb{T}^{[\Sigma_a \to \Sigma_{\textsc{w}}]}\right)^{-1} \widehat{\bs{\nu}}_a^{[\Sigma_{\textsc{w}}]} \mb{T}^{[\Sigma_a \to \Sigma_{\textsc{w}}]}.
\end{align}

From it, we can also obtain that
\begin{equation}\label{equ:twist_relationship_2}
\widehat{\bs{\nu}}_a^{[\Sigma_\textsc{w}]} = \mb{T}^{[\Sigma_a \to \Sigma_{\textsc{w}}]} \widehat{\bs{\nu}}_a^{[\Sigma_a]} (\mb{T}^{[\Sigma_a \to \Sigma_{\textsc{w}}]})^{-1}.
\end{equation}

In other word, via the \textit{homogeneous transformation matrix}, we can project the \textit{twist velocity} vectors subject to the $\land$ operator to each other.

\subsubsection{Relationship Between ${\bs{\nu}}_a^{[\Sigma_a]}$ And ${\bs{\nu}}_a^{[\Sigma_\textsc{w}]}$}\label{subsubsec:twist_relationship}\label{subsubsec:adjoint_representation_homogeneous_transformation_matrix}

Based on the above analysis, we can further derive the relationship between the \textit{twist velocity} vectors ${\bs{\nu}}_a^{[\Sigma_a]}$ and ${\bs{\nu}}_a^{[\Sigma_\textsc{w}]}$. By replace the terms $\widehat{\bs{\nu}}_a^{[\Sigma_\textsc{w}]}$, $\mb{T}^{[\Sigma_a \to \Sigma_{\textsc{w}}]}$ and $\widehat{\bs{\nu}}_a^{[\Sigma_a]}$ with their concrete matrix representation into Equation~\ref{equ:twist_relationship_2}, we can get
\begin{align}\label{equ:twist_relationship_3}
&\begin{bmatrix}
\widehat{\bs{\omega}}_a^{[\Sigma_{\textsc{w}}]} & \mb{v}_a^{[\Sigma_{\textsc{w}}]}\\
\mb{0} & 0\\
\end{bmatrix} \\
&= 
\begin{bmatrix}
\mb{R}^{[\Sigma_a \to \Sigma_{\textsc{w}}]} & \mb{p}_a^{[\Sigma_{\textsc{w}}]} \\
\mb{0} & 1\\
\end{bmatrix} 
\begin{bmatrix}
\widehat{\bs{\omega}}_a^{[\Sigma_a]} & \mb{v}_a^{[\Sigma_a]}\\
\mb{0} & 0\\
\end{bmatrix} 
\begin{bmatrix}
(\mb{R}^{[\Sigma_a \to \Sigma_{\textsc{w}}]})^{-1} &- (\mb{R}^{[\Sigma_a \to \Sigma_{\textsc{w}}]})^{-1} \mb{p}_a^{[\Sigma_{\textsc{w}}]}\\
\mb{0} & 1\\
\end{bmatrix} \\
&= 
\begin{bmatrix}
\mb{R}^{[\Sigma_a \to \Sigma_{\textsc{w}}]} \widehat{\bs{\omega}}_a^{[\Sigma_a]} & \mb{R}^{[\Sigma_a \to \Sigma_{\textsc{w}}]} \mb{v}_a^{[\Sigma_a]} \\
\mb{0} & 0\\
\end{bmatrix}
\begin{bmatrix}
(\mb{R}^{[\Sigma_a \to \Sigma_{\textsc{w}}]})^{-1} &- (\mb{R}^{[\Sigma_a \to \Sigma_{\textsc{w}}]})^{-1} \mb{p}_a^{[\Sigma_{\textsc{w}}]}\\
\mb{0} & 1\\
\end{bmatrix} \\
&= 
\begin{bmatrix}
\mb{R}^{[\Sigma_a \to \Sigma_{\textsc{w}}]} \widehat{\bs{\omega}}_a^{[\Sigma_a]}(\mb{R}^{[\Sigma_a \to \Sigma_{\textsc{w}}]})^{-1} & -\mb{R}^{[\Sigma_a \to \Sigma_{\textsc{w}}]} \widehat{\bs{\omega}}_a^{[\Sigma_a]} (\mb{R}^{[\Sigma_a \to \Sigma_{\textsc{w}}]})^{-1} \mb{p}_a^{[\Sigma_{\textsc{w}}]} + \mb{R}^{[\Sigma_a \to \Sigma_{\textsc{w}}]} \mb{v}_a^{[\Sigma_a]} \\
\mb{0} & 0\\
\end{bmatrix}\\
&= 
\begin{bmatrix}
\mb{R}^{[\Sigma_a \to \Sigma_{\textsc{w}}]} \widehat{\bs{\omega}}_a^{[\Sigma_a]}(\mb{R}^{[\Sigma_a \to \Sigma_{\textsc{w}}]})^{\top} & -\mb{R}^{[\Sigma_a \to \Sigma_{\textsc{w}}]} \widehat{\bs{\omega}}_a^{[\Sigma_a]} (\mb{R}^{[\Sigma_a \to \Sigma_{\textsc{w}}]})^{\top} \mb{p}_a^{[\Sigma_{\textsc{w}}]} + \mb{R}^{[\Sigma_a \to \Sigma_{\textsc{w}}]} \mb{v}_a^{[\Sigma_a]} \\
\mb{0} & 0\\
\end{bmatrix}\\
&= 
\begin{bmatrix}
\left(\mb{R}^{[\Sigma_a \to \Sigma_{\textsc{w}}]} {\bs{\omega}}_a^{[\Sigma_a]}\right)^{\land}
& - \left(\mb{R}^{[\Sigma_a \to \Sigma_{\textsc{w}}]} {\bs{\omega}}_a^{[\Sigma_a]}\right)^{\land} \mb{p}_a^{[\Sigma_{\textsc{w}}]} + \mb{R}^{[\Sigma_a \to \Sigma_{\textsc{w}}]} \mb{v}_a^{[\Sigma_a]} \\
\mb{0} & 0\\
\end{bmatrix}\\
\end{align}

Based on some basic properties about cross-product and the $\land$ operator on the rotation matrix, the term $- \left(\mb{R}^{[\Sigma_a \to \Sigma_{\textsc{w}}]} {\bs{\omega}}_a^{[\Sigma_a]}\right)^{\land} \mb{p}_a^{[\Sigma_{\textsc{w}}]}$ can be rewritten as
\begin{align}
- \left(\mb{R}^{[\Sigma_a \to \Sigma_{\textsc{w}}]} {\bs{\omega}}_a^{[\Sigma_a]}\right)^{\land} \mb{p}_a^{[\Sigma_{\textsc{w}}]} &=
- \left(\mb{R}^{[\Sigma_a \to \Sigma_{\textsc{w}}]} {\bs{\omega}}_a^{[\Sigma_a]}\right) \times \mb{p}_a^{[\Sigma_{\textsc{w}}]} \\
&= \mb{p}_a^{[\Sigma_{\textsc{w}}]} \times \left(\mb{R}^{[\Sigma_a \to \Sigma_{\textsc{w}}]} {\bs{\omega}}_a^{[\Sigma_a]}\right) \\
&= (\mb{p}_a^{[\Sigma_{\textsc{w}}]})^\land \left(\mb{R}^{[\Sigma_a \to \Sigma_{\textsc{w}}]} {\bs{\omega}}_a^{[\Sigma_a]}\right) \\
&= \widehat{\mb{p}}_a^{[\Sigma_{\textsc{w}}]} \mb{R}^{[\Sigma_a \to \Sigma_{\textsc{w}}]} {\bs{\omega}}_a^{[\Sigma_a]}.
\end{align}

Therefore, from the above Equation~\ref{equ:twist_relationship_3}, we can get that
\begin{equation}\label{equ:linear_equations}
\begin{cases}
{\bs{\omega}}_a^{[\Sigma_{\textsc{w}}]} &= \mb{R}^{[\Sigma_a \to \Sigma_{\textsc{w}}]} {\bs{\omega}}_a^{[\Sigma_a]} \\
\mb{v}_a^{[\Sigma_{\textsc{w}}]} &= \widehat{\mb{p}}_a^{[\Sigma_{\textsc{w}}]} \mb{R}^{[\Sigma_a \to \Sigma_{\textsc{w}}]} {\bs{\omega}}_a^{[\Sigma_a]} + \mb{R}^{[\Sigma_a \to \Sigma_{\textsc{w}}]} \mb{v}_a^{[\Sigma_a]}.
\end{cases}
\end{equation}


We can also represent the above Equation~\ref{equ:linear_equations} with linear algebra representation, which will illustrate the relationship between ${\bs{\nu}}_a^{[\Sigma_a]}$ and ${\bs{\nu}}_a^{[\Sigma_\textsc{w}]}$ as follows:
\begin{equation}\label{equ:adjoint_representation_homogeneous_transformation}
\begin{bmatrix}
{\bs{\omega}}_a^{[\Sigma_{\textsc{w}}]}\\
\mb{v}_a^{[\Sigma_{\textsc{w}}]}
\end{bmatrix}
= \underbrace{\begin{bmatrix}
\mb{R}^{[\Sigma_a \to \Sigma_{\textsc{w}}]} & \mb{0}\\
\widehat{\mb{p}}_a^{[\Sigma_{\textsc{w}}]} \mb{R}^{[\Sigma_a \to \Sigma_{\textsc{w}}]} & \mb{R}^{[\Sigma_a \to \Sigma_{\textsc{w}}]}\\
\end{bmatrix}}_{\text{adjoint representation } Ad({{\mb{T}}^{[\Sigma_a \to \Sigma_{\textsc{w}}]}})}
\begin{bmatrix}
{\bs{\omega}}_a^{[\Sigma_a]}\\
\mb{v}_a^{[\Sigma_a]}
\end{bmatrix}
\end{equation}

\begin{definition}
(\textbf{Adjoint Representation}): Given the  \textit{homogeneous transformation matrix} 
\begin{equation}
\mb{T}^{[\Sigma_a \to \Sigma_{\textsc{w}}]} = \begin{bmatrix}
\mb{R}^{[\Sigma_a \to \Sigma_{\textsc{w}}]} & \mb{p}_a^{[\Sigma_{\textsc{w}}]}\\
\mb{0} & 1\\
\end{bmatrix}
\end{equation}
We can represent its \textit{adjoint representation} as follows:
\begin{equation}
Ad({{\mb{T}}^{[\Sigma_a \to \Sigma_{\textsc{w}}]}}) = \begin{bmatrix}
\mb{R}^{[\Sigma_a \to \Sigma_{\textsc{w}}]} & \mb{0}\\
\widehat{\mb{p}}_a^{[\Sigma_{\textsc{w}}]} \mb{R}^{[\Sigma_a \to \Sigma_{\textsc{w}}]} & \mb{R}^{[\Sigma_a \to \Sigma_{\textsc{w}}]}\\
\end{bmatrix}
\end{equation}
\end{definition}

The \textit{adjoint representation} of the homogeneous transformation matrix will project the \textit{twist velocity} vectors across different coordinate systems. For instance, from the above analysis, we can get that
\begin{equation}\label{equ:twist_transformation_1}
\bs{\nu}_a^{[\Sigma_{\textsc{w}}]} = Ad({{\mb{T}}^{[\Sigma_a \to \Sigma_{\textsc{w}}]}}) \bs{\nu}_a^{[\Sigma_a]}.
\end{equation}

With a similar process, we can also obtain that
\begin{equation}\label{equ:twist_transformation_2}
\bs{\nu}_a^{[\Sigma_a]} = Ad({{\mb{T}}^{[\Sigma_{\textsc{w}} \to \Sigma_a]}}) \bs{\nu}_a^{[\Sigma_{\textsc{w}}]}.
\end{equation}

\subsection{Screw}

We have been discussing about the robot \textit{body motion} and \textit{twist velocity} vector above. Some questions may naturally arise in readers' mind: ``why do we need to study both `robot rotation' and `robot motion'?''

To answer the question, we need to distinguish the differences between ``robot rotation'' and ``robot motion''. As indicated by the name, ``robot rotation'' as introduced in \cite{RobotBasics_2022} denotes the rotational movement of the robot body (or part of the body). Meanwhile, ``robot motion''  involves both the \textit{rotational} movement, as well as the \textit{translational} movement, so it will be much more complicated to model than the the \textit{robot rotation}.

\subsubsection{A Toy Example}

Some readers probably may propose another question ``Can we project the `robot motion' as a type of `robot rotation' movement, or combine them and model them as one movement?'' In stead of directly answering the question, we can use a toy example below to illustrate our answer.

\begin{figure}[t]
    \centering
    \includegraphics[width=0.9\textwidth]{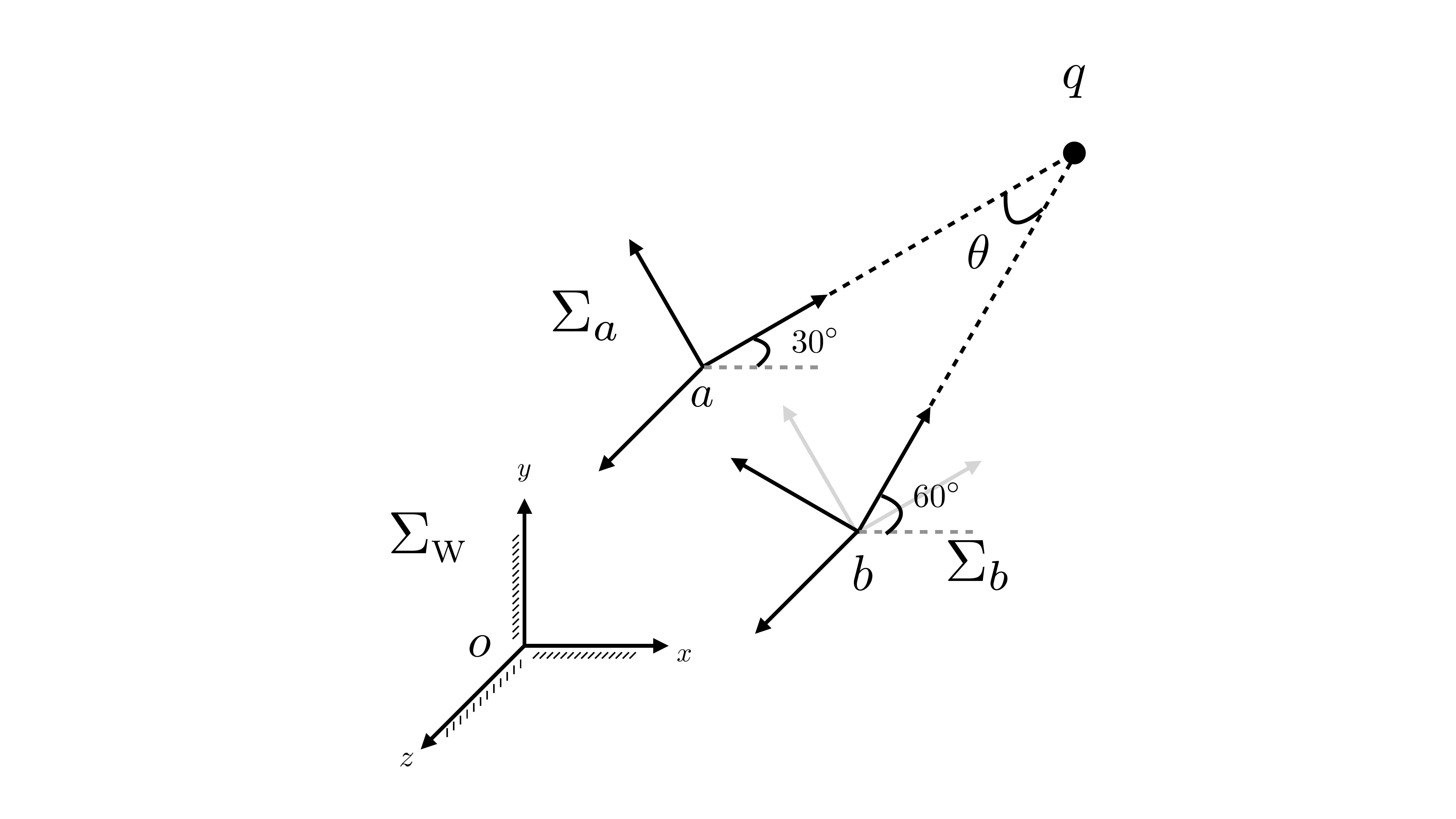}
    \caption{A Example of Robot Motion. (The motion of coordinate $\Sigma_a$ to $\Sigma_b$ via both rotation and translation can be modeled as a rotation around point $q$ with an angle of $\theta$ degrees.)}
    \label{fig:robot_motion_example}
\end{figure}

\begin{example}
As shown in Figure~\ref{fig:robot_motion_example}, in the 3D space, we provide three coordinate systems: (1) the world coordinate system $\Sigma_{\textsc{w}}$ with origin $o$, (2) the local coordinate system $\Sigma_a$ before motion with origin at point $\mb{p}_a = (1, 2, 0)^\top$, and (3) the local coordinate system $\Sigma_b$ after motion with origin at point $\mb{p}_a = (2, 1, 0)^\top$. As to the orientation, compared with the world coordinate system $\Sigma_{\textsc{w}}$, coordinate system $\Sigma_a$ rotates in the counter clockwise direction around axis $z$ with an angle of $30^{\circ}$, and $\Sigma_b$ rotates in he counter clockwise direction with an angle of $60^{\circ}$.

Based on the orientation and positions of these two local coordinate systems, we can define the \textit{homogeneous transformation matrices} from them to the world coordinate system as
\begin{equation}
\mb{T}^{[\Sigma_a \to \Sigma_{\textsc{w}}]} = \begin{bmatrix}
\cos 30^{\circ} & -\sin 30^{\circ} & 0 & 1 \\
\sin 30^{\circ} & \cos 30^{\circ} & 0 & 2 \\
0 & 0 & 1 & 0 \\
0 & 0 & 0 & 1
\end{bmatrix}, 
\mb{T}^{[\Sigma_b \to \Sigma_{\textsc{w}}]} = \begin{bmatrix}
\cos 60^{\circ} & -\sin 60^{\circ} & 0 & 2 \\
\sin 60^{\circ} & \cos 60^{\circ} & 0 & 1 \\
0 & 0 & 1 & 0 \\
0 & 0 & 0 & 1
\end{bmatrix}.
\end{equation}

Based on them, we can calculate the transformation matrix from $\Sigma_a$ to $\Sigma_b$ as
\begin{align}
\mb{T}^{[\Sigma_b \to \Sigma_a]} &= (\mb{T}^{[\Sigma_a \to \Sigma_{\textsc{w}}]})^{-1} \mb{T}^{[\Sigma_b \to \Sigma_{\textsc{w}}]} \\
&= \begin{bmatrix}
0.866 & -0.5 & 0 & 2.134 \\
0.5 & 0.866 & 0 & -1.232 \\
0 & 0 & 1 & 0 \\
0 & 0 & 0 & 1\\
\end{bmatrix}\\
&= \begin{bmatrix}
\cos 30^{\circ} & -\sin 30^{\circ} & 0 & 0.366 \\
\sin 30^{\circ} & \cos 30^{\circ} & 0 & -1.366 \\
0 & 0 & 1 & 0 \\
0 & 0 & 0 & 1\\
\end{bmatrix}.
\end{align}

So, according to matrix $\mb{T}^{[\Sigma_b \to \Sigma_a]}$, looking from $\Sigma_a$, the motion of the local coordinate system from $\Sigma_a$ to $\Sigma_b$ involves both (1) a rotation in the counter clockwise direction with an angle of $30^{\circ}$ around the $z$ axis, and (2) a translation with a position vector $(0.366, -1.366, 0)^\top$.

Meanwhile, looking from the world coordinate system $\Sigma_{\textsc{w}}$, it can also be modeled as a pure rotational movement around the axis vector at point $q$, {\ie}, $\mb{p}_q = (3.37, 3.37, 0)^\top$, in the counter clockwise direction with an angle of $\theta$ degrees actually.
\end{example}

In the above example, the movement involves both a rotational movement and a translational movement, which can be modeled as a pure rotational movement around certain axis in the space. How about a pure translational movement without any rotations? Actually, the translational movement can still be modeled as a rotational movement with the rotation radius approaches $+\infty$. 

\subsubsection{Screw Axis}

To formally model the rigid robot motion as a rotational movement, in this part, we will introduce the concept of \textit{screw axis} and will also introduce two operators to project \textit{screw axis} with the \textit{transformation matrix}.

If the readers still remember, when introducing the \textit{angular velocity} vector $\bs{\omega}$, we mention that it can be represented as $\bs{\omega} = \dot{q} \mb{e}_{\bs{\omega}}$, where $\mb{e}_{\bs{\omega}}$ is a unit vector indicating the direction of the \textit{angular velocity} vector and scalar $\dot{q}$ denotes the rate of ration around the axis with the unit rad/s. It helps interpret the physical meaning of the \textit{angular velocity} vector. Meanwhile, when it comes to the \textit{twist velocity} vector $\bs{\nu}$, the readers probably may also wonder if it can also be interpreted in a similar way or not. In this part, we aim to answer this question and represent the \textit{twist velocity} vector with the \textit{screw} motion.

\begin{example}
\begin{figure}[t]
    \centering
    \includegraphics[width=0.9\textwidth]{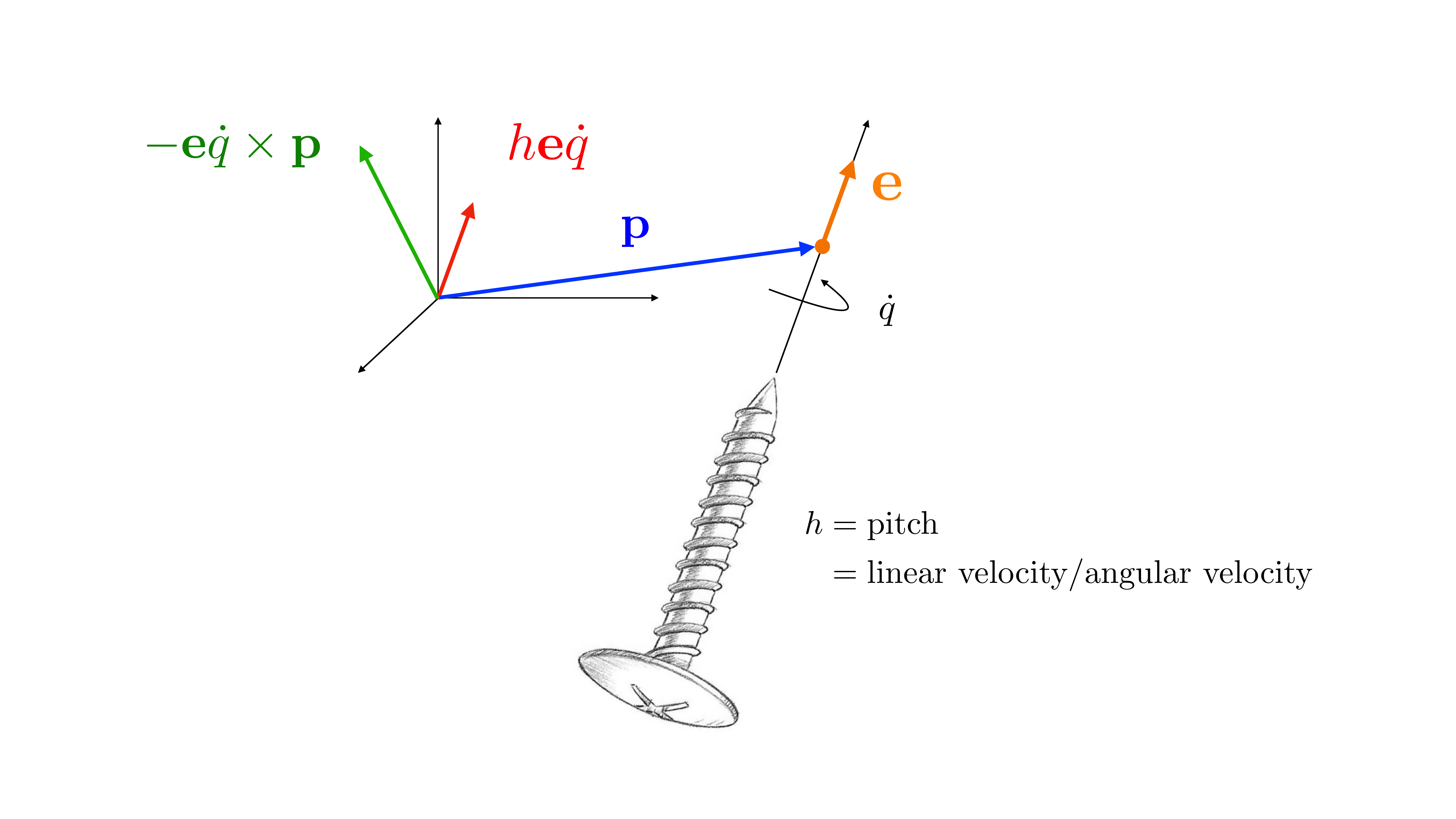}
    \caption{An Example of Screw Rotational Movement. (Unit vector $\mb{e}$: the direction of the screw axis; Scalar $\dot{q}$: rotation rate of screw; Scalar $h$: pitch of screw and it equals to ``linear velocity scalar along screw axis/angular velocity scalar''; Vector $\mb{p}$: it denotes a point on the screw axis.)}
    \label{fig:robot_motion_example}
\end{figure}
The \textit{twist velocity} vector contains both \textit{angular velocity} and \textit{linear velocity}, which may remind us about the \textit{screw} rotational movement. In Figure~\ref{fig:robot_motion_example}, we show an example of a \textit{screw} rotating around the axis with direction indicated by a unit vector $\mb{e}_s$. The \textit{screw} rotating rate can be denoted by a scalar $\dot{q}$ with unit rad/s, and the \textit{screw pitch} is represented by another scalar $h$ in this example, which denotes the forward movement distance as the screw rotates an angle of $360^{\circ}$ degrees. One of the point on the axis is denoted by position vector $\mb{p}$. The collection of $\mb{s} = (\mb{p}, \mb{e}_s, h)$ together defines the \textit{screw axis} representation, and $\dot{q}$ denotes the \textit{angular velocity} scalar around the \textit{screw axis}.
\end{example}

\noindent \textbf{Screw Representation}: Based on the above screw representation, given a \textit{twist velocity} vector $\bs{\nu}$, by choosing proper elements of the \textit{screw axis}, we can also represent it with the \textit{screw axis} together with the corresponding rotational \textit{angular velocity} scalar $\dot{q}$ as follows:
\begin{equation}\label{equ:screw_representation}
\bs{\nu} = \begin{bmatrix}
\bs{\omega}\\
\mb{v}\\
\end{bmatrix}=\begin{bmatrix}
\dot{q} \mb{e}_s\\
-\dot{q} \mb{e}_s \times \mb{p} + h \mb{e}_s \dot{q}\\
\end{bmatrix}.
\end{equation}
The physical meaning of $\dot{q} \mb{e}_s$ denotes the \textit{angular velocity} of the \textit{screw} in rotation, As to the \textit{linear velocity} representation $-\dot{q} \mb{e}_s \times \mb{p} + h \mb{e}_s \dot{q}$, it has two parts: (1) term $h \mb{e}_s \dot{q}$ denotes the translational linear velocity that the screw moves forward as it rotates, and (2) $-\dot{q} \mb{e}_s \times \mb{p}$ denotes the linear velocity at the origin due to the rotational movement about the axis. 

\subsubsection{Screw Axis Representation Simplification}

The above \textit{screw axis} vector $\mb{s}$ and \textit{rotational velocity} scalar $\dot{q}$ representations carry concrete physical meanings. Meanwhile, given a random \textit{twist velocity} vector $\bs{\nu}$, calculating the configurations of the \textit{screw axis} vector $\mb{s}$ and \textit{rotational velocity} scalar $\dot{q}$ that can make the equation hold is not an easy task. In this part, we propose to simplify the above calculation process.
\vspace{5pt}

\noindent \textbf{Simplified Version 1}: In application, for any \textit{twist velocity} vector $\mb{\nu} = \begin{bmatrix}
\bs{\omega}\\
\mb{v}\\
\end{bmatrix}$ input,
 \begin{itemize}
 \item if $\bs{\omega} \neq \mb{0}$, there also exist an equivalent representation of the \textit{screw axis} vector $\mb{s}$ and \textit{rotational velocity} scalar $\dot{q}$ that can make Equation~\ref{equ:screw_representation} hold, where $\mb{e}_s = \bs{\omega}/\left\| \bs{\omega}  \right\|$, $\dot{q} = \left\| \bs{\omega}  \right\|$, and $h = \mb{e}_s^\top \mb{v}/\dot{q}$. Meanwhile, the position vector $\mb{p}$ is chosen so that the term $-\dot{q} \mb{e}_s \times \mb{p}$ provides the portion of $\mb{v}$ orthogonal to the screw axis;
 \item if $\bs{\omega} = \mb{0}$, then the pitch $h$ will be $+\infty$ and $\mb{e}_s = \mb{v}/\left\| \mb{v} \right\|$ and $\dot{q}$ is interpreted as the linear velocity $\left\| \mb{v} \right\|$ along $\mb{e}_s$.
\end{itemize}

\vspace{5pt}
The readers may have also observe some potential problems with the above simplified representation of the \textit{twist velocity} vector as the \textit{screw axis} and \textit{angular velocity} representation, {\eg}, the pitch $h$ can be $+\infty$, vector $\mb{p}$ is not unique, and the calculation process is very cumbersome. Therefore, we will introduce to define the \textit{screw axis} vector $\mb{s}$ as a normalized version of any \textit{twist velocity} vector $\bs{\nu}$ corresponding to the motion along the screw.
\vspace{5pt}

\noindent \textbf{Simplified Version 2}: For any \textit{twist velocity} vector $\mb{\nu} = \begin{bmatrix}
\bs{\omega}\\
\mb{v}\\
\end{bmatrix}$ input,
\begin{itemize}
\item if $\bs{\omega} \neq \mb{0}$, we define the \textit{screw axis} $\mb{s} = \mb{\nu}/\left\| \bs{\omega} \right\| = \begin{bmatrix}
\frac{\bs{\omega}}{\left\| \bs{\omega} \right\|}\\
\frac{\mb{v}}{\left\| \bs{\omega} \right\|}\\
\end{bmatrix}$, and the angular velocity scalar about the axis as $\dot{q} = \left\| \bs{\omega} \right\|$.
\item if $\bs{\omega} = \mb{0}$, we define the \textit{screw axis} $\mb{s} = \mb{\nu}/\left\| \mb{v} \right\| = \begin{bmatrix}
\mb{0}\\
\frac{\mb{v}}{\left\| \mb{v} \right\|}\\
\end{bmatrix}$, and the angular velocity scalar about the axis as $\dot{q} = \left\| \mb{v} \right\|$.
\end{itemize}

Based on the above simplified version, we can provide the formal definition of \textit{screw axis} as follows:
\begin{definition}
(\textbf{Screw Axis}): For a given coordinate system, a \textit{screw axis} can be formally defined as 
\begin{equation}
\mb{s} = \begin{bmatrix}
\bs{\omega}\\
\mb{v}\\
\end{bmatrix} \in \mathbbm{R}^6,
\end{equation}
where either (1) $\left\| \bs{\omega} \right\| = 1$ or (2) $\bs{\omega} = \mb{0}$ and $\left\| \mb{v} \right\| = 1$.
\end{definition}

According to the above definition, the \textit{screw axis} vector is actually a normalized version of the \textit{twist velocity} vector, and we have the equation $\bs{\nu} = \dot{q} \mb{s}$ hold. Also many of the properties, operators and transformation that can be applied to the \textit{twist velocity} vector will also hold and applied to the \textit{screw axis} vector, {\eg}, the $\land$ operator.

Given a \textit{screw axis} representation $\mb{s} = \begin{bmatrix}
\bs{\omega}\\
\mb{v}\\
\end{bmatrix}$, we can represent its matrix representation subject to the $\land$ operator as follows:
\begin{equation}
\widehat{\mb{s}} = \begin{bmatrix}
\widehat{\bs{\omega}} & \mb{v}\\
\mb{0} & 0\\
\end{bmatrix}, \text{ where }
\widehat{\bs{\omega}} = \begin{bmatrix}
0 & - \omega_3 & \omega_2\\
\omega_3 & 0 & -\omega_1\\
-\omega_2 & \omega_1 & 0\\
\end{bmatrix}.
\end{equation}

With the \textit{adjoint representation} of the \textit{homogeneous transformation matrix}, we can also convert the \textit{screw axis} vectors across coordinate systems, {\eg},
\begin{equation}
\widehat{\mb{s}}^{[\Sigma_{\textsc{w}}]} = Ad(\mb{T}^{[\Sigma_a \to \Sigma_{\textsc{w}}]}) \widehat{\mb{s}}^{[\Sigma_a]} \text{, and } 
\widehat{\mb{s}}^{[\Sigma_a]} = Ad(\mb{T}^{[\Sigma_{\textsc{w}} \to \Sigma_a]}) \widehat{\mb{s}}^{[\Sigma_{\textsc{w}}]},
\end{equation}
where $\widehat{\mb{s}}^{[\Sigma_{\textsc{w}}]}$ and $\widehat{\mb{s}}^{[\Sigma_a]}$ denote two \textit{screw axis} vectors in the coordinate systems $\Sigma_{\textsc{w}}$ and $\Sigma_a$, respectively. Terms $\mb{T}^{[\Sigma_a \to \Sigma_{\textsc{w}}]}$ and $\mb{T}^{[\Sigma_{\textsc{w}} \to \Sigma_a]}$ represent the \textit{homogeneous transformation matrices} between these two coordinate systems.

\subsection{Motion Exponential and Logarithm}\label{subsec:motion_exponential_logarithm}

In the previous tutorial article \cite{RobotBasics_2022}, we have introduced that the \textit{rotation matrix} $\mb{R}$ can be represented as the exponential of its corresponding \textit{angular velocity vector} $\bs{\omega}$, {\ie},
\begin{equation}
\mb{R}(t) = \exp^{\widehat{\bs{\omega}}t}.
\end{equation} 

Since we have modeled the rigid-body motion of robots as the rotational movement in this section, the readers probably may also wonder if the \textit{homogeneous transformation matrix} $\mb{T}$ will also have similar exponential representations of the \textit{screw axis} representation $\mb{s}$ or not? This is what we plan to introduce in this part.

\subsubsection{Motion Exponential Representation}

We have spent lots of space introducing the rigid-body \textit{robot motion}, \textit{twist velocity} and the \textit{screw axis} before already. Here, instead of showing the detailed derivation steps, we will directly define the \textit{exponential operator} that will project the \textit{screw axis} to the \textit{homogeneous transformation matrix}.

\begin{definition}
(\textbf{Exponential Operator}): Given the \textit{screw axis} $\mb{s} = \begin{bmatrix}
\bs{\omega}\\
\mb{v}\\
\end{bmatrix} \in \mathbbm{R}^6$ and the rotation angle $q$, we define a matrix exponential operator to project the \textit{screw rotation} to the corresponding  \textit{homogeneous transformation matrix} $\mb{T} \in \mathbbm{R}^{4 \times 4}$ as follows:
\begin{equation}
\exp: \widehat{\mb{s}} q \to \mb{T}.
\end{equation}

Specifically, for the \textit{screw axis} $\mb{s} = \begin{bmatrix}
\bs{\omega}\\
\mb{v}\\
\end{bmatrix}$, depending on its element values, we can also provide the closed-form representation of term $\exp^{\widehat{\mb{s}} q}$ illustrated as follows:
\begin{itemize}
\item if $\left\| \bs{\omega} \right\|=1$, we can also provide the closed-form representation of the exponential term $\exp^{\widehat{\mb{s}} q}$ as follows:
\begin{equation}
\exp^{\widehat{\mb{s}} q} = \begin{bmatrix}
\exp^{\widehat{\bs{\omega}}q} & \left( \mb{I} q + (1 - \cos q) \widehat{\bs{\omega}} + (q - \sin q) \widehat{\bs{\omega}}^2 \right) \mb{v} \\
\mb{0} & 1\\
\end{bmatrix}.
\end{equation}

\item if the \textit{angular velocity} vector $\bs{\omega} = \mb{0}$ and $\left\| \mb{v} \right\| =1$, the above matrix representation can also be simplified as
\begin{equation}
\exp^{\widehat{\mb{s}} q} = \begin{bmatrix}
\mb{i} & \mb{v} q \\
\mb{0} & 1\\
\end{bmatrix}.
\end{equation}
\end{itemize}
\end{definition}

Based on the exponential operator, given a \textit{screw axis} vector, we can calculate the corresponding \textit{homogeneous transformation matrix} that can describe the identical robot motion across coordinate systems.

\subsubsection{Matrix Logarithm Representation}

Let's also look at the reversed direction. Given a \textit{homogeneous transformation matrix} $\mb{T} = \begin{bmatrix} \mb{R} & \mb{p}\\ \mb{0} & 0\end{bmatrix}$, can we also find a \textit{screw axis} $\mb{s} \in \mathbbm{R}^6$ and a rotation angle scalar $q \in \mathbbm{R}$ such that $\exp^{\widehat{\mb{s}} q} = \mb{T}$? Besides the \textit{exponential operator}, we will also introduce the a \textit{logarithm operator} here to achieve such an objective.

\begin{definition}
(\textbf{Logarithm Operator}): Given a \textit{homogeneous transformation matrix} $\mb{T} = \begin{bmatrix} \mb{R} & \mb{p}\\ \mb{0} & 0\end{bmatrix}$, via the \textit{logarithm operator}, we can project $\mb{T}$ to a \textit{screw axis} $\mb{s} \in \mathbbm{R}^6$ and a rotation angle scalar $q \in \mathbbm{R}$ shown as follows:
\begin{equation}
\ln: \mb{T} \to \widehat{\mb{s}} q.
\end{equation}

Specifically, depending on the values in matrix $\mb{T}$, we can also provide the closed-form representation of the corresponding \textit{screw axis} $\mb{s} \in \mathbbm{R}^6$ and rotation angle scalar $q \in \mathbbm{R}$:
\begin{itemize}
\item if $\mb{R} = \mb{I}$, then we can set $\bs{\omega} = \mb{0}$, $\mb{v} = \mb{p}/\left\| \mb{p} \right\|$, and $q = \left\| \mb{p} \right\|$.
\item otherwise, we can use the \textit{logarithm operator} defined before for \textit{rotation matrix} to calculate the \textit{angular velocity} vector and the angle scalar $q$ as
\begin{align}
\bs{\omega} &= \left( \ln \mb{R} \right)^{\lor}\\
q& = \text{atan2} \left(\left\|( \mb{R}_{32}-\mb{R}_{23}, \mb{R}_{13}-\mb{R}_{31}, \mb{R}_{21}-\mb{R}_{12} )^\top \right\|, \mb{R}_{11}+\mb{R}_{22}+\mb{R}_{33}-1 \right).
\end{align}
As to the \textit{linear velocity} vector $\mb{v}$, we can represent it as
\begin{equation}
\mb{v} = \mb{G}^{-1} \mb{p} \text{, where matrix } \mb{G} = \mb{I} q + (1 - \cos q) \widehat{\bs{\omega}} + (q - \sin q) \widehat{\bs{\omega}}^2.
\end{equation}
\end{itemize}
\end{definition}

\subsection{Wrench}

At the end of this section, we also plan to provide a brief introduction about the \textit{wrench} of robot motion, which illustrate the relationship between forces and torques. More detailed information about force and torque will be discussed in the last section about \textit{robot dynamics} for readers.

\subsubsection{Wrench Definition}

From basic knowledge we learn from physics, given a force denoted by vector $\mb{f}_e$ acting on the end point $e$ of a robot body, whose position can be denoted as vector $\mb{p}_e$, we can represent the generated \textit{torque} as
\begin{equation}
\bs{\tau}_e = \mb{p}_e \times \mb{f}_e.
\end{equation}

Similar to \textit{twist velocity} vector, the force and generated torque vectors can also be organized into a vector of $6$ elements, which is formally defined as the \textit{wrench vector} as follows:
\begin{equation}
\mb{w}_e = \begin{bmatrix}
\bs{\tau}_e\\
\mb{f}_e\\
\end{bmatrix} \in \mathbbm{R}^6.
\end{equation}

For the multiple \textit{wrenches} are acting on the same body, the total \textit{wrench} is defined as the sum of these individual \textit{wrench vectors}. Readers probably have also noticed that the above \textit{wrench vector} is very similar to the \textit{twist velocity} vector, and the \textit{torque} vector can also be expressed as the cross-product of the \textit{force} and the \textit{position} vectors. Meanwhile, they also have minor differences, {\eg}, (1) $\bs{\tau} = \mb{p} \times \mb{f}$ (vector $\mb{p}$ is multiplied at the left side) but $\mb{v} = \bs{\omega} \times \mb{p}$ (vector $\mb{p}$ at the right side); and (2) in the wrench vector $\mb{w}$, term $\bs{\tau}$ obtained via $\mb{p} \times \mb{f}$ is at the top, but in the twist velocity vector $\bs{\nu}$, the linear velocity $\mb{v}$ obtained by cross product term $\bs{\omega} \times \mb{p}$ is at the bottom instead. Due to these differences, we cannot simply apply the operators introduced before to the \textit{wrench vector} here.

\subsubsection{Wrench Transformation}

Actually, via the power quantity that drives the motion, we can further illustrate the relationships between \textit{wrench} and \textit{twist} more clearly. The \textit{wrench} vectors in different coordinate systems can also be converted into each other via the \textit{homogeneous transformation matrix}. Given two coordinate systems $\Sigma_a$ and $\Sigma_b$, we can represent two \textit{wrench vectors} acting on the end point $e$ defined in each of them as $\mb{w}_e^{[\Sigma_a]}$ and $\mb{w}_e^{[\Sigma_b]}$, respectively. At the same time, we can also represent the \textit{twist vectors} of the end point $e$ in these two coordinate systems as vectors $\bs{\nu}_e^{[\Sigma_a]}$ and $\bs{\nu}_e^{[\Sigma_b]}$. We know that the power that drives the motion of the rigid-body robot can be represented as
\begin{equation}
P = \mb{f}_e^\top \times \mb{v}_e = \bs{\tau}_e^\top \bs{\omega}_e.
\end{equation}
Also the power $P$ is scalar that is independent of the coordinate systems. Viewed in such a perspective, we can get that
\begin{equation}\label{equ:power_equal}
\underbrace{(\bs{\nu}_e^{[\Sigma_a]})^\top \mb{w}_e^{[\Sigma_a]}}_{(\bs{\tau}_e^{[\Sigma_a]})^\top \bs{\omega}_e^{[\Sigma_a]}+(\mb{f}_e^{[\Sigma_a]})^\top \times \mb{v}_e^{[\Sigma_a]}  =2P} = \underbrace{(\bs{\nu}_e^{[\Sigma_b]})^\top \mb{w}_e^{[\Sigma_b]}}_{(\bs{\tau}_e^{[\Sigma_b]})^\top \bs{\omega}_e^{[\Sigma_b]}+(\mb{f}_e^{[\Sigma_b]})^\top \times \mb{v}_e^{[\Sigma_b]}  =2P}.
\end{equation} 

According to the above Equation~\ref{equ:twist_transformation_1} and Equation~\ref{equ:twist_transformation_2}, we know that
\begin{equation}
\bs{\nu}_e^{[\Sigma_b]} = Ad(\mb{T}^{[\Sigma_a \to \Sigma_b]}) \bs{\nu}_e^{[\Sigma_a]}
\end{equation}
By replacing the above representation of $\bs{\nu}_e^{[\Sigma_b]}$ into Equation~\ref{equ:power_equal}, we will have
\begin{align}
(\bs{\nu}_e^{[\Sigma_a]})^\top \mb{w}_e^{[\Sigma_a]} &= (\bs{\nu}_e^{[\Sigma_b]})^\top \mb{w}_e^{[\Sigma_b]} \\
&= \left( Ad(\mb{T}^{[\Sigma_a \to \Sigma_b]}) \bs{\nu}_e^{[\Sigma_a]} \right)^\top \mb{w}_e^{[\Sigma_b]} \\
&= (\bs{\nu}_e^{[\Sigma_a]})^\top Ad(\mb{T}^{[\Sigma_a \to \Sigma_b]})^\top \mb{w}_e^{[\Sigma_b]}.
\end{align}
Since the above equation should hold for any \textit{twist velocity} vector $\bs{\nu}_e^{[\Sigma_a]} \in \mathbbm{R}^6$, it can lead to
\begin{equation}
\mb{w}_e^{[\Sigma_a]} = Ad(\mb{T}^{[\Sigma_a \to \Sigma_b]})^\top \mb{w}_e^{[\Sigma_b]}.
\end{equation}
With the same method, we can also derive that
\begin{equation}
\mb{w}_e^{[\Sigma_b]} = Ad(\mb{T}^{[\Sigma_b \to \Sigma_a]})^\top \mb{w}_e^{[\Sigma_a]}.
\end{equation}


\section{Forward Kinematics}\label{sec:forward_kinematics}

In this section, we will talk about the \textit{forward kinematics} for readers on robot motion. Based on the robot motion introduced in the previous article \cite{RobotBasics_2022}, in this part, we will show readers how the robot body (involving multiple joints and links) will move to the desired position and state. 


\subsection{What is Forward Kinematics?}

The rigid robot body can actually be viewed as mapping, which projects the input motor movement parameters ({\eg}, desired angles $\mb{q}$, velocity $\dot{\mb{q}}$ and acceleration $\ddot{\mb{q}}$) to the desired robot position and state ({\eg}, position $\mb{p}$, orientation $\mb{R}$, velocity $\mb{v}$ and force $\mb{f}$ of certain robot body parts).

\begin{definition}
(\textbf{Forward Kinematics}): Formally, \textit{forward kinematics} is a calculation to obtain the desired robot position, orientation and other properties of links in the body from a given input parameter.

If we represent the input parameters as a vector $\bs{\theta}(t)$, subject the pre-defined rigid robot body, we can represent the desired robot outputs as
\begin{equation}\label{equ:forward_kinematics}
\mb{o}(t) = g \left( \bs{\theta}(t) \right).
\end{equation}
Since the input parameter and the desired output are both changing with time, we specify that vectors $\bs{\theta}(t)$ and $\mb{o}(t)$ both have time as their variable. Meanwhile, the calculation process of the mapping $g \left(\cdot\right)$ to achieve the output is called the \textit{forward kinematics} process.
\end{definition}

Robot \textit{forward kinematics} serves as the basis for robot simulation and control. It helps display the current state of the robot, calculate the center of mass of the robot body and detect potential collisions of the robot with its body parts and the environment. In this section, we will introduce the \textit{forward kinematics} of robots in terms of the \textit{velocity}, \textit{orientation}, \textit{velocity} \textit{force} and \textit{torque} for readers, respectively.


\subsection{Pose in Forward Kinematics}\label{subsec:position_orientation_kinematics}

We have been discussing about the \textit{position} and \textit{orientation} changes of robot parts in both the previous article \cite{RobotBasics_2022} and in the previous Section~\ref{sec:robot_motion} already. Here, we will just provide a very brief introduction about updating robot \textit{pose}, covering both \textit{position} and \textit{orientation}, in forward kinematics with the \textit{chain rule} and \textit{exponential representation}, respectively.


\begin{figure}[t]
    \centering
    \includegraphics[width=0.9\textwidth]{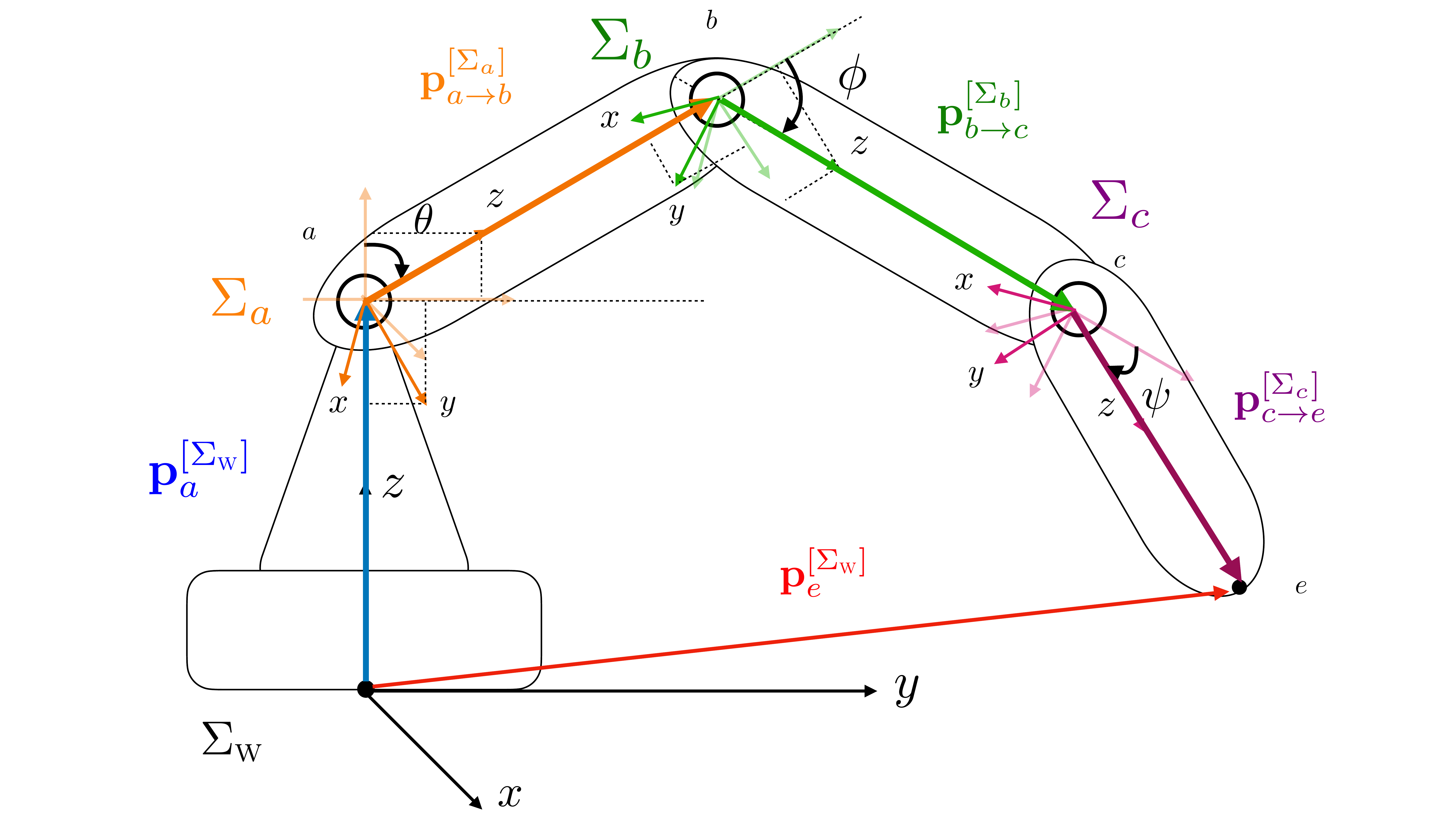}
    \caption{The Robot Arm Example We Use Before.}
    \label{fig:robot_arm_use_here_again}
\end{figure}

\subsubsection{Chain Rule}

According to our previous discussions, readers should be able to understand how the \textit{rotation matrix} projects \textit{position} vectors across coordinate systems. In some sense, the \textit{rotation matrix} also indicates the \textit{orientation} (relative to the world coordinate system) of robot actually within local or world coordinate systems.

\begin{definition}
(\textbf{Pose}): The combination of both \textit{position} and \textit{orientation} defines the \textit{pose} of the robot, which can be represented as the \textit{position} vector and \textit{rotation} matrix pair $\left(\mb{p}, \mb{R} \right)$. Since the \textit{homogeneous transformation matrix} $\mb{T} = \begin{bmatrix} \mb{R} & \mb{p} \\ \mb{0} & 1 \end{bmatrix}$ covers both $\mb{p}$ and $\mb{R}$, we can also say the \textit{homogeneous transformation matrix} $\mb{T}$ represents the \textit{pose} of robots.
\end{definition}

As introduced in the previous section, for the robot arm shown in Figure~\ref{fig:robot_arm}, the position and orientation of its end point can be calculated via the \textit{chain rule}, which will obtain the \textit{homogeneous transition matrix} $\mb{T}^{[\Sigma_c \to \Sigma_{\textsc{w}}]}$:
\begin{align}
&\mb{T}^{[\Sigma_c \to \Sigma_{\textsc{w}}]} \\
&= \mb{T}^{[\Sigma_a \to \Sigma_{\textsc{w}}]} \mb{T}^{[\Sigma_b \to \Sigma_{a}]} \mb{T}^{[\Sigma_c \to \Sigma_b]} \\
&= \begin{bmatrix}
\mb{R}^{[\Sigma_a \to \Sigma_{\textsc{w}}]} & \mb{p}_a^{[\Sigma_{\textsc{w}}]} \\
\mb{0} & 1\\
\end{bmatrix}
\begin{bmatrix}
\mb{R}^{[\Sigma_b \to \Sigma_{a}]} & \mb{p}_{a \to b}^{[\Sigma_{a}]} \\
\mb{0} & 1\\
\end{bmatrix}
\begin{bmatrix}
\mb{R}^{[\Sigma_c \to \Sigma_b]} & \mb{p}_{b \to c}^{[\Sigma_b]} \\
\mb{0} & 1\\
\end{bmatrix} \\
&= \begin{bmatrix}
\mb{R}^{[\Sigma_a \to \Sigma_{\textsc{w}}]} \mb{R}^{[\Sigma_b \to \Sigma_{a}]} \mb{R}^{[\Sigma_c \to \Sigma_b]} & \mb{p}_a^{[\Sigma_{\textsc{w}}]} +  \mb{R}^{[\Sigma_a \to \Sigma_{\textsc{w}}]} \mb{p}_{a \to b}^{[\Sigma_{a}]} + \mb{R}^{[\Sigma_a \to \Sigma_{\textsc{w}}]} \mb{R}^{[\Sigma_b \to \Sigma_{a}]} \mb{p}_{b \to c}^{[\Sigma_b]} \\
\mb{0} & 1\\
\end{bmatrix}
\end{align} 

Since the rigid-body robot arm is normally straight, the orientation of the end point is actually the same as the orientation of the joint that controls the last arm link. From the above representation of the \textit{homogeneous transformation matrix} $\mb{T}^{[\Sigma_c \to \Sigma_{\textsc{w}}]}$, we can denote the orientation of the end point of the arm with the rotation matrix
\begin{equation}
\mb{R}^{[\Sigma_c \to \Sigma_{\textsc{w}}]} = \mb{R}^{[\Sigma_a \to \Sigma_{\textsc{w}}]} \mb{R}^{[\Sigma_b \to \Sigma_{a}]} \mb{R}^{[\Sigma_c \to \Sigma_b]}.
\end{equation}

As introduced in the previous section, based on the end point's current position vector within the local coordinate $\Sigma_c$, we can calculate its end point's position within the world coordinate as
\begin{equation}
\begin{bmatrix}
\mb{p}_e^{[\Sigma_\textsc{w}]}\\
1
\end{bmatrix} = \mb{T}^{[\Sigma_c \to \Sigma_{\textsc{w}}]}
\begin{bmatrix}
\mb{p}_{c \to e}^{[\Sigma_c]}\\
1
\end{bmatrix}.
\end{equation}
The vector $\mb{p}_e^{[\Sigma_\textsc{w}]}$ indicating the robot arm end point's position can also be represented as the following summation formula:
\begin{align}
\mb{p}_e^{[\Sigma_\textsc{w}]} = & \mb{p}_a^{[\Sigma_{\textsc{w}}]} +  \mb{R}^{[\Sigma_a \to \Sigma_{\textsc{w}}]} \mb{p}_{a \to b}^{[\Sigma_{a}]} + \mb{R}^{[\Sigma_a \to \Sigma_{\textsc{w}}]} \mb{R}^{[\Sigma_b \to \Sigma_{a}]} \mb{p}_{b \to c}^{[\Sigma_b]} \\
&+ \mb{R}^{[\Sigma_a \to \Sigma_{\textsc{w}}]} \mb{R}^{[\Sigma_b \to \Sigma_{a}]} \mb{R}^{[\Sigma_c \to \Sigma_b]} \mb{p}_{c \to e}^{[\Sigma_c]} .
\end{align}


\subsubsection{Product of Exponentials}

In the above derivation, we assume the rotation matrix for each joint is known and can be used for calculating both the orientation and position of the robot arm end point. Meanwhile, when the rotation matrices are unknown, we can still calculate the end point orientation and position via the exponential operator we introduce in the previous article \cite{RobotBasics_2022} based on the rotation angles or the joints.

As introduced in \cite{RobotBasics_2022}, the \textit{homogeneous transformation matrix} of the robot arm joint can be represented with its \textit{angular velocity vector} with the \textit{exponential operator} as follows:
\begin{align}
\mb{T}^{[\Sigma_a \to \Sigma_{\textsc{w}}]} = \begin{bmatrix}
\mb{R}^{[\Sigma_a \to \Sigma_{\textsc{w}}]} & \mb{p}_a^{[\Sigma_{\textsc{w}}]} \\
\mb{0} & 1\\
\end{bmatrix} = \begin{bmatrix}
\exp^{\widehat{\mb{e}}_a^{[\Sigma_{\textsc{w}}]} \dot{q}_a} & \mb{p}_a^{[\Sigma_{\textsc{w}}]} \\
\mb{0} & 1\\
\end{bmatrix},
\end{align}
where vector ${\mb{e}}_a^{[\Sigma_{\textsc{w}}]}$ and scalar $\dot{q}_a$ denote the rotation axis and rotation velocity, respectively.

In a similar way, we can also represent the \textit{rotation matrix} and \textit{homogeneous transformation matrix} with origins at other joints with their corresponding \textit{angular velocity vector} as follows:
\begin{align}
\mb{T}^{[\Sigma_b \to \Sigma_a]} &= \begin{bmatrix}
\mb{R}^{[\Sigma_b \to \Sigma_a]} & \mb{p}_{a \to b}^{[\Sigma_a]} \\
\mb{0} & 1\\
\end{bmatrix} = \begin{bmatrix}
\exp^{\widehat{\mb{e}}_b^{[\Sigma_a]} \dot{q}_b} & \mb{p}_{a \to b}^{[\Sigma_a]} \\
\mb{0} & 1\\
\end{bmatrix},\\
\mb{T}^{[\Sigma_c \to \Sigma_b]} &= \begin{bmatrix}
\mb{R}^{[\Sigma_c \to \Sigma_b]} & \mb{p}_{b \to c}^{[\Sigma_b]} \\
\mb{0} & 1\\
\end{bmatrix} = \begin{bmatrix}
\exp^{\widehat{\mb{e}}_c^{[\Sigma_b]} \dot{q}_c} & \mb{p}_{b \to c}^{[\Sigma_b]} \\
\mb{0} & 1\\
\end{bmatrix}.
\end{align}

By replacing them into the rotation matrix and end point position vector calculation, we can obtain their representations as follows:
\begin{align}
\mb{R}^{[\Sigma_c \to \Sigma_{\textsc{w}}]} &= \left( \exp^{\widehat{\mb{e}}_a^{[\Sigma_{\textsc{w}}]} \dot{q}_a} \exp^{\widehat{\mb{e}}_b^{[\Sigma_a]} \dot{q}_b} \exp^{\widehat{\mb{e}}_c^{[\Sigma_b]} \dot{q}_c} \right), \\
\mb{p}_e^{[\Sigma_\textsc{w}]} &= \mb{p}_a^{[\Sigma_{\textsc{w}}]} +  \left( \exp^{\widehat{\mb{e}}_a^{[\Sigma_{\textsc{w}}]} \dot{q}_a} 
\mb{p}_{a \to b}^{[\Sigma_{a}]} \right) + \left( \exp^{\widehat{\mb{e}}_a^{[\Sigma_{\textsc{w}}]} \dot{q}_a} \exp^{\widehat{\mb{e}}_b^{[\Sigma_a]} \dot{q}_b} \mb{p}_{b \to c}^{[\Sigma_b]} \right) \\
&+ \left( \exp^{\widehat{\mb{e}}_a^{[\Sigma_{\textsc{w}}]} \dot{q}_a} 
\exp^{\widehat{\mb{e}}_b^{[\Sigma_a]} \dot{q}_b} 
\exp^{\widehat{\mb{e}}_c^{[\Sigma_b]} \dot{q}_c} 
\mb{p}_{c \to e}^{[\Sigma_c]} \right).
\end{align}
The above representations of rotation matrix and end point position vector is also named as the \textit{product of exponentials} of the joint angular velocity vectors.


\subsection{Velocity in Forward Kinematics}\label{subsec:velocity_kinematics}

Let's revisit the Equaton~\ref{equ:forward_kinematics} provided at the beginning of this section, if we take the desired output vector as the position vector $\mb{p}(t)$ and the input as the angles of the joints $\mb{q}(t)$, we can represent the desired output as follows:
\begin{equation}
\mb{p}(t) = g \left( \mb{q}(t) \right).
\end{equation}


\subsubsection{Jacobian Matrix}

By taking the derivatives of both size of the above equation with regard to the time variable $t$, we can obtain
\begin{align}\label{equ:jacobian_matrix}
\dot{\mb{p}}(t) &= \frac{\partial g \left( \mb{q} \right) }{\partial \mb{q}} \frac{\partial \mb{q}(t)}{\partial t} = \frac{\partial g \left( \mb{q} \right) }{\partial \mb{q}} \dot{\mb{q}}(t) \\
&= \mb{J} \dot{\mb{q}}(t),
\end{align}
where term $\dot{\mb{p}}(t)$ denotes the \textit{linear velocity} vector and the matrix $\mb{J}=\frac{\partial g \left( \mb{q} \right) }{\partial \mb{q}}$ about the joint variables $\mb{q}$ is called the Jacobian matrix.

\begin{example}
For instance, let's assume the length of the robot arm links starting from the origin $o$, joints $a$, $b$ and $c$ are $l_1$, $l_2$, $l_3$ and $l_4$, respectively. Based on the rotation angles of these links, we can provide the analytic representation of the end point position in the world coordinate system as
\begin{align}
x_e &= 0,\\
y_e &= l_2 \sin \theta + l_3 \sin (\theta + \phi) + l_4 \sin (\theta + \phi + \psi),\\
z_e &= l_1 + l_2 \cos \theta + l_3 \cos(\theta + \phi) + l_4 \cos (\theta + \phi + \psi).
\end{align} 

By taking derivatives of both sides of the above equations, we can obtain
\begin{align}
\dot{x}_e &= 0,\\
\dot{y}_e &= l_2 \dot{\theta} \cos \theta + l_3 (\dot{\theta} + \dot{\phi}) \cos (\theta + \phi)  + l_4 (\dot{\theta} + \dot{\phi} + \dot{\psi}) \cos (\theta + \phi + \psi),\\
\dot{z}_e &= -l_2 \dot{\theta} \sin \theta - l_3 (\dot{\theta} + \dot{\phi}) \sin (\theta + \phi) - l_4 (\dot{\theta} + \dot{\phi} + \dot{\psi}) \sin (\theta + \phi + \psi).
\end{align}
\end{example}

The above equations can also be represented in the form of $\dot{\mb{p}}=\mb{J}\dot{\mb{q}}$ as follows:
\begin{equation}
\underbrace{\begin{bmatrix}
\dot{x}_e\\
\dot{y}_e\\
\dot{z}_e\\
\end{bmatrix}}_{\dot{\mb{p}}}
=
\underbrace{\begin{bmatrix}
0 & 0 & 0\\
\mb{J}_{2,1} & \mb{J}_{2,2} & \mb{J}_{2,3}\\
\mb{J}_{3,1} & \mb{J}_{3,2}  & \mb{J}_{3,3}\\
\end{bmatrix}}_{\mb{J}}
\underbrace{\begin{bmatrix}
\dot{\theta}\\
\dot{\phi}\\
\dot{\psi}\\
\end{bmatrix}}_{\dot{\mb{q}}},
\end{equation}
where the matrix elements
\begin{align}
\mb{J}_{2,1} &= l_2\cos \theta + l_3\cos (\theta + \phi) + l_4\cos (\theta + \phi + \psi),\\
\mb{J}_{2,2} &= l_3\cos (\theta + \phi) + l_4\cos (\theta + \phi + \psi),\\
\mb{J}_{2,3} &= l_4\cos (\theta + \phi + \psi),\\
\mb{J}_{3,1} &= -l_2\sin \theta-l_3\sin (\theta + \phi)-l_4\sin (\theta + \phi + \psi),\\
\mb{J}_{3,2} &= -l_3\sin (\theta + \phi)-l_4\sin (\theta + \phi + \psi),\\
\mb{J}_{3,3} &= -l_4\sin (\theta + \phi + \psi).
\end{align}

\begin{example}
For instance, if the arm link length $l_1 = l_2 = l_3 = l_4 = 1$, and the arm joint angles $\theta=\frac{\pi}{4}$, $\phi=\frac{\pi}{2}$ and $\psi=\frac{3\pi}{4}$, we can represent the corresponding \textit{Jacobian matrix} as follows:
\begin{align}
\mb{J} &= \begin{bmatrix}
0 & 0 & 0 \\
\left( \cos \frac{\pi}{4} + \cos \frac{3\pi}{4} + \cos \frac{3\pi}{2} \right) & \left( \cos \frac{3\pi}{4} + \cos \frac{3\pi}{2} \right) & \left( \cos \frac{3\pi}{2} \right) \\
\left( -\sin \frac{\pi}{4} - \sin \frac{3\pi}{4} - \sin \frac{3\pi}{2} \right) & \left( - \sin \frac{3\pi}{4} - \sin \frac{3\pi}{2} \right) & \left( - \sin \frac{3\pi}{2} \right) \\
\end{bmatrix} \\
&= \begin{bmatrix}
0 & 0 & 0\\
0 & -\frac{\sqrt{2}}{2} & 0\\
1-\sqrt{2} &1-\frac{\sqrt{2}}{2}  & 1\\
\end{bmatrix}.
\end{align}
\end{example}


\subsubsection{Velocity Kinematics in World Coordinate System}

As introduced in Section~\ref{subsubsec:twist_world}, the \textit{linear velocity} and \textit{angular velocity} of robots can be described with its \textit{twist velocity}, which can be represented with the \textit{homogeneous transformation matrix} effectively. In this part, we will discuss about the robot \textit{velocity} within the world coordinate system, whereas the \textit{velocity kinematics} within the local coordinate system will be introduced in the following subsection.

According to Equation~\ref{equ:twist_world_coordinate_system} introduced in the previous Section~\ref{subsubsec:twist_world}, the \textit{twist velocity} vector of the robot end point can be represented as
\begin{equation}
\widehat{\bs{\nu}}_e^{[\Sigma_{\textsc{w}}]} = \dot{\mb{T}}^{[\Sigma_c \to \Sigma_{\textsc{w}}]} \left(\mb{T}^{[\Sigma_c \to \Sigma_{\textsc{w}}]} \right)^{-1}.
\end{equation}


Meanwhile, as introduced in the previous Section~\ref{subsec:motion_exponential_logarithm}, the \textit{homogeneous transformation matrix} can also be effectively represented as the \textit{screw axis} and \textit{rotation angle} with the \textit{exponential operator}. Viewed in such a perspective, we can represent the matrix ${\mb{T}}^{[\Sigma_c \to \Sigma_{\textsc{w}}]}$ as follows:
\begin{align}
{\mb{T}}^{[\Sigma_c \to \Sigma_{\textsc{w}}]} &= \mb{T}^{[\Sigma_a \to \Sigma_{\textsc{w}}]} \mb{T}^{[\Sigma_b \to \Sigma_{a}]} \mb{T}^{[\Sigma_c \to \Sigma_b]} \\
&= \exp^{\widehat{\mb{s}}_a q_a} \exp^{\widehat{\mb{s}}_b q_b} \exp^{\widehat{\mb{s}}_c q_c} \mb{T}_0,
\end{align}
where the constant matrix $\mb{T}_0$ to represent the initial state of the robot when $q_a = q_b = q_c = 0$.

With the above exponential representation of matrix ${\mb{T}}^{[\Sigma_c \to \Sigma_{\textsc{w}}]}$, we can represent its derivative with respect to the time as
\begin{align}
\dot{\mb{T}}^{[\Sigma_c \to \Sigma_{\textsc{w}}]} &= \left( \frac{{d}}{{d}t} \exp^{\widehat{\mb{s}}_a q_a} \right) \exp^{\widehat{\mb{s}}_b q_b} \exp^{\widehat{\mb{s}}_c q_c} \mb{T}_0 + \exp^{\widehat{\mb{s}}_a q_a} \left( \frac{{d}}{{d}t} \exp^{\widehat{\mb{s}}_b q_b} \right) \exp^{\widehat{\mb{s}}_c q_c} \mb{T}_0 \\
&+ \exp^{\widehat{\mb{s}}_a q_a} \exp^{\widehat{\mb{s}}_b q_b} \left( \frac{{d}}{{d}t} \exp^{\widehat{\mb{s}}_c q_c} \right) \mb{T}_0\\
&=\Big( \exp^{\widehat{\mb{s}}_a q_a} \widehat{\mb{s}}_a \dot{q}_a \exp^{\widehat{\mb{s}}_b q_b} \exp^{\widehat{\mb{s}}_c q_c} + \exp^{\widehat{\mb{s}}_a q_a} \exp^{\widehat{\mb{s}}_b q_b} \widehat{\mb{s}}_b \dot{q}_b \exp^{\widehat{\mb{s}}_c q_c} \\
&+ \exp^{\widehat{\mb{s}}_a q_a} \exp^{\widehat{\mb{s}}_b q_b} \exp^{\widehat{\mb{s}}_c q_c} \widehat{\mb{s}}_c \dot{q}_c \Big) \mb{T}_0.
\end{align}

On the other hand, we can also represent the inverse of matrix $\left(\mb{T}^{[\Sigma_c \to \Sigma_{\textsc{w}}]} \right)^{-1}$ as follows:
\begin{align}
\left(\mb{T}^{[\Sigma_c \to \Sigma_{\textsc{w}}]} \right)^{-1} &= \left( \mb{T}^{[\Sigma_a \to \Sigma_{\textsc{w}}]} \mb{T}^{[\Sigma_b \to \Sigma_{a}]} \mb{T}^{[\Sigma_c \to \Sigma_b]} \right)^{-1} \\
&= \left(\mb{T}^{[\Sigma_c \to \Sigma_b]} \right)^{-1} \left( \mb{T}^{[\Sigma_b \to \Sigma_{a}]} \right)^{-1} \left( \mb{T}^{[\Sigma_a \to \Sigma_{\textsc{w}}]} \right)^{-1} \\
&= \mb{T}_0^{-1} \exp^{-\widehat{\mb{s}}_c q_c} \exp^{-\widehat{\mb{s}}_b q_b} \exp^{-\widehat{\mb{s}}_a q_a}.
\end{align}

Based on the above derivation, we can formally represent the \textit{twist velocity} vector with the above exponential representation as follows:
\begin{align}
\widehat{\bs{\nu}}_e^{[\Sigma_{\textsc{w}}]} &= \dot{\mb{T}}^{[\Sigma_c \to \Sigma_{\textsc{w}}]} \left(\mb{T}^{[\Sigma_c \to \Sigma_{\textsc{w}}]} \right)^{-1} \\
&= \exp^{\widehat{\mb{s}}_a q_a} \widehat{\mb{s}}_a \dot{q}_a \exp^{-\widehat{\mb{s}}_a q_a}  +  \exp^{\widehat{\mb{s}}_a q_a} \exp^{\widehat{\mb{s}}_b q_b}\widehat{\mb{s}}_b \dot{q}_b \exp^{-\widehat{\mb{s}}_b q_b} \exp^{-\widehat{\mb{s}}_a q_a}\\
& + \exp^{\widehat{\mb{s}}_a q_a} \exp^{\widehat{\mb{s}}_b q_b} \exp^{\widehat{\mb{s}}_c q_c} \widehat{\mb{s}}_c \dot{q}_c \exp^{-\widehat{\mb{s}}_c q_c} \exp^{-\widehat{\mb{s}}_b q_b} \exp^{-\widehat{\mb{s}}_a q_a}.
\end{align}
From the previous Section~\ref{subsubsec:twist_hat_relationship} and Section~\ref{subsubsec:twist_relationship}, we already know that for $\widehat{\bs{\nu}}_a^{[\Sigma_\textsc{w}]} = \mb{T}^{[\Sigma_a \to \Sigma_{\textsc{w}}]} \widehat{\bs{\nu}}_a^{[\Sigma_a]} (\mb{T}^{[\Sigma_a \to \Sigma_{\textsc{w}}]})^{-1}$, we can get $\bs{\nu}_a^{[\Sigma_{\textsc{w}}]} = Ad({{\mb{T}}^{[\Sigma_a \to \Sigma_{\textsc{w}}]}}) \bs{\nu}_a^{[\Sigma_a]}$. Since the \textit{screw axis} is also defined as a normalized \textit{twist velocity}, with the $\lor$ operator be applied to both sides of the above equation, we can simplify the equation as  
\begin{align}
{\bs{\nu}}_e^{[\Sigma_{\textsc{w}}]} &= \underbrace{ Ad(\exp^{\widehat{\mb{s}}_a q_a}) {\mb{s}}_a}_{\mb{j}_a^{[\Sigma_a \to \Sigma_{\textsc{w}}]}} \dot{q}_a  + \underbrace{Ad(\exp^{\widehat{\mb{s}}_a q_a} \exp^{\widehat{\mb{s}}_b q_b})  {\mb{s}}_b}_{\mb{j}_b^{[\Sigma_b \to \Sigma_{\textsc{w}}]}} \dot{q}_b + \underbrace{Ad(\exp^{\widehat{\mb{s}}_a q_a} \exp^{\widehat{\mb{s}}_b q_b} \exp^{\widehat{\mb{s}}_c q_c}) {\mb{s}}_c}_{\mb{j}_c^{[\Sigma_c \to \Sigma_{\textsc{w}}]}} \dot{q}_c \\
&= \mb{j}_a^{[\Sigma_a \to \Sigma_{\textsc{w}}]} \dot{q}_a + \mb{j}_b^{[\Sigma_b \to \Sigma_{\textsc{w}}]} \dot{q}_b + \mb{j}_c^{[\Sigma_c \to \Sigma_{\textsc{w}}]} \dot{q}_c \\
&= \underbrace{\left[ \mb{j}_a^{[\Sigma_a \to \Sigma_{\textsc{w}}]}, \mb{j}_b^{[\Sigma_b \to \Sigma_{\textsc{w}}]}, \mb{j}_c^{[\Sigma_c \to \Sigma_{\textsc{w}}]} \right]}_{\mb{J}^{[\Sigma_{\textsc{w}}]}} \begin{bmatrix}
\dot{q}_a\\
\dot{q}_b\\
\dot{q}_c\\
\end{bmatrix} \\
&= \mb{J}^{[\Sigma_{\textsc{w}}]} \dot{\mb{q}}.
\end{align}

\begin{definition}
(\textbf{Space Jacobian}): Formally, the matrix 
\begin{equation}
\mb{J}^{[\Sigma_{\textsc{w}}]} = \begin{bmatrix}
Ad(\exp^{\widehat{\mb{s}}_a q_a}) & Ad(\exp^{\widehat{\mb{s}}_a q_a}\exp^{\widehat{\mb{s}}_b q_b}) {\mb{s}}_b & Ad(\exp^{\widehat{\mb{s}}_a q_a} \exp^{\widehat{\mb{s}}_b q_b} \exp^{\widehat{\mb{s}}_c q_c}) {\mb{s}}_c
\end{bmatrix}
\end{equation} 
we introduce above is also named as the \textit{space Jacobian} matrix in the world coordinate system. The column vectors in matrix $\mb{J}^{[\Sigma_{\textsc{w}}]}$ actually denote the \textit{screw axes} of different joints of the robot body in the world coordinate, respectively.
\end{definition}


\subsubsection{Velocity Kinematics in Local Coordinate System}

Meanwhile, according to Equation~\ref{equ:twist_local_coordinate_system} introduced in the previous Section~\ref{subsubsec:twist_local}, we can also obtain the \textit{twist velocity} vector representation of the end point within a local coordinate system $\Sigma_c$, which can be denoted as follows:
\begin{align}
\widehat{\bs{\nu}}_e^{[\Sigma_c]} &= \left(\mb{T}^{[\Sigma_c \to \Sigma_{\textsc{w}}]} \right)^{-1} \dot{\mb{T}}^{[\Sigma_c \to \Sigma_{\textsc{w}}]} \\
&= \left(\exp^{-\widehat{\mb{s}}_c q_c} \exp^{-\widehat{\mb{s}}_b q_b} \widehat{\mb{s}}_a \exp^{\widehat{\mb{s}}_b q_b} \exp^{\widehat{\mb{s}}_c q_c}\right) \dot{q}_a + \left(\exp^{-\widehat{\mb{s}}_c q_c} \widehat{\mb{s}}_b \exp^{\widehat{\mb{s}}_c q_c}\right) \dot{q}_b+ \left( \widehat{\mb{s}}_c \right) \dot{q}_c \\
&= \underbrace{Ad\left(\exp^{-\widehat{\mb{s}}_c q_c} \exp^{-\widehat{\mb{s}}_b q_b}\right) \widehat{\mb{s}}_a}_{\mb{j}_a^{[\Sigma_a \to \Sigma_c]}} \dot{q}_a + \underbrace{Ad\left(\exp^{-\widehat{\mb{s}}_c q_c} \right) \widehat{\mb{s}}_b}_{\mb{j}_b^{[\Sigma_b \to \Sigma_c]}} \dot{q}_b+ \underbrace{\left( \widehat{\mb{s}}_c \right)}_{\mb{j}_c^{[\Sigma_c]}} \dot{q}_c \\
&= \underbrace{\left[ \mb{j}_a^{[\Sigma_a \to \Sigma_c]}, \mb{j}_b^{[\Sigma_b \to \Sigma_c]}, \mb{j}_c^{[\Sigma_c]} \right]}_{\mb{J}^{[\Sigma_c]}} \begin{bmatrix}
\dot{q}_a\\
\dot{q}_b\\
\dot{q}_c\\
\end{bmatrix} \\
&= \mb{J}^{[\Sigma_c]} \dot{\mb{q}}.
\end{align}

\begin{definition}
(\textbf{Body Jacobian}): Formally, the matrix 
\begin{equation}
\mb{J}^{[\Sigma_c]} = \begin{bmatrix}
Ad\left(\exp^{-\widehat{\mb{s}}_c q_c} \exp^{-\widehat{\mb{s}}_b q_b}\right) & Ad\left(\exp^{-\widehat{\mb{s}}_c q_c} \right) & \left( \widehat{\mb{s}}_c \right)
\end{bmatrix}
\end{equation} 
we introduce above is also named as the \textit{body Jacobian} matrix in the local coordinate system $\Sigma_c$ that the end point lies in. The column vectors in matrix $\mb{J}^{[\Sigma_c]}$ actually denote the \textit{screw axes} of different joints of the robot body in the local coordinate $\Sigma_c$, respectively.
\end{definition}


\subsection{Torque in Forward Kinematics}\label{subsec:torque_kinematics}

The \textit{Jacobian matrix} also plays an important role in robot \textit{static analysis}. Formally, at the end of the robot arm, we can represent the generated force vector as $\mb{f}_e$ and the joint torque vector as $\bs{\tau}_e$. 


\subsubsection{Static Analysis}

According to the physics, in the rotational system, the system power equals to the product of \textit{torque vector} $\bs{\tau}_e$ and the \textit{angular velocity vector} $\bs{\omega}_e$ of the end point, {\ie},
\begin{equation}\label{equ:power_representation}
P = \bs{\tau}_e^\top \bs{\omega}_e.
\end{equation}

Meanwhile, the power of the end point can also be represented with its end point force vector $\mb{f}_e$ and its \textit{linear velocity vector} $\mb{v}_e$, {\ie},
\begin{equation}
P = \mb{f}_e^\top \mb{v}_e.
\end{equation}

These above two equations will derive that
\begin{equation}
\mb{f}_e^\top \mb{v}_e = \bs{\tau}_e^\top \bs{\omega}_e.
\end{equation}

\subsubsection{Force-Torque Relationship}\label{subsubsec:force_torque_relationship}

Meanwhile, according to the Equation~\ref{equ:jacobian_matrix} we present before when defining the \textit{Jacobian Matrix} before, we know that the \textit{linear velocity vector} $\mb{v}_e = \mb{J} \bs{\omega}_e$, so we can have the following equation,
\begin{equation}
\mb{f}_e^\top \mb{J} \bs{\omega}_e = \bs{\tau}_e^\top \bs{\omega}_e.
\end{equation}

The above equation holds for any \textit{angular velocity vector} $\bs{\omega}_e$ if 
\begin{equation}\label{equ:force_torque_relationship}
\bs{\tau}_e = \left(\mb{f}_e^\top \mb{J} \right)^\top = \mb{J}^\top \mb{f}_e.
\end{equation}

The above equation calculates the required torque $\bs{\tau}_e$ needed to generate the desired force $\mb{f}_e$ at the end point. Meanwhile, when the \textit{Jacobian Matrix} is not singular and its transpose is invertible, we can also obtain
\begin{equation}
\mb{f}_e = (\mb{J}^\top)^{-1} \bs{\tau}_e = \mb{J}^{-\top} \bs{\tau}_e.
\end{equation}
Such an equation indicates that, under the static equilibrium state, the force that can be generated at the end point given the torques of the robot arm. This equation is very important, and will be used in many scenarios. For instance, to pick up a box of certain weight $w$, we need to decide the needed torques of each joint motors so the robot can hold the box. More discussions about the robot torque and generated forces will be provided in the following sections when talking about the robot dynamics.


\section{Inverse Kinematics}\label{sec:inverse_kinematics}

In this section, we will introduce another important topic about robotic for readers, the \textit{inverse kinematics}. Just from the name, we can know that it is a reversed process of \textit{forward kinematics} introduced in the previous section. In this section, we will introduce two different methods to address the \textit{inverse kinematics} problem for readers, and also provide detailed analysis of the Jacobian singularity.

\subsection{What is Inverse Kinematics?}

We can still take the Equation~\ref{equ:forward_kinematics} we use at the beginning of the previous Section~\ref{sec:forward_kinematics} on \textit{forward kinematics} to explain ``what is \textit{inverse kinematics}'' for readers.

\begin{itemize}
\item \textbf{Forward Kinematics}: Given the parameter inputs $\mb{q}(t)$, the process of calculating the desired output $\mb{o}(t)$ of the robot body is called \textit{forward kinematics}, {\ie},
\begin{equation}
\mb{o}(t) = g \left( \mb{q}(t) \right).
\end{equation}

\item \textbf{Inverse Kinematics}: Given the desired output $\mb{O}(t)$, the problem to find the parameter solution $\mb{q}(t)$ that can satisfy 
\begin{equation}
\arg_{\mb{q}(t)} g \left( \mb{q}(t) \right) = \mb{O}(t), \forall t \in \{1, 2, \cdots\},
\end{equation}
is named as the \textit{inverse kinematics} problem.
\end{itemize}

Compared with \textit{forward kinematics} that give clear ``instructions'' (denoted by parameters $\mb{q}(t)$) to robots that will perform accordingly to illustrate the outputs, \textit{inverse kinematics} is more frequently used in real-world robot control, {\eg}, \textit{standing}, \textit{walking}, \textit{picking up items}. In this part, we will try to introduce how to find the parameter vector $\mb{q}(t)$ that will generate the output satisfying the constraint $g \left( \mb{q}(t) \right) = \mb{O}(t)$.

\subsection{Analytic Solution}

Before we introduce the \textit{numerical solution} to the \textit{inverse kinematics} problem, we would like to use a toy example to illustrate how to solve the problem analytically. 

\begin{example}
\begin{figure}[t]
    \centering
    \includegraphics[width=0.9\textwidth]{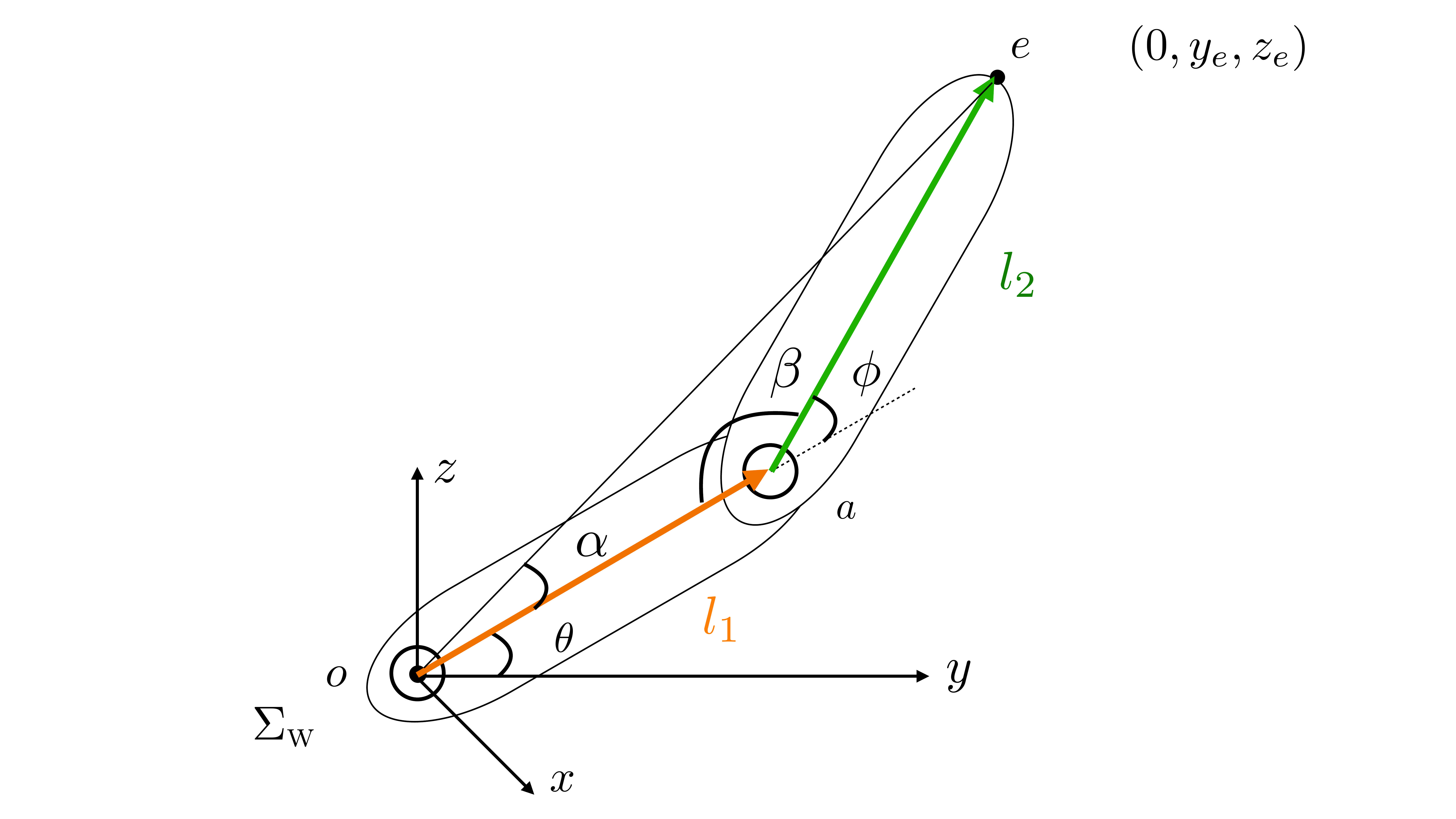}
    \caption{An Example of 2-Link Robot Ram.}
    \label{fig:analytic_solution_example}
\end{figure}

As shown in Figure~\ref{fig:analytic_solution_example}, we illustrate a robot arm in the 3D coordinate system with two links of length $l_1$ and $l_2$, respectively. For the first arm link, it forms an angle of $\theta$ degrees with respect to the $y$ axis. Meanwhile, the second arm link further forms an angle of $\phi$ degrees with the first arm link. We aim to configure the parameter values of the angles $\theta$ and $\phi$ to make the end point reaches position denoted by coordinate pair $(x_e, y_e, 0)$ in the space.
\end{example}

\noindent \textbf{Analytic Solution}: Based on the robot arm example shown in Figure~\ref{fig:analytic_solution_example}, we will show below about how to use the \textit{analytic solution} to calculate the desired angles $\theta$ and $\phi$. 

First of all, for the triangle $\triangle oae$, according to the \textit{law of cosine}, we know that
\begin{align}
&oe^2 = oa^2 + ae^2 - 2(oa)(ae) \cos \beta \\
\implies & y_e^2 + z_e^2 = l_1^2 + l_2^2 - 2 l_1 l_2 \cos \beta \\
\implies & \beta = \cos^{-1} \left( \frac{l_1^2 + l_2^2 - y_e^2 - z_e^2}{2 l_1 l_2} \right).
\end{align}

Following the same law, we can also obtain that
\begin{align}
&ae^2 = oa^2 + oe^2 - 2(oa)(oe) \cos \alpha \\
\implies & l_2^2 = l_1^2 + y_e^2 + z_e^2 - 2 l_1 \sqrt{y_e^2 + z_e^2} \cos \alpha \\
\implies & \alpha = \cos^{-1} \left( \frac{l_1^2 + y_e^2 + z_e^2 - l_2^2}{2 l_1 \sqrt{y_e^2 + z_e^2}} \right).
\end{align}

Meanwhile, according to the coordinate of end point $e$, {\ie}, $(0, y_e, z_e)$, we can get that
\begin{align}
& \tan2 (\alpha + \theta) = (z_e, y_e)\\
\implies & (\alpha + \theta) = atan2(z_e, y_e).
\end{align}

Based on the above analysis, we can easily obtain the representations of angles $\theta$ and $\alpha$ as follows:
\begin{align}
\theta &= (\alpha + \theta) - \alpha = atan2(z_e, y_e) - \cos^{-1} \left( \frac{l_1^2 + y_e^2 + z_e^2 - l_2^2}{2 l_1 \sqrt{y_e^2 + z_e^2}} \right),\\
\phi &= \pi - \beta = \pi - \cos^{-1} \left( \frac{l_1^2 + l_2^2 - y_e^2 - z_e^2}{2 l_1 l_2} \right).
\end{align}

According to the above analysis, via the \textit{analytic method} shown in the example, we can obtain the closed-form solution of the \textit{inverse kinematics} problem. The readers may have also notice that the \textit{analytic method} calculation process is already very cumbersome for the 2-link robot arm. When it comes to more complicated robot structures, like a 6-link robot arm, it will become much more challenging to solve the problem analytically.

\subsection{Inverse Position and Orientation Kinematics}

Compared with the \textit{inverse kinematics}, we have noticed that the \textit{forward kinematics} calculation process we introduce in the previous section is much simpler and easier. Readers may also wonder if we can use \textit{forward kinematics} introduced before to help identify the solution to the \textit{inverse kinematics} problem. This is also the problem we plan to study in this part, and we will introduce a \textit{numerical solution} that doesn't need to analyze the 


\subsubsection{Numerical Solution Method}

Before we introduce the detailed information about the \textit{numerical solution}, we would like to first provide the general framework of the \textit{numerical solution} algorithm.
\begin{itemize}
\item \textbf{Step 1}: Define the position vector $\mb{p}_{start}$ and rotation matrix $\mb{R}_{start}$ of the robot's start position and orientation. 
\item \textbf{Step 2}: Define the position vector $\mb{p}_{target}$ and rotation matrix $\mb{R}_{target}$ of the robot's target position and orientation. 
\item \textbf{Step 3}: Initialize a variable vector $\mb{q}$ that denotes the angles of the robot's link angles.
\item \textbf{Step 4}: Given the variable vector $\mb{q}$ and the current robot position and orientation, calculate the robot position vector $\mb{p}$ and orientation matrix $\mb{R}$ outputs with \textit{forward kinematics}.
\item \textbf{Step 5}: Calculate the introduced error (or loss) term of the \textit{forward kinematics} outputs $\mb{p}$, $\mb{R}$ compared with the target outputs $\mb{p}_{target}$, $\mb{R}_{target}$, {\ie}, $\delta \mb{p} = \mb{p} - \mb{p}_{target}$ and $\delta \mb{R} = \mb{R} - \mb{R}_{target}$.
\item \textbf{Step 6}: If $\delta \mb{p}$ and $\delta \mb{R}$ are sufficiently small (less than the pre-defined threshold), stop the calculation and return the current variable vector $\mb{q}$ as the output.
\item \textbf{Step 7}: Otherwise, calculate the term $\delta \mb{q}$ with the introduced error terms $\delta \mb{p}$ and $\delta \mb{R}$, and update the current variable vector $\mb{q} = \mb{q} + \delta \mb{q}$ and go to \textbf{Step 4}.
\end{itemize}

\begin{figure}[t]
    \centering
    \includegraphics[width=0.9\textwidth]{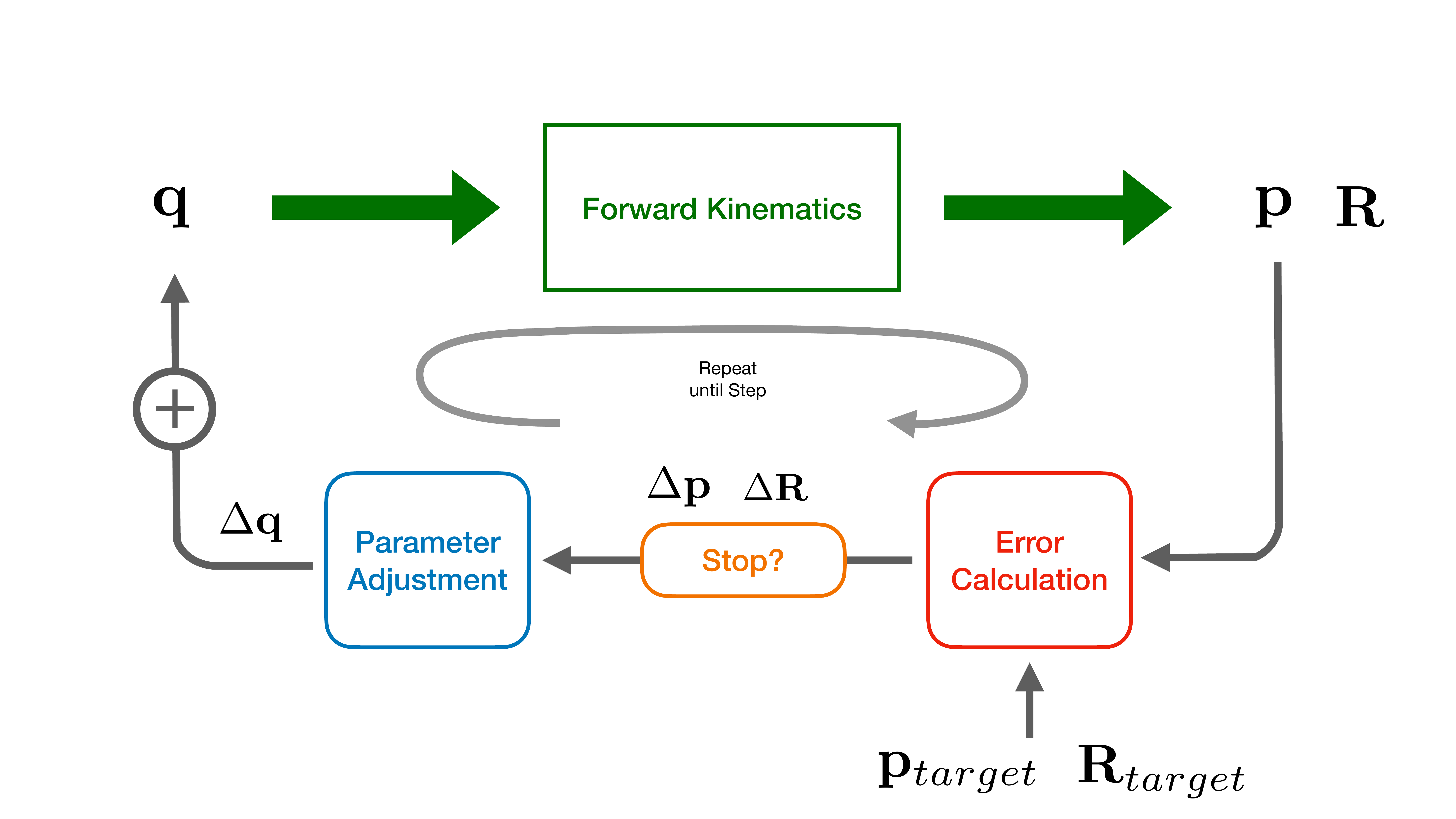}
    \caption{Outline of the Numerical Solution in Updating Variable Vector via Forward Kinematics.}
    \label{fig:numerical_algorithm_outline}
\end{figure}

An outline of the above algorithm is also illustrated in Figure~\ref{fig:numerical_algorithm_outline}. Such an iterative adjustment process continues until the stop criteria is met. Meanwhile, the above algorithm also create several new open questions:
\begin{enumerate}
\item \textbf{Error Function}: How to define ``whether $\delta \mb{p}$ and $\delta \mb{R}$ are sufficiently small''?
\item \textbf{Adjustment Term}: How to calculate the ``$\delta \mb{q}$'' term to be used for updating $\mb{q}$?
\end{enumerate}

To address the first problem, a loss function is introduced to quantify ``whether $\delta \mb{p}$ and $\delta \mb{R}$ are sufficiently small'' or not. Formally, given $\delta \mb{p} = \mb{p} - \mb{p}_{target}$ and $\delta \mb{R} = \mb{R} - \mb{R}_{target}$, we can represent the introduced loss term to be
\begin{equation}
\ell (\delta \mb{p}, \delta \mb{R}) = \left\| \delta \mb{p} \right\|^2 + \left\| \left( \ln \delta \mb{R} \right)^{\lor} \right\|^2
\end{equation}
According to the previous tutorial article, the ``$\ln$'' and ``$\lor$'' operators will project the \textit{rotation matrix} to the corresponding \textit{angular velocity} vector, {\ie}, $\delta (\bs{\omega} t) = (\ln \delta \mb{R})^{\lor}$. The term $\bs{\omega}$ denotes the \textit{angular velocity} vector of the interested point of the robot body. For the loss term $\ell (\delta \mb{p}, \delta \mb{R}) \le \gamma$ (where $\gamma = 1e^{-6}$ is a pre-defined threshold parameter), then the updating process will stop and we can also generally guarantee that the robot position vector and orientation matrix approaches to the targets.

As to the second problem, we will introduce its solution in the following subsection for readers with the \textit{Newton-Raphson Method}.


\subsubsection{Newton-Raphson Method}\label{subsec:newton_raphson_method}

Let's assume we change the parameter vector $\mb{q}$ with a very minor term $\delta \mb{q}$, where $\delta \mb{q} \to \mb{0}$. Its impact on the position vector $\mb{q}$ and orientation matrix $\mb{R}$ can be represented as
\begin{align}
\delta \mb{p} &= g_{\mb{p}}(\mb{q} + \delta \mb{q}) - g_p(\mb{q}),\\
\delta (\bs{\omega}t) &= g_{\bs{\omega}}(\mb{q} + \delta \mb{q}) - g_{\bs{\omega}}(\mb{q}).
\end{align}
In the above equation, we already replace the term $\delta \mb{R}$ with the $\delta \bs{\omega}$ just to simplify the representations below.

As introduced before in Equation~\ref{equ:jacobian_matrix} on \textit{velocity kinematics}, if the term $\delta \mb{q}$ is small enough and approaches to $\mb{0}$, we can actually combine the above two equations and rewrite them as follows:
\begin{equation}\label{equ:inverse_velocity_equation}
\begin{bmatrix}
\delta {\mb{p}}\\
\delta ({\bs{\omega}}t)\\
\end{bmatrix} = \begin{bmatrix}
\mb{v}\\
\bs{\omega}\\
\end{bmatrix} = \mb{J} \delta {\mb{q}},
\end{equation}
where $\mb{J} \in \mathbbm{R}^{6 \times n}$ is the \textit{Jacobian matrix} we introduce before and vectors $\mb{v} = \dot{\mb{p}}$ and $\bs{\omega}$ denote the \textit{linear} and \textit{angular velocity}, respectively. The notation $n$ denotes the number of links in the robot, which is equal to the dimension of the variable vector $\mb{q}$. The combined vector $\bs{\nu} = \begin{bmatrix}
\mb{v}\\
\bs{\omega}\\
\end{bmatrix}$ is also the \textit{twist velocity} vector we define before.

If matrix $\mb{J}$ is invertible, we can represent the term $\delta \mb{q}$ as follows:
\begin{equation}
\delta {\mb{q}} = \mb{J}^{-1} \begin{bmatrix}
\mb{v}\\
\bs{\omega}\\
\end{bmatrix} = \mb{J}^{-1} \bs{\nu}.
\end{equation}

The above equation answer the second problem on ``\textit{how to calculate the `$\delta \mb{q}$' term to be used for updating $\mb{q}$}''. In application, instead of directly applying the calculated term $\delta {\mb{q}}$ to update the variable vector $\mb{q}$, we usually also includes a parameter $\lambda \in (0, 1)$ to control the variable updating speed, {\ie},
\begin{equation}
\mb{q} = \mb{q}  + \lambda \cdot \delta {\mb{q}}.
\end{equation}
The introduced parameter $\lambda$ can stabilize the updating process and avoid drastic changes to the variables, and this method is formally named as the \textit{Newton-Raphson method}.

If the readers are familiar with \textit{mathematical optimization} or \textit{gradient descent} algorithms, the \textit{Newton-Raphson method} has been widely applied in parameter updating to optimize the model performance. Meanwhile, in the above calculation process, we have a strong assumption that ``the \textit{Jacobian matrix} $\mb{J}$ is invertible''. At the end of this section, we will have a detailed analysis about the \textit{Jacobian matrix} and discuss about how to update the variables when $\mb{J}$ is \textit{singular}.

\subsection{Inverse Velocity and Torque Kinematics}

In the previous Section~\ref{subsec:velocity_kinematics} and Section~\ref{subsec:torque_kinematics}, we discuss about the robot \textit{forward velocity kinematics} and \textit{forward torque kinematics}, which calculates the robot part \textit{twist velocity} vector and generated \textit{force} vector based on the inputs. In this part, we will talk about the reversed process, {\ie}, the \textit{inverse velocity kinematics} and \textit{inverse torque kinematics}. 

\subsubsection{Inverse Velocity Kinematics}

From the name, we can know that \textit{inverse velocity kinematics} literally denotes the process to figure out the desired joint \textit{angular velocity} to achieve the desired \textit{twist velocity} vector. From Equation~\ref{equ:jacobian_matrix} we provide in the previous section on defining the \textit{Jacobian matrix}, we already know the relationship between \textit{twist velocity} vector and the joint \textit{angular velocity}. Furthermore, from Equation~\ref{equ:inverse_velocity_equation} in the previous subsection, we also know that if the \textit{Jacobian matrix} $\mb{J}$ is invertible, the joint \textit{angular velocity} can be calculated from its \textit{twist velocity} as follows:
\begin{equation}
\dot{\mb{q}} = \mb{J}^{-1} \bs{\nu}.
\end{equation}
Depending on the coordinate chosen to study the \textit{inverse velocity kinematics}, the \textit{twist velocity} vector $\bs{\nu}$ can be defined as either $\mb{T}^{-1}\dot{\mb{T}}$ (the twist velocity in the local coordinate system) or $\dot{\mb{T}} \mb{T}^{-1}$ (the twist velocity within the world coordinate system) as we introduce before. In the case if the \textit{Jacobian matrix} $\mb{J}$ is not invertible, we will introduce the solution in the following part at the end of this section for readers.

\subsubsection{Inverse Torque Kinematics}

In the previous Section~\ref{subsec:torque_kinematics}, we have also discussed about the relationship between the robot \textit{torque} and the \textit{force}. In Equation~\ref{equ:force_torque_relationship} introduced before, via the \textit{Jacobian Matrix}, we have already calculate the required joint \textit{torque} to generate the desired force for the end point, {\ie},
\begin{equation}
\bs{\tau}_e = \mb{J}^\top \mb{f}_e.
\end{equation}

The above equation is derived based on pure static analysis, and no kinetic energy or potential energy are considered actually. In the following section on \textit{robot dynamics}, we will further discuss about the robot \textit{velocity}, \textit{torque} and \textit{force} for readers.

\subsection{Singular Jacobian Matrix}

In this part, we will talk about an important problem about the Jacobian matrix. According to the previous discussion, the Jacobian matrix plays an important role in calculating (1) \textit{twist velocity} vector and \textit{force} vector of the end point in \textit{forward kinematics}, (2) needed joint \textit{angle} vector and \textit{torque} vector to generate the \textit{twist velocity} and \textit{force} in \textit{inverse kinematics}. We make lots of assumption on the Jacobian matrix before in the derivation, {\eg}, it is full rank and invertible. In this part, we will discuss about some potential problems with Jacobian matrix, {\ie}, the \textit{singularity} problem.

\subsubsection{Singularity and Gimbal Lock}

We can first use the classic ``gimbal lock'' problem to describe ``what is singularity''. The ``gimbal lock'' term becomes famous in the movie Apollo 13, which happens when the rotational axis of the middle term in the sequence becomes parallel to the rotational axis of the first or the third term.

\begin{figure}[t]
    \centering
    \includegraphics[width=0.9\textwidth]{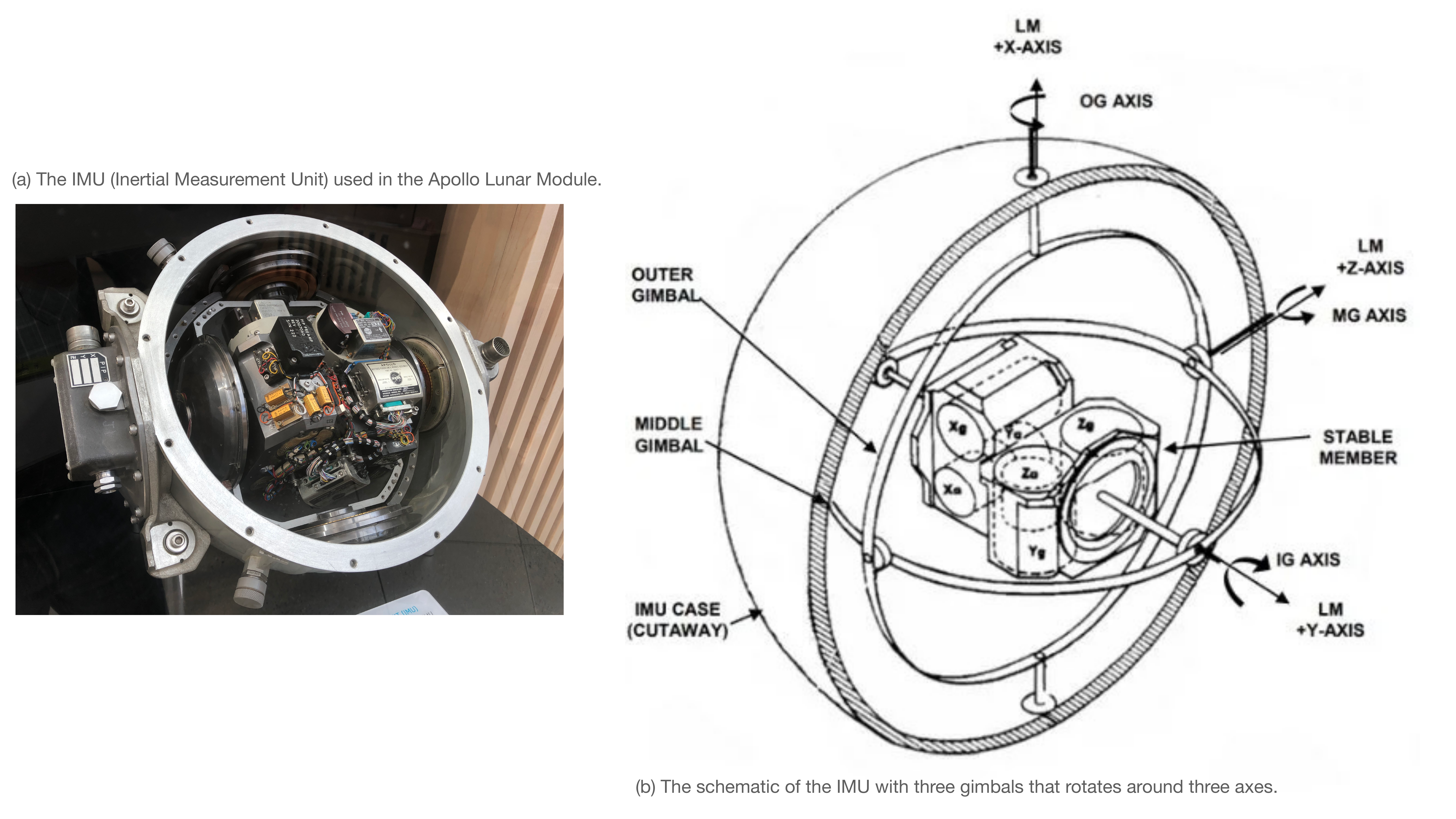}
    \caption{An Illustration of the IMU (Inertial Measure Unit) used in Apollo Lunar Module.}
    \label{fig:inertial_measurement_unit_illustration}
\end{figure}

\begin{example}
As illustrated in Figure~\ref{fig:inertial_measurement_unit_illustration}, we show the IMU (inertial measurement unit) used in the Apollo Lunar Module and its schematic representation, which is a mechanical gyroscope used for spacecraft navigation. This IMU has three orthogonal gyroscopes that hold it at a constant orientation with respect to the universe. According to the orientations of these three axes and the right-hand rule, we can mark them as the $x$, $y$ and $z$ axis, respectively. 

\begin{figure}[t]
    \centering
    \includegraphics[width=0.9\textwidth]{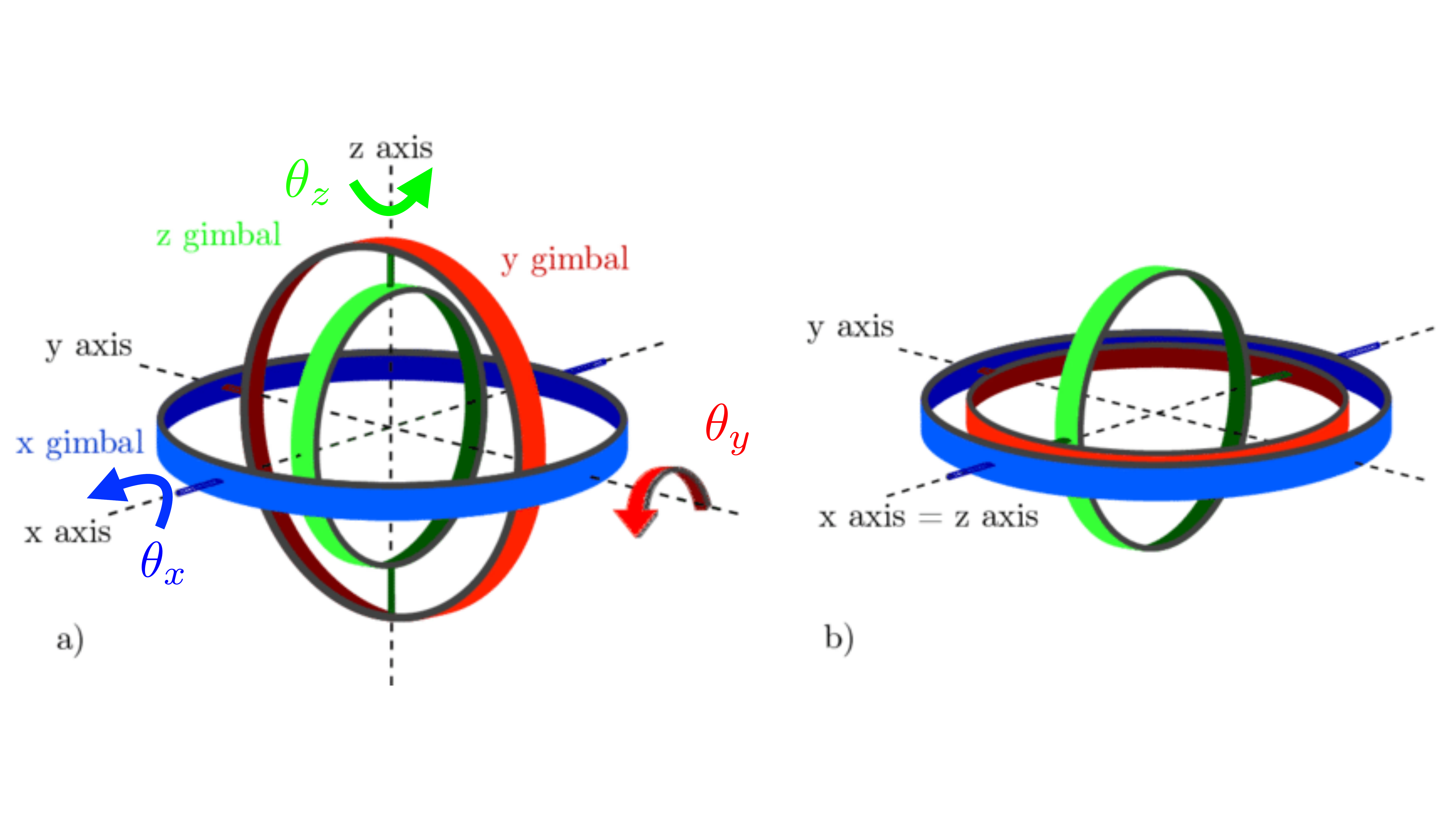}
    \caption{An Illustration of the IMU Gimbal Lock Problem.}
    \label{fig:gimbal_lock_problem}
\end{figure}

As the spacecraft pose changes in the universe, the stable member of the IMU will remain relatively stable in terms of the orientation, but the outer, middle and inner gimbal will rotate around their axes with certain angles. As shown in Figure~\ref{fig:gimbal_lock_problem}, we can represent the rotation angles around these three axes as $\theta_x$, $\theta_y$ and $\theta_z$, respectively. From these rotation angles, we will be able to infer the orientation of the spacecraft.

Meanwhile, when the middle gimbal rotates around the $y$ axis with an angle $\theta_y = \pi/2$, we observe that the axis of the inner and outer gimbal ({\ie}, the $x$ axis and $z$ axis) will be aligned and share the same rotation axis. For such a case, the IMU will have only two degrees of freedom and one degree of freedom will be lost. 
\end{example}

Such a degree of freedom lose in the above example can also be illustrated via the mathematical representations of the rotation matrix about the IMU. For the rotation around the $x$, $y$ and $z$ axis with angles of $\theta_x$, $\theta_y$ and $\theta_z$ degrees, we can represent the corresponding rotation matrix of the IMU to be
\begin{equation}
\mb{R} = \mb{R}_x(\theta_x) \mb{R}_y(\theta_y) \mb{R}_z(\theta_z).
\end{equation}
Such rotation matrix follows the cyclic rotation rules, {\ie},
\begin{align}
&\mb{R}_x(\pi/2) \mb{R}_y(\theta) \mb{R}_x(\pi/2)^\top = \mb{R}_z(\theta),\\
&\mb{R}_y(\pi/2) \mb{R}_z(\theta) \mb{R}_y(\pi/2)^\top = \mb{R}_x(\theta),\\
&\mb{R}_z(\pi/2) \mb{R}_x(\theta) \mb{R}_z(\pi/2)^\top = \mb{R}_y(\theta);
\end{align}
as well as the anti-cyclic rotation rules, {\ie},
\begin{align}
&\mb{R}_y(\pi/2)^\top \mb{R}_x(\theta) \mb{R}_y(\pi/2) = \mb{R}_z(\theta),\\
&\mb{R}_z(\pi/2)^\top \mb{R}_y(\theta) \mb{R}_z(\pi/2) = \mb{R}_x(\theta),\\
&\mb{R}_x(\pi/2)^\top \mb{R}_z(\theta) \mb{R}_x(\pi/2) = \mb{R}_y(\theta).
\end{align}
We will not provide the proofs of the above rules, and readers can easily prove them with the concrete representations of rotation matrices $\mb{R}_x$, $\mb{R}_y$ and $\mb{R}_z$ to calculate their product result representations by yourselves.

According to the above rules as well as the \textit{orthogonality} of rotation matrix, when the middle gimbal rotation angle $\theta_y = \pi/2$, we can rewrite the rotation matrix of the IMU as follows:
\begin{align}
\mb{R} &= \mb{R}_x(\theta_x) \mb{R}_y(\pi/2) \mb{R}_z(\theta_z) \\
&= \mb{R}_x(\theta_x) \mb{R}_y(\pi/2) \mb{R}_z(\theta_z) \left( \mb{R}_y(\pi/2)^\top \mb{R}_y(\pi/2) \right) \\
&= \mb{R}_x(\theta_x) \left( \mb{R}_y(\pi/2) \mb{R}_z(\theta_z) \mb{R}_y(\pi/2)^\top \right) \mb{R}_y(\pi/2) \\
&= \mb{R}_x(\theta_x) \mb{R}_x(\theta_z) \mb{R}_y(\pi/2) \\
&= \mb{R}_x(\theta_x + \theta_z) \mb{R}_y(\pi/2).
\end{align} 
The above representation of the rotation matrix also indicates that the IMU will not be able to represent the rotation in the $z$ axis anymore and render the IMU fail to work. Such a problem happens only at certain special circumstances, {\eg}, $\theta_y = \pi/2$ as show above.

\subsubsection{Jacobian Singularity}

A robot singularity is a physical blockage, not some kind of abstract mathematical problem. 

\begin{definition}
(\textbf{Robot Singularity}): At a \textit{singularity}, a robot loses one or more degrees of freedom.
\end{definition}

In the previous subsection, we have illustrate the \textit{singularity} of a simple ``robot'', {\ie}, the IMU with three degrees of freedom, which will lose one degree of freedom when the middle gimbal rotates with an angle of $\theta_y = \pi/2$ degrees. 

Besides the simple IMU, \textit{singularity} widely exists in many robots, like a robot arm or a legged robot. Readers may also wonder what are the causes of the \textit{robot singularity}. In this part, we will illustrate that \textit{robot singularity} is closely related to the singularity of the corresponding \textit{Jacobian matrix}.

\begin{example}

To illustrate the relationship of robot singularity with the \textit{Jacobian matrix}, we provide an example of robot arm in Figure~\ref{fig:robot_singularity_Jacobian_matrix}, whose links are connected via the $6$ joints, and the DH parameters are illustrated in the figure as well. 

\vspace{5pt}
\noindent \textbf{A Note}: We haven't introduced the robot DH parameters for readers yet. For now, you just need to know robot DH parameters can describe the pose and structure of rigid-body robots. Once the DH parameters are specified, then we can build the mathematical model to recover the current pose and the full structure of the robot. More detailed information about DH parameter will be described in detail in the follow-up tutorial articles instead. 
\vspace{5pt}

In the DH parameter representation, the local coordinates are attached to the links on the far-away side of the joints with the $z$ axis collinear with the joint rotation axis. For some of the axes not clearly indicated, readers can infer their orientation with the right-hand rule, {\ie}, thumb: $x$ axis, index finger: $y$ axis, and middle finger: $z$ axis.

\begin{figure}[t]
    \centering
    \includegraphics[width=0.9\textwidth]{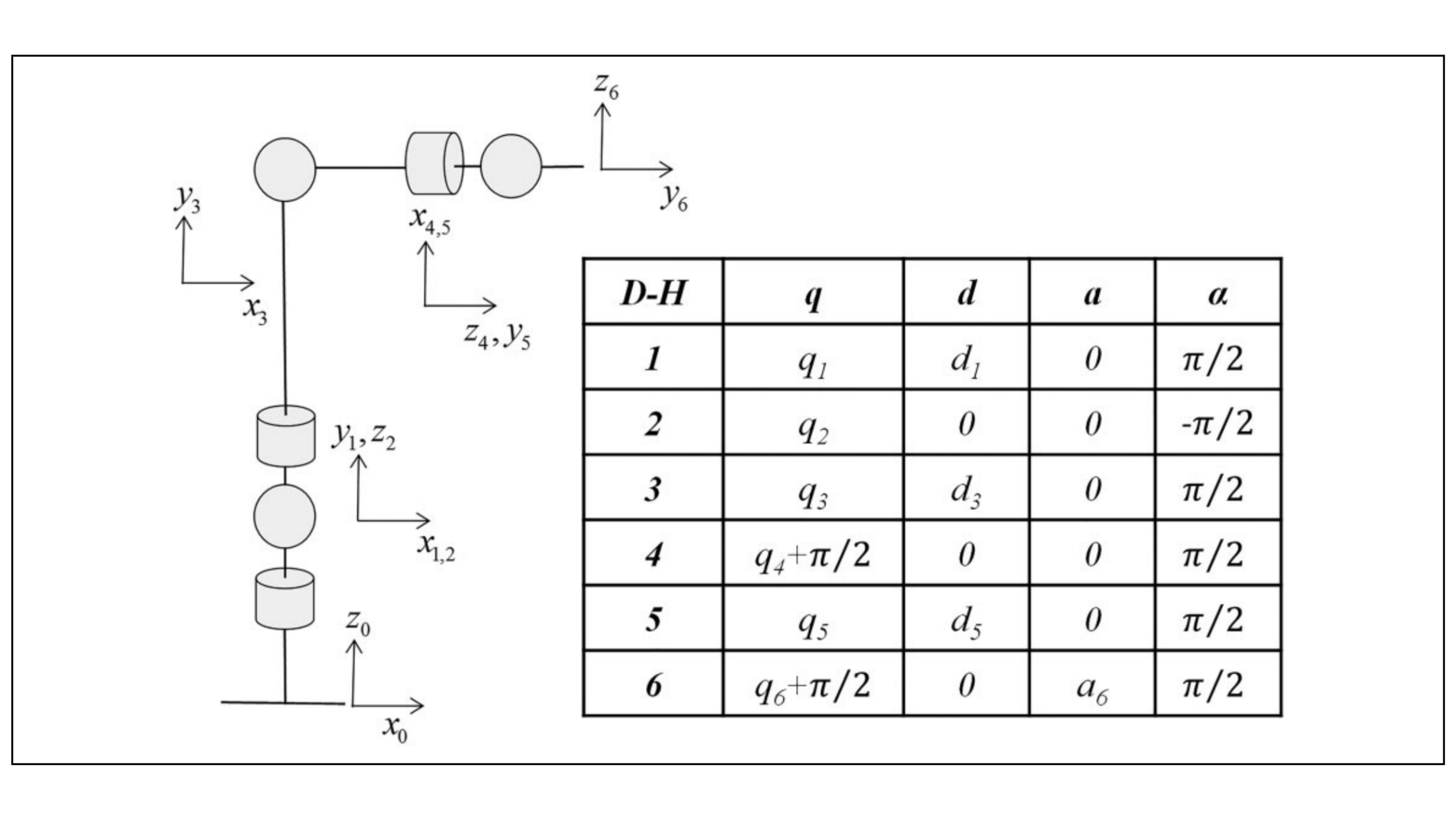}
    \caption{An Example of $6$-DOF Robot Arm and Its DH Parameter Table. (DOF: Degree of Freedom).}
    \label{fig:robot_singularity_Jacobian_matrix}
\end{figure}

According to the DH table, we can define its corresponding Jacobian matrix as follows:
\begin{equation}
\mb{J} = \begin{bmatrix}
X & X & X & X & X & X\\
X & X & X & X & X & X\\
0 & X & X & X & X & X\\
0 & s_1 & -c_1s_2 & -c_3s_1-c_{123} & c_1s_{24} - c_4s_{13} + c_{1234} & X\\
0 & -c_1 & -s_{12} & c_{13} -c_2s_{13} & s_{124}+c_{14}s_3+c_{234}s_1 & X\\
1 & 0 & c_2 & -s_{23} & c_{34}s_2-c_2s_4 & c_{24}s_5 -c_5s_{23}+c_3s_{245}\\
\end{bmatrix},
\end{equation}
where the simplified notations of terms $c_i$ and $s_i$ denote $\cos q_i$ and $\sin q_i$. Meanwhile, terms $s_{ijk\cdots}$ and $c_{ijk\cdots}$ represent $\sin(q_i + q_j + q_k + \cdots)$ and $\cos(q_i + q_j + q_k + \cdots)$ instead.

Via symbolic computation, we can calculate the determinant of the Jacobian matrix as 
\begin{equation}
\det (\mb{J}) = - \cos(q_5) \sin(q_2) \Gamma(q_4) \text{, where } \Gamma(q_4) = (d_3^2d_5)/2+d_3^2d_5\cos(2q_4)/2.
\end{equation}

In the derivation of \textit{inverse kinematics} we introduce before, the inverse of the Jacobian matrix play a critical role. We assumed the Jacobian matrix is invertible, whose inverse will help calculate the desired joint angles, angular velocity and torque vector parameters. Meanwhile, the Jacobian matrix is invertible when Jacobian is not singular, and a matrix is singular iff its determinant is $0$. Therefore, we can derive the \textit{singularity} condition to be
\begin{equation}
\det (\mb{J}) = 0 \Longleftrightarrow \begin{cases}
&\text{Condition 1}: \cos(q_5) = 0\\
&\text{Condition 2}: \sin(q_2) = 0\\
&\text{Condition 3}: \Gamma(q_4) = 0\\
\end{cases}
\Longleftrightarrow \begin{cases}
&q_5 = \pm \pi/2\\
&q_2 = 0\\
&q_4 = - \pi/2\\
\end{cases},
\end{equation}
where assume the joint angles are all within the range $[-\pi/2, \pi/2]$ and some cases like $q_2 = \pi$ have been pruned from the results.

Based on the above solutions of the \textit{singularity} conditions, as shown in Figure~\ref{fig:robot_singular_pose}, we can categorize the robot singularity into two main types:
\begin{itemize}

\item \textbf{Boundary Singularity}: The condition that $q_4 = - \pi/2$ will cause the robot arm manipulator in a straight pose, as shown in Figure~\ref{fig:robot_singular_pose}, where the end point of the arm will actually reach the boundary of its workspace. Therefore, we also call such singularity as the \textit{boundary singularity} or \textit{workspace singularity}. It is usually caused by a full extension of a joint, and asking the manipulator to move beyond where it can be positioned. An illustration of robot arm \textit{boundary singularity} is also shown in Figure~\ref{fig:robot_boundary_singular_region}. 
\begin{figure}[H]
    \centering
    \includegraphics[width=0.7\textwidth]{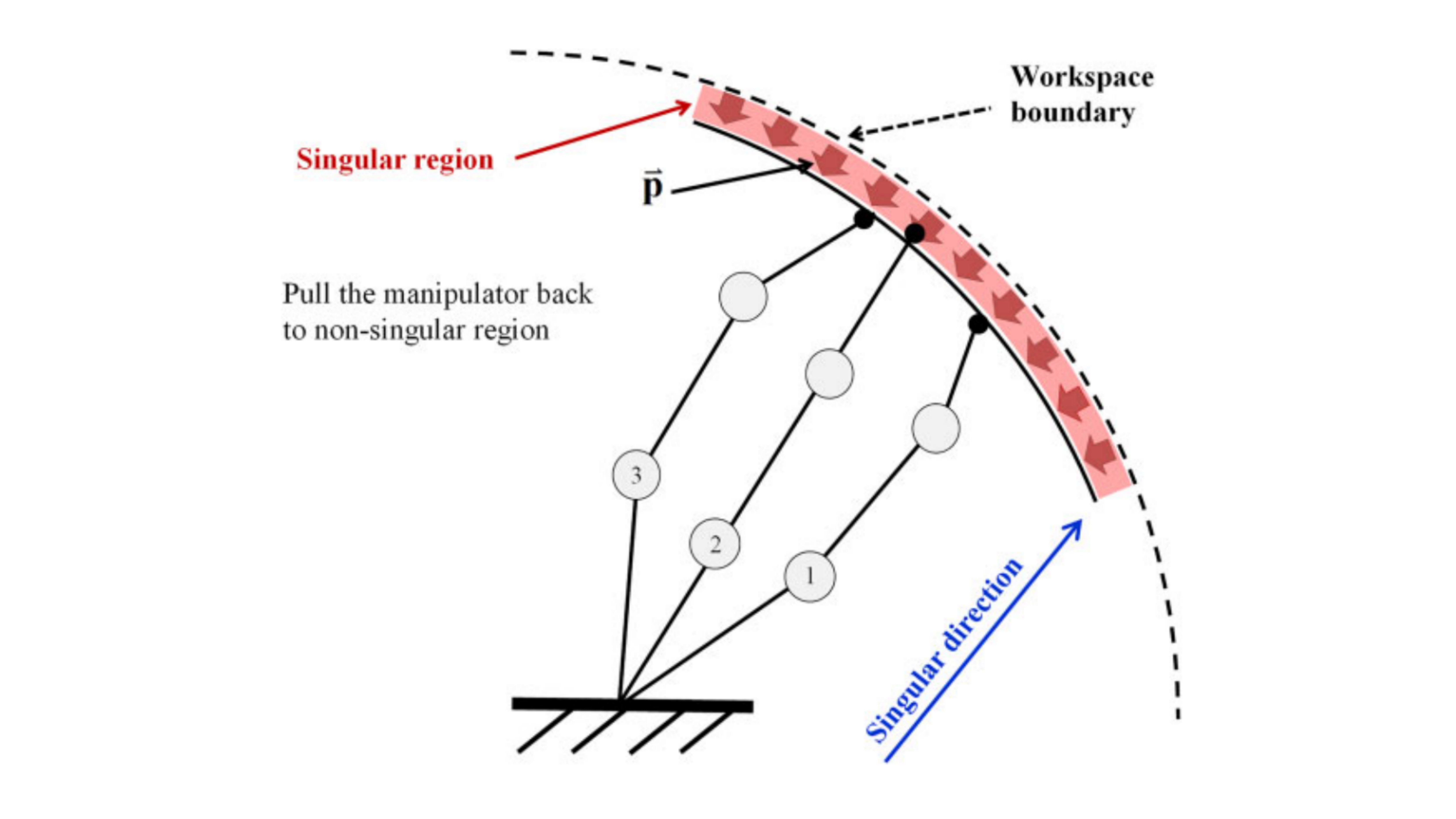}
    \caption{An Illustration of Robot Arm Boundary Singularity and Singular Region.}
    \label{fig:robot_boundary_singular_region}
\end{figure}

\item \textbf{Internal Singularity}: Except the condition $q_4 = - \pi/2$, the other conditions do not yield a straight posture, and we name them as the \textit{internal singularity}. The \textit{internal singularity} (also known as \textit{joint space singularity}) are generally caused by an alignment of the robots axes in space. The conditions derived above can introduce three different \textit{internal singularity} as follows:
\begin{itemize}
\item \textbf{Case 1}: $q_2 = 0$, $q_4 \neq - \pi/2$ and $q_5 \neq \pm \pi/2$. When $q_2 = 0$, the axes of joints $q_1$ and $q_3$ will be collinear so the rotation of these two joints correspond to the directions of the coordinate system is coupled in the $y_1$-direction and the $z_1$-direction as illustrated in Figure~\ref{fig:robot_boundary_singular_region}. 

\item \textbf{Case 2}: $q_2 \neq 0$, $q_4 \neq - \pi/2$ and $q_5 = \pm \pi/2$. The axes joints $q_1$ and $q_6$ (blue color), as well as those of joints $q_3$ and $1_6$ (red color) will be collinear. 

\item \textbf{Case 3}: $q_2 = 0$, $q_4 \neq - \pi/2$ and $q_5 = \pm \pi/2$. The collinear joint axes will include $q_1$, $q_3$, and $q_6$, or $q_1$ and $q_3$.

\end{itemize}
\end{itemize}

\begin{figure}[t]
    \centering
    \includegraphics[width=0.9\textwidth]{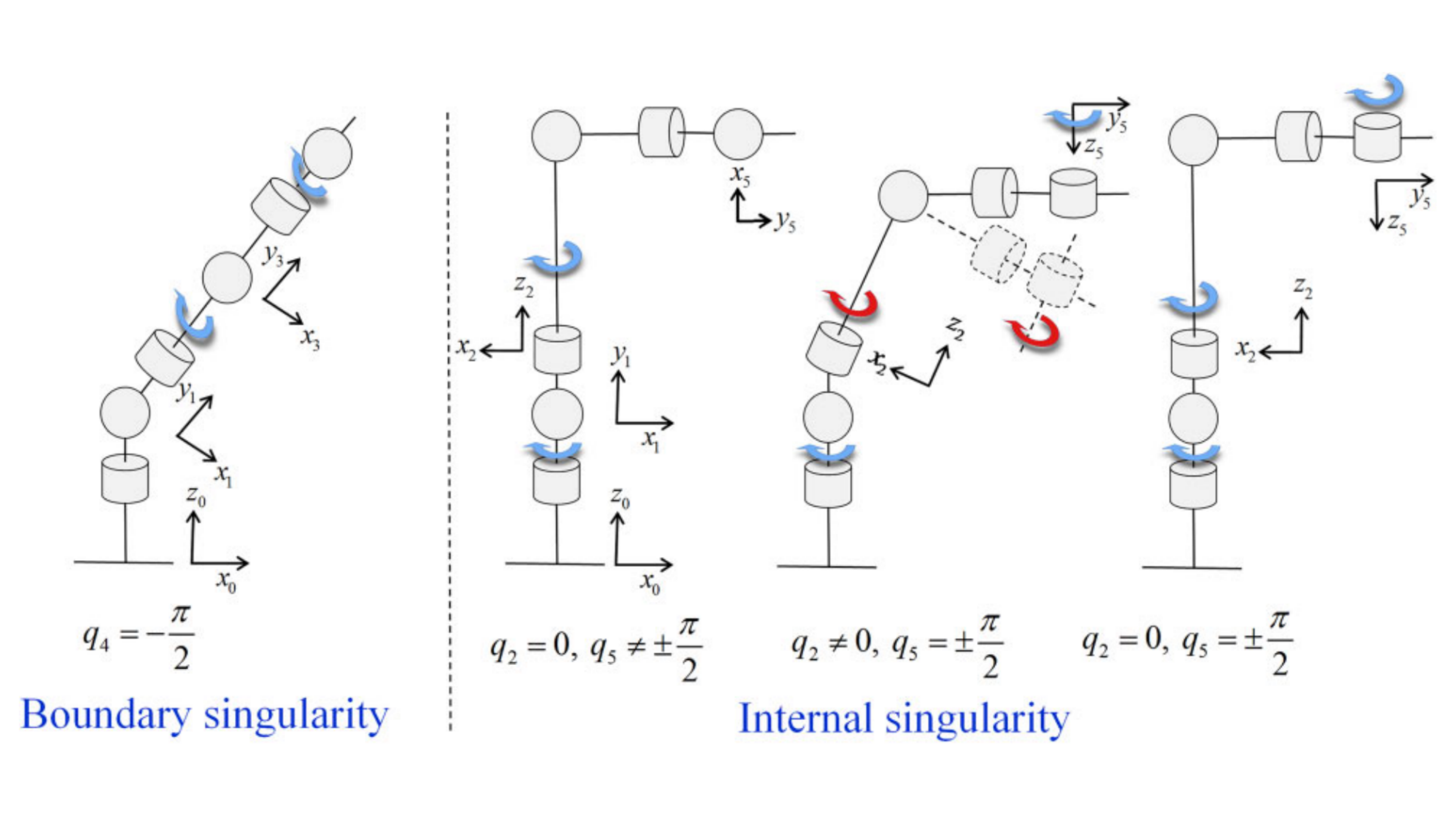}
    \caption{An Illustration of Robot Arm Boundary Singularity and Internal Singularity.}
    \label{fig:robot_singular_pose}
\end{figure}
\end{example}

\vspace{5pt}
\noindent \textbf{A Remark}: The above derivation about the \textit{singularity conditions} are obtained based on the robot arm shown in Figure~\ref{fig:robot_singularity_Jacobian_matrix}. For other robot arms with different structures, their DH parameter table will be different, and we will also obtain very different derivation results for the \textit{singularity conditions} as well.
\vspace{5pt}

\subsubsection{Robot Singular Pose in Application}

\begin{figure}[t]
    \centering
    \includegraphics[width=0.9\textwidth]{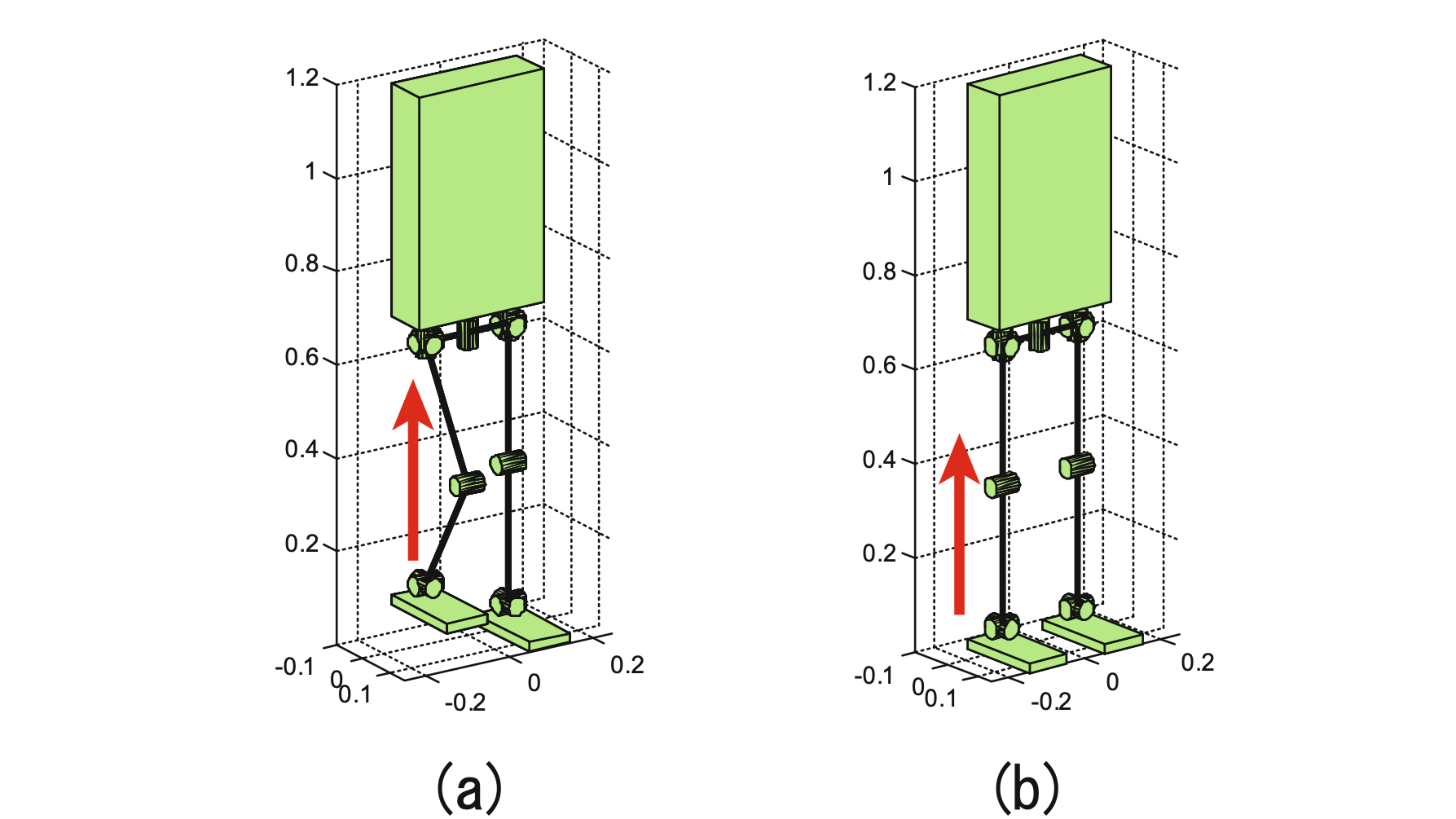}
    \caption{Velocity Inverse Kinematics of Legged Robot \cite{10.5555/3100040}.}
    \label{fig:legged_robot}
\end{figure}

Besides the robot arm, \textit{singularity} also exist in other types of robots in real-world applications, like the \textit{legged robots}. 

\begin{example}
As shown in Figure~\ref{fig:legged_robot}, we show a biped legged robot, where each leg has 6 DOF and the trunk is just simply represented as a cuboid. In Plot (a), the robot right leg is slightly raised and the angles of each joint are $(0, 0, -\pi/6, \pi/3, -\pi/6, 0)$, respectively. We aim to calculate the desired angular velocity of the robot leg joints so that the robot right foot ({\ie}, the end point) can lift vertically with a linear velocity of $0.1m/s$.

According to our previous discussion, we know that the robot joint angular velocity can be represented as
\begin{equation}
\dot{\mb{q}} = \mb{J}^{-1} \bs{\nu},
\end{equation}
where vector $\bs{\nu} = [0, 0, 0.1, 0, 0, 0]^\top$ denotes the \textit{twist velocity} vector of the right foot end point and the Jacobian matrix can be defined based on the current leg pose and its joint angles $(0, 0, -\pi/6, \pi/3, -\pi/6, 0)$. 

With the above equation, we can calculate the vector $\dot{\mb{q}}$ to be
\begin{equation}
\dot{\mb{q}} = \mb{J}^{-1} \begin{bmatrix}
0\\
0\\
0.1\\
0\\
0\\
0\\
\end{bmatrix} = \begin{bmatrix}
0\\
0\\
-0.3333\\
0.6667\\
-0.3333\\
0\\
\end{bmatrix}.
\end{equation}
In other words, the angular velocity of the hip, knee and ankle joint should be $-0.3333$ rad/s, $0.6667$ rad/s and $-0.3333$ rad/s, respectively.

Meanwhile, when it comes to the robot in the Plot (b) of Figure~\ref{fig:legged_robot}, we observe that its right leg is fully stretched and the angles of the current joints are all zeros. If we calculate the inverse of the corresponding Jacobian matrix and the angular velocity vector $\dot{\mb{q}}$ with the above equation, we can get
\begin{equation}
\mb{J}^{-1} = \begin{bmatrix}
\inf & \inf & \inf & \inf & \inf & \inf \\
\inf & \inf & \inf & \inf & \inf & \inf \\
\inf & \inf & \inf & \inf & \inf & \inf \\
\inf & \inf & \inf & \inf & \inf & \inf \\
\inf & \inf & \inf & \inf & \inf & \inf \\
\inf & \inf & \inf & \inf & \inf & \inf \\
\end{bmatrix} \text{, and } \dot{\mb{q}} = \begin{bmatrix}
\text{NaN}\\
\text{NaN}\\
\text{NaN}\\
- \inf \\
\inf \\
0\\
\end{bmatrix}.
\end{equation}
From the result, we observe that the robot right leg is at its singular pose. The corresponding Jacobian at the current pose is singular and its inverse doesn't exist. If we still use it to calculate the angular velocity vector $\dot{\mb{q}}$ to achieve a vertical linear velocity of $0.1$m/s, its result will consists of either NaN (Not a Number) of $\inf$ (Infinity). It is also the reason why the legged robots will always have their legs bended slightly to avoid be stretched too much and suffer from the singularity.
\end{example}

\begin{figure}[t]
    \centering
    \includegraphics[width=0.9\textwidth]{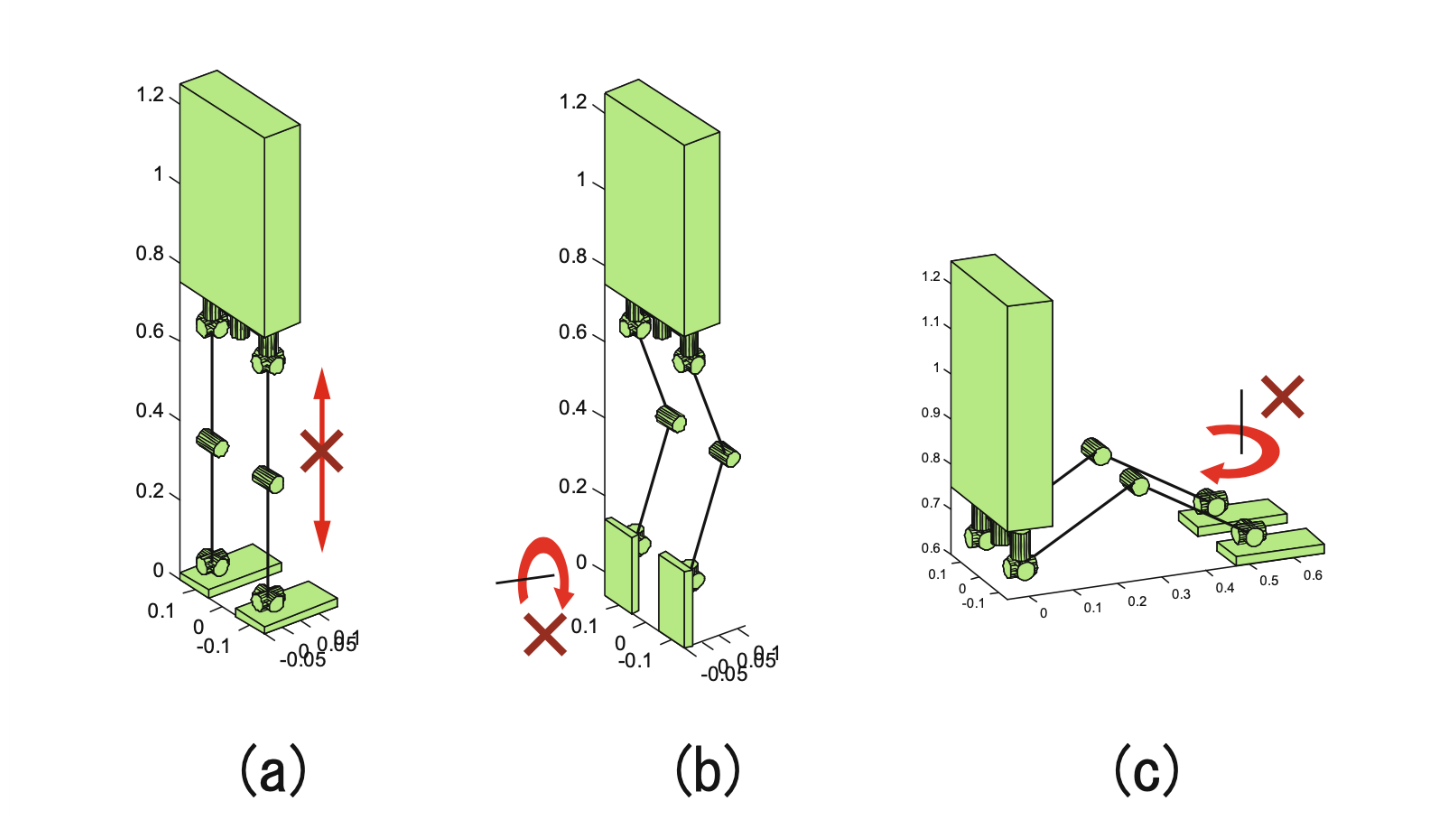}
    \caption{Legged Robot Singularity Poses \cite{10.5555/3100040}.}
    \label{fig:legged_robot_singularity}
\end{figure}

In addition to such a singular pose, in Figure~\ref{fig:legged_robot_singularity}, we also show several other singular poses of the legged robot.
\begin{itemize}
\item \textbf{Case (a)}: As discussed above, the legs are fully stretched and cannot move vertically due to singularity of the Jacobian matrix.
\item \textbf{Case (b)}: The hip yaw axis and the ankle roll axis are collinear, and the leg is also in a singular pose.
\item \textbf{Case (c)}: The robot hip roll axis and ankle roll axis are collinear and it is at a singular pose as well.
\end{itemize}

\subsection{Singularity-Robust Inverse of Jacobian}

By now, readers should know that the Jacobian matrix and its inverse plays an important role in robot inverse kinematics. When the Jacobian matrix is singular, we will suffer from lots of problems in calculating its inverse kinematics parameters. Meanwhile, in applications, people has almost zero tolerance for wired performance of the robots due to the singularity problem. At the end of this section, we will introduce a \textit{singularity robust inverse} of the Jacobian matrix, which allows the robots still be able to work with no worries about the singularity problem.

To illustrate the effectiveness, we also use an example in this part to compare the performance of the robot using conventional Jacobian inverse in Newton Raphson Method introduced in Section~\ref{subsec:newton_raphson_method} and the new method to calculate the \textit{singular robust inverse} of Jacobian matrix.

\begin{figure}[t]
    \centering
    \includegraphics[width=0.9\textwidth]{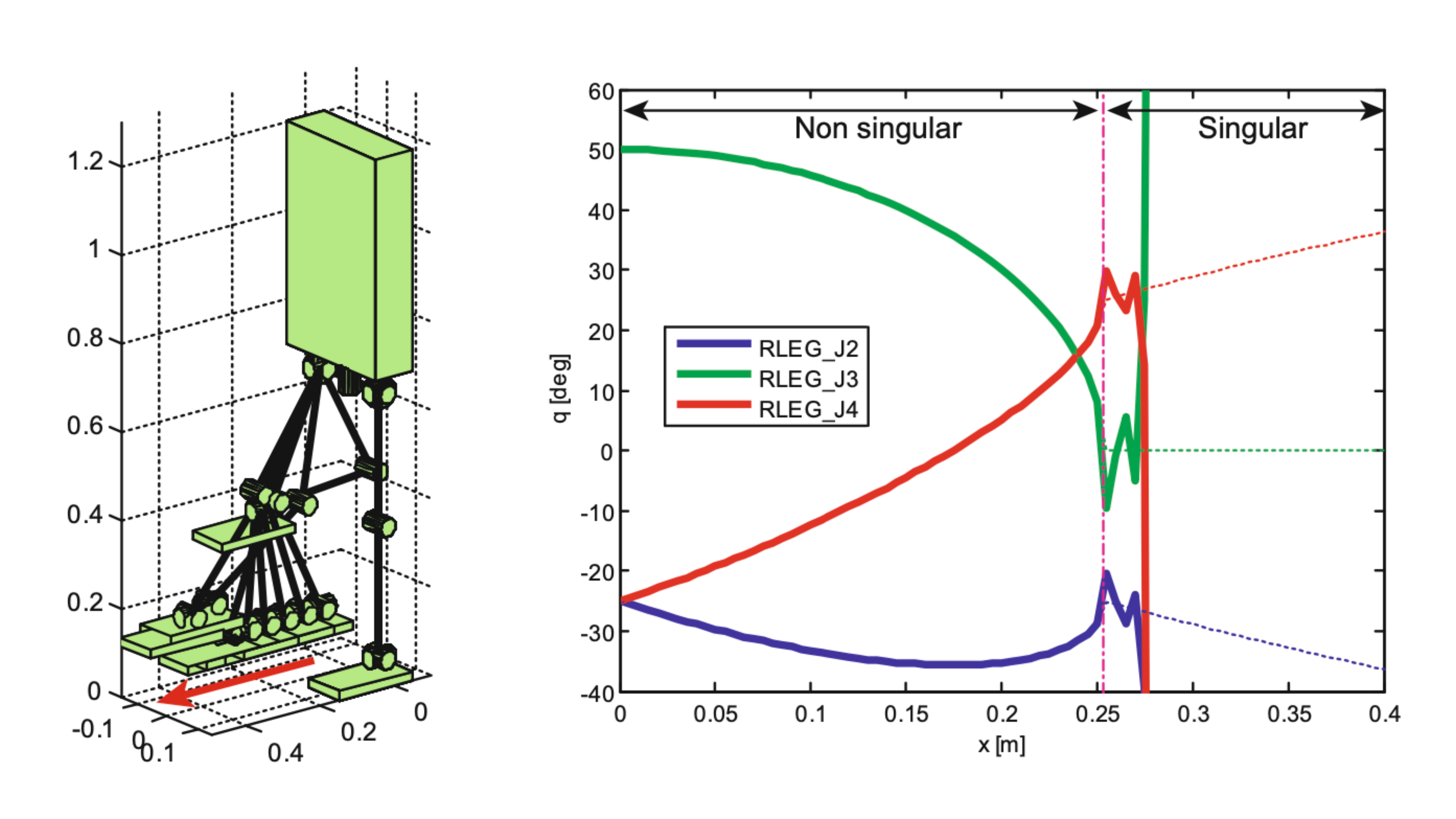}
    \caption{An Illustration of Legged Robot Suffering from Singularity \cite{10.5555/3100040}.}
    \label{fig:singularity_before}
\end{figure}

\begin{example}
As illustrated in Figure~\ref{fig:singularity_before}, we aim to calculate the robot joint angularity velocity to move the robot right foot forward from a non-singular pose with the Newton-Raphson method. The right plot shows the angles of the hip joint, keen joint and ankle joint at the given target foot position. The solid line denote the solution obtained by the numerical method. For comparison, we also provide the analytic solution in the plot, which is denoted by the dotted line on the right-hand side when the robot reaches its singular pose.

We observe that the robot will reach the singularity, since the axes of the hip pitch, knee pitch and ankle pitch will collinear. On the right plot, we observe that the angles of the hip, knee and ankle angles vibrate and go off the chart.
\end{example}

Let's take a look at the equation illustrating the relationship between the \textit{twist velocity} vector of the end point and the robot joint \textit{angular velocity}:
\begin{equation}
\bs{\nu} = \mb{J} \dot{\mb{q}}.
\end{equation}

We aim to find the desired vector $\dot{\mb{q}}$ to make the above equation hold. When the Jacobian matrix $\mb{J}$ is not singular, we can easily solve the problem with $\dot{\mb{q}} = \mb{J}^{-1}\bs{\nu}$. However, when matrix $\mb{J}$ is singular, we have to find another way to calculate the optimal $\dot{\mb{q}}$ vector.

From the above description, some readers probably already get a sense that ``can we solve the problem as an mathematical optimization problem?''. Finding the solution to equation $\bs{\nu} = \mb{J} \dot{\mb{q}}$ is actually equivalent to find the vector $\dot{\mb{q}}$ that can make the error vector $\mb{e}$ equal to zero, where
\begin{equation}
\mb{e} = \bs{\nu} - \mb{J} \dot{\mb{q}}.
\end{equation}

Making the error vector $\mb{e} = \mb{0}$ is not an easy task. Sometimes, we will not be able to find such a solution actually, which will render the robot failing to work. Then, can we slightly relax our expectation? Instead of find the error vector $\dot{\mb{q}}$ to make $\mb{e} = \mb{0}$, we propose to find the optimal vector $\dot{\mb{q}}$ to make the error vector $\mb{e}$ as close to the zero vector $\mb{0}$ as possible, {\ie},
\begin{equation}
\min_{\dot{\mb{q}}} \mb{E}(\dot{\mb{q}}),
\end{equation}
where we can introduce a loss function notation $\mb{E}(\dot{\mb{q}}) = \frac{1}{2} \mb{e}^\top \mb{e}$ to represent the optimization objective function. If the function $\mb{E}(\dot{\mb{q}})$ approaches $0$, the error vector $\mb{e}$ will be close to the zero vector as well, and vice versa.

The optimal variable $\dot{\mb{q}}$ can be obtained by making the derivative of $\mb{E}(\dot{\mb{q}})$ with respect to variable $\dot{\mb{q}}$ equal to zero, {\ie},
\begin{align}
&\frac{\partial \mb{E}(\dot{\mb{q}})}{\partial \dot{\mb{q}}} = \mb{0}\\
\Longrightarrow &\ \  - \mb{J}^\top \bs{\nu} + \mb{J}^\top \mb{J} \dot{\mb{q}} = \mb{0} \\
\Longrightarrow &\ \  \dot{\mb{q}} = \left( \mb{J}^\top \mb{J} \right)^{-1} \mb{J}^\top \bs{\nu}.
\end{align}
Meanwhile, the matrix $\mb{J}^\top \mb{J}$ is actually \textit{positive semidefinite}, and it can still be singular. Since $\det(\mb{J}^\top \mb{J}) = \det(\mb{J}^\top) \det(\mb{J})$, if the Jacobian matrix $\mb{J}$ is singular, {\ie}, $\det(\mb{J})=0$, we can easily get that $\det(\mb{J}^\top \mb{J}) = 0$, {\ie}, matrix product $\mb{J}^\top \mb{J}$ is singular as well. In other words, if $\mb{J}$ is singular, the above representation doesn't exist neither.

If the readers have any experiences with mathematical optimization, especially with variable regularizations, you probably can propose different definition of the loss function $\mb{E}(\dot{\mb{q}})$ definition. To address the above solution non-existence problem when $\mb{J}$ is singular, we propose to modify the loss function slightly by incorporating a regularization term on variable $\dot{\mb{q}}$ as follows:
\begin{equation}
\mb{E}(\dot{\mb{q}}) = \frac{1}{2} \mb{e}^\top \mb{e} + \frac{\lambda}{2} \dot{\mb{q}}^\top \dot{\mb{q}}.
\end{equation}
By solving the equation $\frac{\partial \mb{E}(\dot{\mb{q}})}{\partial \dot{\mb{q}}} = \mb{0}$, we can obtain the representation of the solution to be
\begin{align}
& \frac{\partial \mb{E}(\dot{\mb{q}})}{\partial \dot{\mb{q}}} = \mb{0}\\
\Longrightarrow &\ \ - \mb{J}^\top \bs{\nu} + \left( \mb{J}^\top \mb{J} + \lambda \mb{I} \right) \dot{\mb{q}} = \mb{0} \\
\Longrightarrow &\ \ \dot{\mb{q}} = \left( \mb{J}^\top \mb{J} + \lambda \mb{I} \right)^{-1} \mb{J}^\top \bs{\nu}.
\end{align}
We know $\mb{J}^\top \mb{J}$ is \textit{positive semidefinite} and if $\lambda \neq 0$, their summation $\mb{J}^\top \mb{J} + \lambda \mb{I}$ will be \textit{positive definite} and its inverse always exists, which ensures we can find a solution $\dot{\mb{q}}$ no matter what the Jacobian matrix $\mb{J}$ is actually. 

We can introduce a new notation for the term $\left( \mb{J}^\top \mb{J} + \lambda \mb{I} \right)^{-1} \mb{J}^\top$ as follows
\begin{equation}
\mb{J}^\dagger = \left( \mb{J}^\top \mb{J} + \lambda \mb{I} \right)^{-1} \mb{J}^\top,
\end{equation}
and the solution can be represented with the new notation as
\begin{equation}
\dot{\mb{q}} = \mb{J}^\dagger \bs{\nu}.
\end{equation}
We also name $\mb{J}^\dagger$ as the \textit{singularity-robust inverse} of the Jacobian matrix $\mb{J}$.

\begin{figure}[t]
    \centering
    \includegraphics[width=0.9\textwidth]{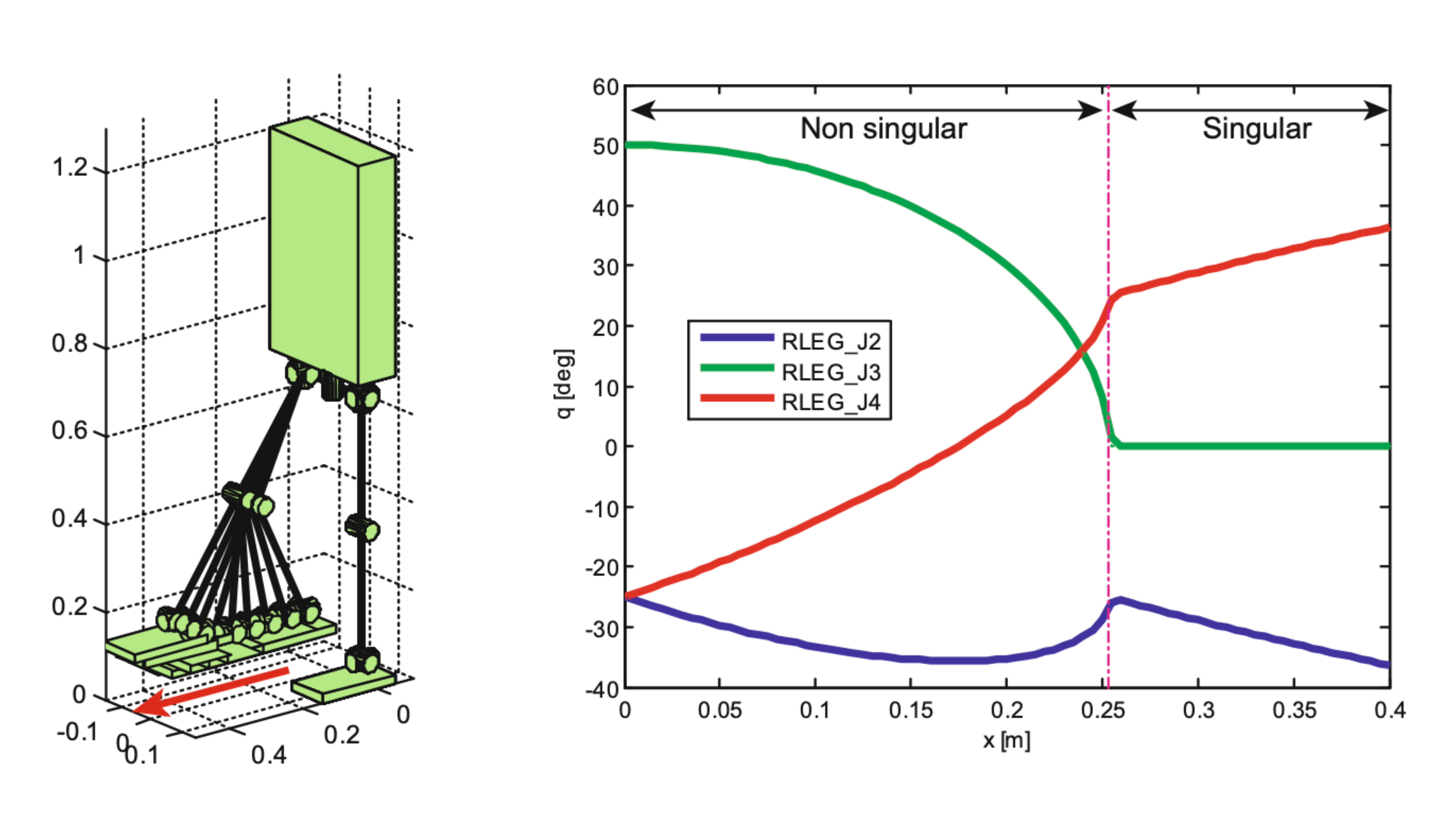}
    \caption{An Illustration of Legged Robot that Gets Rid of Singularity \cite{10.5555/3100040}.}
    \label{fig:singularity_after}
\end{figure}

\begin{example}
Let's come back to the example we show at the beginning of this part. As shown in Figure~\ref{fig:singularity_after}, to move the robot right foot forward, with the new \textit{singularity-robust inverse} of the Jacobian matrix $\mb{J}$, we can get rid of the \textit{singular pose} of the robot perfectly in the movement. As shown in the right plot of the joint angles, we can still control the robot pretty without singularity problem at all.
\end{example}

\section{Robot Dynamics}\label{sec:robot_dynamics}

We have been introducing the robot joint rotation, robot motion, forward kinematics and inverse kinematics for readers already, which covers the transformation of robot \textit{position}, \textit{orientation}, \textit{linear velocity} and \textit{angular velocity} within as well as across coordinate systems. Meanwhile, we didn't discuss much about the driven forces that lead to such changes yet, which will be the main topic to be studied in this section for readers, {\ie}, the \textit{robot dynamics}.


\subsection{What is Robot Dynamics?}

Formally, \textit{robot dynamics} aims to study the relationship between the forces (or torques) that act on a robot and the accelerations that they produce. For the robot studied in this article, the forces covers both the internal forces, {\eg}, the driving forces of joint motors, as well as external forces, {\eg}, the gravity of robot links and body, as well as the items the robot end points trying to pickup. Meanwhile, the generated accelerations overtime will create velocity and further change the pose of robot parts and body.

There exist different formulation methods of \textit{robot dynamics}. In this section, we will introduce two dynamics formulation methods for readers that are initially proposed in Physics but can also be used for formulating robot dynamics as well, {\ie}, 
\begin{itemize}
\item \textbf{Lagrangian Dynamics Formulation}: In the \textit{Lagrangian dynamics}, motion is described by energies as scalars and it is based on the principle of least action. Generalized coordinates are used in \textit{Lagrangian dynamics} instead of constraint forces, and conservation laws can be derived easily. \textit{Lagrangian dynamics} is not idea for non-conservative forces, {\eg}, friction, and widely used in all areas of physics.
\item \textbf{Newton-Euler Dynamics Formulation}: In the \textit{Newtonian dynamics}, motion is described by forces with vectors and it is based on the Newton's laws of motion. \textit{Newtonian dynamics} involves constraint forces, and doesn't have a systematic methods for deriving conservation laws. \textit{Lagrangian dynamics} handles non-conservative forces very well, and mainly applicable to classic physics.
\end{itemize}

There are two main problems that are studied in \textit{robot dynamics}, which include
\begin{itemize}
\item \textbf{Forward Dynamics}: Given the known \textit{forces}, calculate the \textit{acceleration} (as well as \textit{velocity} and \textit{pose}) of the robot generated by the forces.
\item \textbf{Inverse Dynamics}: Given the desired \textit{acceleration} (as well as \textit{velocity} and \textit{pose}), figure out the needed \textit{forces} to generate them.
\end{itemize}
Similar to the \textit{forward kinematics} and \textit{inverse kinematics} discussed before, the \textit{forward dynamics} is normally used for simulation, while the \textit{inverse dynamics} has very diverse usage in \textit{robot control}.

In this section, we first will introduce \textit{Lagrangian dynamics formulation} and \textit{Newton-Euler dynamics formulation} for readers. At the end, we will use a concrete example of motor driven robot link to discuss about the \textit{forward dynamics} and \textit{inverse dynamics} for readers, and the following subsections are also organized according to these topics.

\subsection{Lagrangian Dynamics Formulation}

In this part, we will introduce the Lagrangian formulation of dynamics and discuss about how to use it to model the dynamics of robot arms.

\subsubsection{Euler-Lagrange Equation}\label{subsubsec:euler_lagrange_equation}

To use Lagrangian formulation of robot dynamics, the first step is to choose a set of independent coordinates $\mb{q} \in \mathbbm{R}^n$ that defines the robot configuration. The chosen coordinates can not only be the Cartesian coordinate but also other configurations of the robot, {\eg}, the angles of the joints, which are also named as the \textit{generalized coordinates}. The set of all potential configurations $\mb{q}$ will define the \textit{configuration space} $\mathbbm{C}$. 

\begin{definition}
(\textbf{Lagrangian Dynamics}): Formally, the \textit{Lagrangian dynamics} formulates a dynamics system as a pair $(\mathbbm{C}, L)$, where $\mathbbm{C}$ denotes the \textit{configuration space} and $L$ is the Lagrangian function defined as follows:
\begin{equation}
L(\mb{q}, \dot{\mb{q}}) = K(\mb{q}, \dot{\mb{q}}) - P(\mb{q}) \text{, where } \mb{q} \in \mathbbm{C}.
\end{equation}
In the above equation, terms $K(\mb{q}, \dot{\mb{q}})$ and $P(\mb{q})$ denote the \textit{kinetic energy} and \textit{potential energy} of the system, respectively. Notation $\dot{\mb{q}}$ denotes the first-order derivative of term $\mb{q} \in \mathbbm{C}$.
\end{definition}

After choosing the \textit{generalized coordinates}, we can also further define the \textit{generalized forces} $\mb{f} \in \mathbbm{R}^n$ subject to the constraint that $\mb{f}^\top \mb{q}$ corresponds to power. Depending on the chosen \textit{generalized coordinates}, the \textit{generalized forces} $\mb{f}$ can not only represent regular forces, but can also represent other properties of the robot, {\eg}, the torque of joints. For instance, if \textit{generalized coordinates} $\mb{q}$ denotes the position in Cartesian coordinate, then $\mb{f}$ will be regular forces, since the product of forces and moving distance will denote the power. Meanwhile, if $\mb{q}$ denotes the rotation angles, then $\mb{f}$ should represent the torques, since the product of torque and rotation angles denotes the power.

Based on the \textit{Lagrangian function} defined above, any equations of motion can now be expressed in terms of the Lagrangian as follows:
\begin{equation}\label{equ:euler_lagrange_equation}
\mb{f} = \frac{d}{dt} \frac{\partial L}{\partial \dot{\mb{q}}} - \frac{\partial L}{\partial \mb{q}},
\end{equation}
where $\mb{f}$ denotes the \textit{generalized forces}. Such an equation is also referred to as the \textit{Euler-Lagrange equation}.

Instead of forces, \textit{Lagrangian dynamics} uses energy in the system formulation, and the \textit{Lagrangian function} defined above summarizes the dynamics of the entire system. Some readers may wonder ``why the \textit{Lagrangian function} is defined as the kinetic energy minus the potential energy?'' According to Physics, defining \textit{Lagrangian function} in this way will allow us to generate the correct equations of motion in agreement with classic physical laws, {\eg}, the ``\textit{Newton's laws of motion}''.

\begin{example}
To illustrate that the above \textit{Lagrangian function} can be used to define the equations of motion, we also provide an example of a particle of mass $m$ that is moving within a 1D space along a vertical line. We represent the position ({\ie}, the height) of the particle as a variable $q$, while its speed and acceleration can be denoted as $\dot{q}$ and $\ddot{q}$, respectively. The gravitational force acting on the particle will generate an acceleration represented by $g$. According to what we learn from high-school Physics course, we can represent the motion of the particle as
\begin{equation}\label{equ:newtow_second_law}
f - mg = m\ddot{q},
\end{equation}
where $f$ is the external force acting on the particle to drive its motion. If readers still remember, the above equation is derived according to the classic \textit{Newton's second law of motion}.

Meanwhile, the \textit{Euler-Lagrange equation} introduced above will also help derive the identical representation of the external force. For the particle, we can represent its \textit{kinetic energy} and \textit{potential energy} as follows:
\begin{equation}
K(q, \dot{q}) = \frac{1}{2}m\dot{q}^2 \text{, } P(q) = mgq,
\end{equation}
which will define the Lagrangian function
\begin{equation}
L(q, \dot{q}) = K(q, \dot{q}) - P(q) = \frac{1}{2}m\dot{q}^2 - mgq.
\end{equation}

According to Equation~\ref{equ:euler_lagrange_equation}, we can represent the equation of motion as 
\begin{align}
f &= \frac{d}{dt} \frac{\partial L}{\partial \dot{{q}}} - \frac{\partial L}{\partial {q}} \\
&= m \ddot{q} + mg,
\end{align}
which is identical to Equation~\ref{equ:newtow_second_law} derived from \textit{Newton's second law of motion}.
\end{example}

\subsubsection{Lagrangian Dynamics of 2-Link Robot Arm}\label{subsubsec:2_link_arm_dynamics}

In the previous subsection, we introduce the \textit{Lagrangian formulation} about the dynamics of a particle, which can derive the identical dynamic equation as \textit{Newton's second law of motion}.

\begin{figure}[t]
    \centering
    \includegraphics[width=0.9\textwidth]{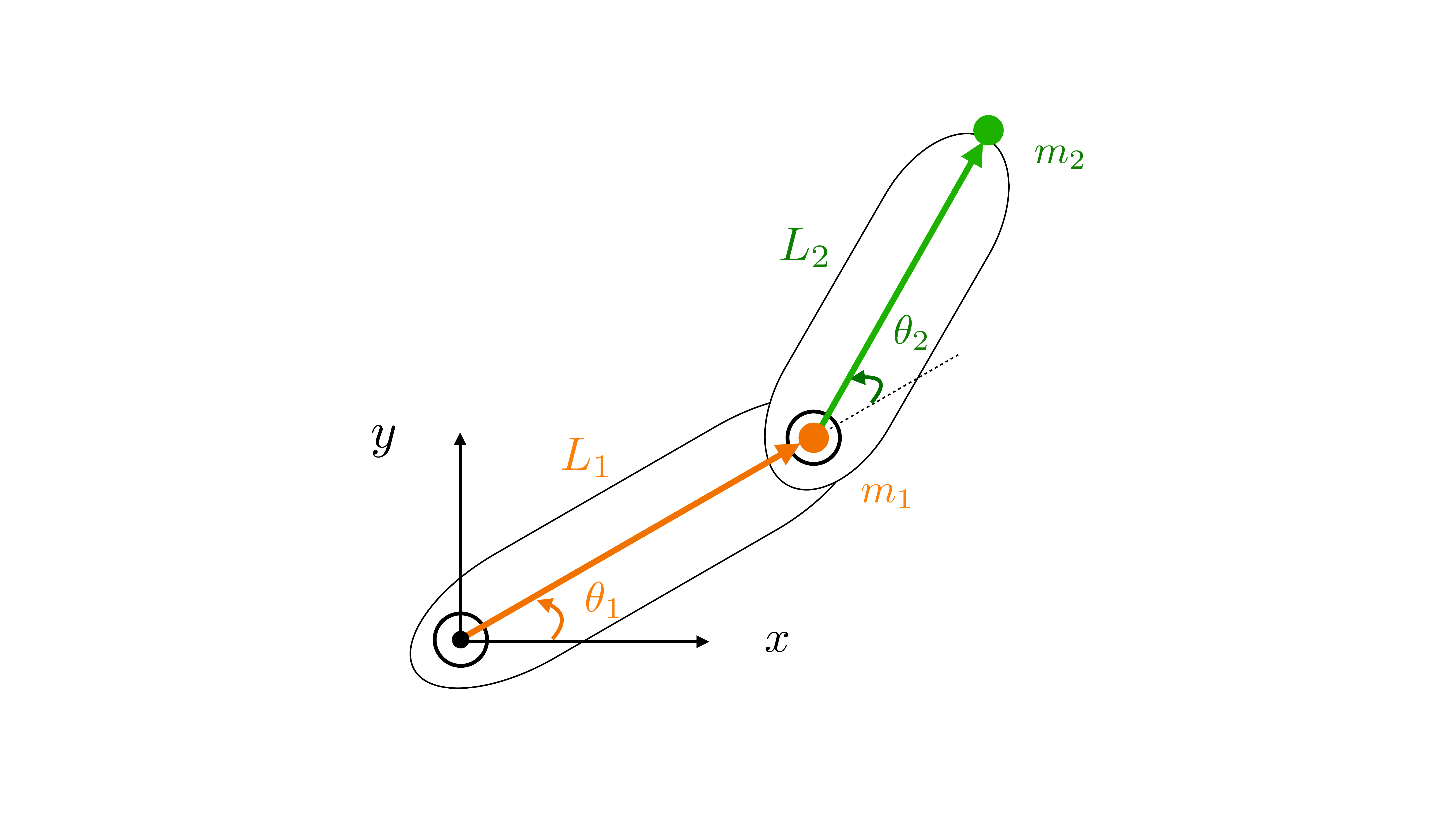}
    \caption{Dynamics of Two-Link Robot Arm.}
    \label{fig:two_link_arm_dynamics}
\end{figure}

\begin{example}
As illustrated in Figure~\ref{fig:two_link_arm_dynamics}, we provide a 2-link robot arm in a 2D space. The positive $x$ and $y$ axis orientations are provided by the coordinate system with origin located at the root of the arm. The length and mass of these two arm links are $L_1$, $m_1$ and $L_2$, $m_2$, respectively. In this example, we assume the mass of these two links are concentrated at the links' end points, {\ie}, the orange and green dots in the figure. The first link rotates in the counter clockwise direction with an angle of $\theta_1$ degrees, and the second link further rotates with another angle of $\theta_2$ degrees.
\end{example}

According to the link length and rotation angles, we can represent the position of the these two arm links' end points in the coordinate system as
\begin{equation}
\begin{bmatrix}
x_1\\
y_1\\
\end{bmatrix} = \begin{bmatrix}
L_1 \cos (\theta_1) \\
L_1 \sin (\theta_1)\\
\end{bmatrix} \text{, and }
\begin{bmatrix}
x_2\\
y_2\\
\end{bmatrix} = \begin{bmatrix}
L_1 \cos (\theta_1)+L_2 \cos (\theta_1+\theta_2) \\
L_1 \sin (\theta_1)+L_2 \sin (\theta_1+\theta_2)\\
\end{bmatrix}.
\end{equation}
The \textit{velocity} of these two arm links' end points can be represented as the derivative of the position with respect to time, {\ie},
\begin{align}
&\begin{bmatrix}
\dot{x}_1\\
\dot{y}_1\\
\end{bmatrix} = \begin{bmatrix}
-L_1 \sin (\theta_1) \\
L_1 \cos (\theta_1) \\
\end{bmatrix} \dot{\theta}_1,\\
&\begin{bmatrix}
\dot{x}_2\\
\dot{y}_2\\
\end{bmatrix} = \begin{bmatrix}
-L_1 \sin (\theta_1) - L_2 \sin (\theta_1+\theta_2) & - L_2 \sin (\theta_1+\theta_2) \\
L_1 \cos (\theta_1) \dot{\theta}_1 + L_2 \cos (\theta_1+\theta_2) & L_2 \cos (\theta_1+\theta_2)\\
\end{bmatrix}
\begin{bmatrix}
\dot{\theta}_1\\
\dot{\theta}_2\\
\end{bmatrix},
\end{align}
where $\dot{\theta}_1 = \frac{d \theta_1}{dt}$ and $\dot{\theta}_2 = \frac{d \theta_2}{dt}$ denote the derivatives of $\theta_1$ and $\theta_2$ with respect to the time.

Meanwhile, based on the above derivations, we can also calculate the current \textit{kinetic energy} and \textit{potential energy} of the robot arm as
\begin{align}\label{equ:two_link_robot_kp_equation}
K_1 &= \frac{1}{2} m_1 (\dot{x}_1^2 + \dot{y}_1^2) = \frac{1}{2} m_1 L_1 \dot{\theta}_1^2,\\
K_2 &= \frac{1}{2} m_2 (\dot{x}_2^2 + \dot{y}_2^2)\\
& = \frac{1}{2} m_2 \left( (L_1^2+2L_1L_2\cos(\theta_2)+L_2^2)\dot{\theta}_1^2 + 2(L_2^2+L_1L_2\cos(\theta_2))\dot{\theta}_1\dot{\theta}_2 + L_2^2 \dot{\theta}_2^2 \right),\\
P_1 &= m_1 g y_1 = m_1 g L_1 \sin(\theta_1),\\
P_2 &= m_1 g y_1 = m_2 g \left(L_1 \sin(\theta_1) + L_2 \sin (\theta_1 + \theta_2) \right).
\end{align}
By combining them together, we will be able to define the \textit{Lagrangian functoin}
\begin{equation}
L(\bs{\theta}, \dot{\bs{\theta}}) = \sum_{i=1}^2 (K_i - P_i).
\end{equation}

By choosing the joint angles $\bs{\theta} = \begin{bmatrix} \theta_1\\ \theta_2\\ \end{bmatrix}$ as the \textit{generalized coordinates}, as discussed in the previous Section~\ref{subsubsec:euler_lagrange_equation}, then the corresponding \textit{generalized forces} derived in Equation~\ref{equ:euler_lagrange_equation} will actually be the \textit{torques} $\bs{\tau} = \begin{bmatrix} \bs{\tau}_1\\ \bs{\tau}_2\\ \end{bmatrix}$:
\begin{align}
\bs{\tau}_1 = \frac{d}{dt} \frac{\partial L}{\partial \dot{\theta}_1} - \frac{\partial L}{\partial \theta_1} \text{, and } \bs{\tau}_2 = \frac{d}{dt} \frac{\partial L}{\partial \dot{\theta}_2} - \frac{\partial L}{\partial \theta_2}.
\end{align}
These two terms above together will define the torque vector $\bs{\tau}$ as follows
\begin{equation}\label{equ:2_link_arm_dynamics}
\bs{\tau} = \mb{M}(\bs{\theta}) \ddot{\bs{\theta}} + { \mb{c}(\bs{\theta}, \dot{\bs{\theta}}) + \mb{g}(\bs{\theta})},
\end{equation}
where
\begin{align}
\mb{M}(\bs{\theta}) &= \underbrace{\begin{bmatrix}
m_1L_1^2 + m_2(L_1^2 + 2L_1 L_2 \cos{\theta_2} + L_2^2) & m_2(L_1 L_2 \cos{\theta_2} + L_2^2)\\
m_2(L_1 L_2 \cos(\theta_2) + L_2^2) & m_2 L_2^2\\
\end{bmatrix}}_{\text{symmetric positive-definite mass matrix}},\\
\mb{c}(\bs{\theta}, \dot{\bs{\theta}}) &= \begin{bmatrix}
-m_2 L_1 L_2 \sin(\theta_2) (2 \dot{\theta}_1\dot{\theta}_2 + \dot{\theta}_2^2 )\\
m_2 L_1 L_2 \dot{\theta}_1^2 \sin (\theta_2)\\
\end{bmatrix}\\
&= \underbrace{\begin{bmatrix}
-m_2 L_1 L_2 \sin(\theta_2)  2\dot{\theta}_1\dot{\theta}_2 \\
0\\
\end{bmatrix}}_{\text{Coriolis torque}}
+ \underbrace{\begin{bmatrix}
-m_2 L_1 L_2 \sin(\theta_2) \dot{\theta}_2^2 \\
m_2 L_1 L_2 \sin (\theta_2) \dot{\theta}_1^2\\
\end{bmatrix}}_{\text{centripetal torque}}\\
\mb{g}(\bs{\theta}) &= \underbrace{\begin{bmatrix}
(m_1+m_2)L_1 g \cos(\theta_1) + m_2 g L_2 \cos(\theta_1 + \theta_2)\\
m_2 g L_2 \cos(\theta_1 + \theta_2)\\
\end{bmatrix}}_{\text{gravitational torque}}.
\end{align}
Formally, term $\mb{M}(\bs{\theta}) \in \mathbbm{R}^{2 \times 2}$ is a symmetric positive-definite mass matrix about the two robot arm links; vector $\mb{c}(\bs{\theta}, \dot{\bs{\theta}}) \in \mathbbm{R}^{2}$ contains both the \textit{Coriolis torque} and the \textit{centripetal torque} torque; and vector $\mb{g}(\bs{\theta}) \in \mathbbm{R}^{2}$ is the \textit{gravitational torque}. All these matrix and vectors are defined based on the general coordinate vector $\bs{\theta}$ and its first-order derivative $\dot{\bs{\theta}}$.

\subsubsection{Lagrangian Dynamics of Multi-Link Robot Arm}

In this part, we further provide a general Lagrangian formulation of robot arms with multiple links. Specifically, we can denote the robot arm link number to be $n$ ($n \ge 2$). Similar to the previous subsection, we can denote the joint rotation angle vector $\bs{\theta} \in \mathbbm{R}^n$ as the chosen \textit{general coordinates}. Meanwhile, the \textit{general forces} calculated with the \textit{Euler Lagrange equation} will represent the \textit{torques} $\bs{\tau} \in \mathbbm{R}^n$ instead.

The \textit{Lagrangian function} defined base don the chosen generalized coordinates can be represented as
\begin{equation}
L(\bs{\theta}, \dot{\bs{\theta}}) = K(\bs{\theta}, \dot{\bs{\theta}}) - P(\bs{\theta}),
\end{equation}
where 
\begin{equation}
K(\bs{\theta}, \dot{\bs{\theta}}) = \frac{1}{2} \sum_{i=1}^n \sum_{j=1}^n m_{i,j}(\bs{\theta}) \dot{\theta}_i \dot{\theta}_j = \frac{1}{2} \dot{\bs{\theta}}^\top \mb{M}(\bs{\theta}) \dot{\bs{\theta}}.
\end{equation}
Meanwhile, term $P(\bs{\theta})$ denotes a vector of elements as represented similar to the $P_1$ and $P_2$ terms in Equation~\ref{equ:two_link_robot_kp_equation}. The matrix $\mb{M}(\bs{\theta}) \in \mathbbm{R}^{n \times n}$ is the mass matrix with term $m_{i,j}(\bs{\theta})$ as its $(i,j)_{th}$ element, which will be discussed in detail in the following Section~\ref{subsec:newton_euler_dynamics_formulation} instead.

According to the \textit{Euler-Lagrange equation}, we can represent the corresponding \textit{torque} vector as
\begin{equation}\label{equ:euler_lagrange_equation}
\bs{\tau} = \mb{M}(\bs{\theta}) \ddot{\bs{\theta}} + \dot{\bs{\theta}}^\top \bs{\Gamma}(\bs{\theta}) \dot{\bs{\theta}} + g(\bs{\theta}),
\end{equation}
where $g(\bs{\theta}) = \frac{\partial L}{\partial \bs{\theta}}$ and $\ddot{\bs{\theta}}$ denotes the second-order derivative of the coordinate vector $\bs{\theta}$. As to $\bs{\Gamma}(\bs{\theta}) \in \mathbbm{R}^{n \times n \times n}$, it is a 3-way tensor with the $(i,j,k)_{th}$ element
\begin{equation}
\Gamma_{ijk}(\bs{\theta}) = \frac{1}{2} \left( \frac{\partial m_{i,j}(\bs{\theta}) }{\partial \theta_k} + \frac{\partial m_{i,k}(\bs{\theta}) }{\partial \theta_j} - \frac{\partial m_{j,k}(\bs{\theta}) }{\partial \theta_i} \right).
\end{equation}

The term $\dot{\bs{\theta}}^\top \bs{\Gamma}(\bs{\theta}) \dot{\bs{\theta}}$ used in the above equation will be a vector in the following form
\begin{equation}
\dot{\bs{\theta}}^\top \bs{\Gamma}(\bs{\theta}) \dot{\bs{\theta}} = \begin{bmatrix}
\dot{\bs{\theta}}^\top \bs{\Gamma}_1(\bs{\theta}) \dot{\bs{\theta}}\\
\dot{\bs{\theta}}^\top \bs{\Gamma}_2(\bs{\theta}) \dot{\bs{\theta}}\\
\vdots\\
\dot{\bs{\theta}}^\top \bs{\Gamma}_n(\bs{\theta}) \dot{\bs{\theta}}\\
\end{bmatrix},
\end{equation}
where notation $\bs{\Gamma}_i(\bs{\theta}) \in \mathbbm{R}^{n \times n}$ is a $n \times n$ matrix with $\Gamma_{ijk}(\bs{\theta})$ as its $(j,k)_{th}$ element.

\subsection{Newton-Euler Dynamics Formulation}\label{subsec:newton_euler_dynamics_formulation}

In this part, we will introduce another formulation method, {\ie}, the \textit{Newton-Euler method}, to model the robot dynamics. Different from the Lagrangian method, which models the dynamics from the perspective of system energy, the \textit{Newton-Euler method} models the dynamics based on the concepts and transformation of motion.

\subsubsection{Newton-Euler Formulation for Rigid Body}

\begin{figure}[t]
    \centering
    \includegraphics[width=0.8\textwidth]{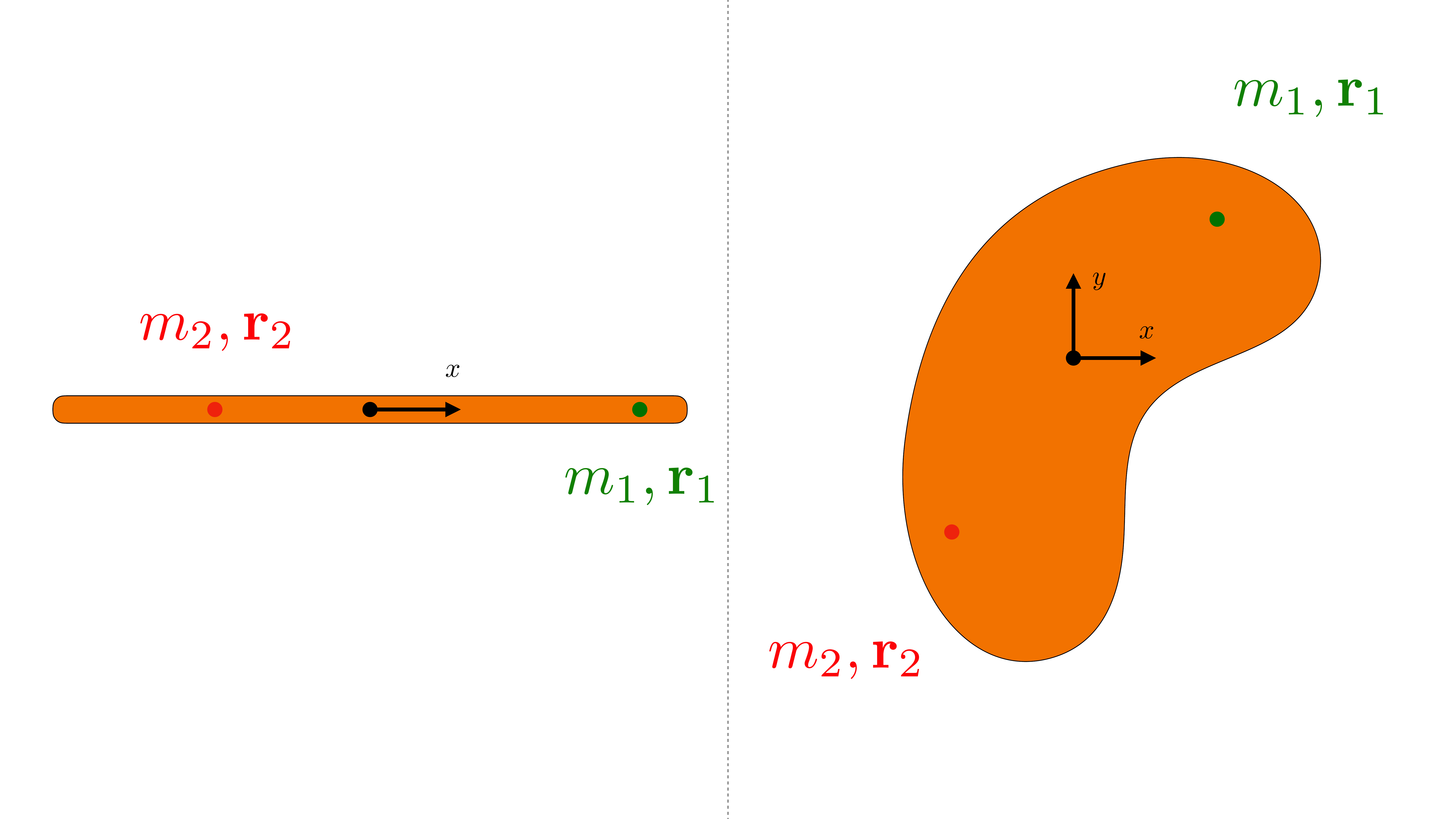}
    \caption{Dynamics of Rigid Body.}
    \label{fig:rigid_body_dynamics}
\end{figure}

In this part, we will take the single rigid body as an example to introduce the \textit{Newton-Euler formulation} method for readers. The rigid body can be in various shapes depending on the dimension of space we are studying the problem. 

\begin{example}
As shown in Figure~\ref{fig:rigid_body_dynamics}, a rigid body consists of a number of rigidly connected particles. Depending on the dimension of space we are studying the rigid body, it can be in different shapes, {\eg}, a line in 1D space as shown in the left plot of Figure~\ref{fig:rigid_body_dynamics} or a cashew shaped diagram in 2D space as shown in the right plot of Figure~\ref{fig:rigid_body_dynamics}.
\end{example}

To be general, we assume we are studying the problem within a 3D space. We can represent the set of particles in the rigid body as $\mc{S} = \{(m_i, \mb{r}_i)\}_{i=1}^n$, where $m_i$ denotes the mass of the $i_{th}$ particles and $\mb{r}_i = (x_i, y_i, z_i)$ denotes the initial position of the particles within the local coordinate about the rigid body. Formally, the origin of the local coordinate system is the unique point such that
\begin{equation}
\sum_{i=1}^n m_i \mb{r}_i = \mb{0}.
\end{equation} 
Such a point is also named as the center of mass about the rigid body.

Here, we assume the rigid body is moving with a body twist $\bs{\nu}_b = \begin{bmatrix} \bs{\omega}_b\\ \mb{v}_b \end{bmatrix}$ (the subscript $b$ denotes the rigid body), then all the particles within the rigid body will also change their position with time. For instance, for the $i_{th}$ particle, we can represent its position as a vector $\mb{p}_i(t)$. Within the local coordinate system, we can also calculate the \textit{linear velocity}, \textit{acceleration}, \textit{external force} and \textit{torque} about the $i_{th}$ particle in the rigid body as
\begin{align}
\textbf{Velocity}: \dot{\mb{p}}_i & = \mb{v}_b + \bs{\omega}_b \times \mb{p}_i,\\
\textbf{Acceleration}: \ddot{\mb{p}}_i &= \frac{d}{dt} \dot{\mb{p}}_i\\
&= \frac{d}{dt} \mb{v}_b + \frac{d}{dt}\bs{\omega}_b \times \mb{p}_i + \bs{\omega}_b \times \frac{d}{dt} \mb{p}_i\\
&= \dot{\mb{v}}_b + \dot{\bs{\omega}}_b \times \mb{p}_i + \bs{\omega}_b \times \dot{\mb{p}}_i \\
&= \dot{\mb{v}}_b + \dot{\bs{\omega}}_b \times \mb{p}_i + \bs{\omega}_b \times (\mb{v}_b + \bs{\omega}_b \times \mb{p}_i)\\
&= \dot{\mb{v}}_b + \widehat{\dot{\bs{\omega}}}_b \mb{r}_i + \widehat{\bs{\omega}}_b \mb{v}_b + \widehat{\bs{\omega}}_b^2 \mb{r}_i,\\
\textbf{Force}: \mb{f}_i &= m_i \ddot{\mb{p}}_i = m_i \left( \dot{\mb{v}}_b + \widehat{\dot{\bs{\omega}}}_b \mb{r}_i + \widehat{\bs{\omega}}_b \mb{v}_b + \widehat{\bs{\omega}}_b^2 \mb{r}_i \right),\\
\textbf{Torque}: \bs{\tau}_i &= \mb{r}_i \times \mb{f}_i = \widehat{\mb{r}}_i \mb{f}_i.
\end{align}

Furthermore, based on the force and torque representations of the $i_{th}$ particle, we can further calculate the force acting on the whole rigid body as
\begin{equation}\label{equ:rigid_body_force}
\begin{aligned}
\mb{f}_b = \sum_{i=1}^n \mb{f}_i &= \sum_{i=1}^n m_i \left( \dot{\mb{v}}_b + \widehat{\dot{\bs{\omega}}}_b \mb{r}_i + \widehat{\bs{\omega}}_b \mb{v}_b + \widehat{\bs{\omega}}_b^2 \mb{r}_i \right) \\
&= \sum_{i=1}^n m_i \left( \dot{\mb{v}}_b + \widehat{\bs{\omega}}_b \mb{v}_b \right) + \sum_{i=1}^n m_i \left(\widehat{\dot{\bs{\omega}}}_b \mb{r}_i \right) + \sum_{i=1}^n m_i \left( \widehat{\bs{\omega}}_b^2 \mb{r}_i \right)\\
&= \sum_{i=1}^n m_i \left( \dot{\mb{v}}_b + \widehat{\bs{\omega}}_b \mb{v}_b \right) - \sum_{i=1}^n m_i \widehat{\mb{r}_i} \dot{\bs{\omega}}_b + \sum_{i=1}^n m_i \widehat{\mb{r}}_i \widehat{\bs{\omega}}_b  \bs{\omega}_b\\
&= \sum_{i=1}^n m_i \left( \dot{\mb{v}}_b + \widehat{\bs{\omega}}_b \mb{v}_b \right) - \underbrace{\left( \sum_{i=1}^n m_i \widehat{\mb{r}_i} \right) \dot{\bs{\omega}}_b}_{= \mb{0}} + \underbrace{\left( \sum_{i=1}^n m_i \widehat{\mb{r}}_i \right) \widehat{\bs{\omega}}_b  \bs{\omega}_b}_{= \mb{0}}\\
&= \sum_{i=1}^n m_i \left( \dot{\mb{v}}_b + \widehat{\bs{\omega}}_b \mb{v}_b \right)
\end{aligned}
\end{equation}
In the above derivation, we use many properties about the $\land$ operator on vectors, {\eg}, (1) $\widehat{\dot{\bs{\omega}}}_b \mb{r}_i = -\widehat{\mb{r}_i} \dot{\bs{\omega}}_b$, (2) $\widehat{\bs{\omega}}_b \widehat{\mb{r}}_i = (\widehat{\mb{r}}_i \widehat{\bs{\omega}}_b)^\top$, and (3) $(\widehat{\mb{r}}_i \widehat{\bs{\omega}}_b)^\top = - \widehat{\mb{r}}_i \widehat{\bs{\omega}}_b$. Also we know that $\sum_{i=1}^n m_i \mb{r}_i = \mb{0}$, by applying the $\land$ operator to both sides of the equation, we can get that $\sum_{i=1}^n m_i \widehat{\mb{r}}_i = \widehat{\mb{0}}$, where $\widehat{\mb{0}}$ is an all-zero matrix.

As to the torque of the whole rigid body, we can represent it as
\begin{equation}\label{equ:rigid_body_torque_2}
\begin{aligned}
\bs{\tau}_b = \sum_{i=1}^n \bs{\tau}_i &= \sum_{i=1}^n \widehat{\mb{r}}_i m_i \left( \dot{\mb{v}}_b + \widehat{\dot{\bs{\omega}}}_b \mb{r}_i + \widehat{\bs{\omega}}_b \mb{v}_b + \widehat{\bs{\omega}}_b^2 \mb{r}_i \right)\\
&= \underbrace{\left(\sum_{i=1}^n \widehat{\mb{r}}_i m_i \right) \left( \dot{\mb{v}}_b + \widehat{\bs{\omega}}_b \mb{v}_b \right)}_{=\mb{0}} + \sum_{i=1}^n \widehat{\mb{r}}_i m_i \left( \widehat{\dot{\bs{\omega}}}_b \mb{r}_i + \widehat{\bs{\omega}}_b^2 \mb{r}_i \right)\\
&= \sum_{i=1}^n m_i \left( - \widehat{\mb{r}}_i^2 \dot{\bs{\omega}}_b - \widehat{\mb{r}}_i \widehat{\bs{\omega}}_b \widehat{\mb{r}}_i \bs{\omega}_b \right)\\
&= \sum_{i=1}^n m_i \left( - \widehat{\mb{r}}_i^2 \dot{\bs{\omega}}_b - \widehat{\bs{\omega}}_b \widehat{\mb{r}}_i^2 \bs{\omega}_b \right)\\
&= \left( -\sum_{i=1}^n m_i \widehat{\mb{r}}_i^2 \right) \dot{\bs{\omega}}_b + \widehat{\bs{\omega}}_b \left( \sum_{i=1}^n m_i \widehat{\mb{r}}_i^2 \right) \bs{\omega}_b.
\end{aligned}
\end{equation}
In the above derivation, we use a property about vector cross-product that $\mb{a} \times \left( \mb{b} \times (\mb{a} \times \mb{b}) \right) = \mb{b} \times \left( \mb{a} \times (\mb{a} \times \mb{b}) \right)$ to convert $\widehat{\mb{r}}_i \widehat{\bs{\omega}}_b \widehat{\mb{r}}_i \bs{\omega}_b$ into $\widehat{\bs{\omega}}_b \widehat{\mb{r}}_i^2 \bs{\omega}_b$. As to proof of equation $\mb{a} \times \left( \mb{b} \times (\mb{a} \times \mb{b}) \right) = \mb{b} \times \left( \mb{a} \times (\mb{a} \times \mb{b}) \right)$, we will leave it as an exercise for the readers. 

\subsubsection{Rotational Inertia Matrix}

In the previous subsection, we try to derive the representations of \textit{velocity}, \textit{acceleration}, \textit{force} and \textit{torque} of both particles in a rigid body, as well as the whole rigid body. For the torque of the rigid body, we can also simplify its representation derived in Equation~\ref{equ:rigid_body_torque} as follows:
\begin{equation}\label{equ:rigid_body_torque}
\begin{aligned}
\bs{\tau}_b &= \left( -\sum_{i=1}^n m_i \widehat{\mb{r}}_i^2 \right) \dot{\bs{\omega}}_b + \widehat{\bs{\omega}}_b \left( \sum_{i=1}^n m_i \widehat{\mb{r}}_i^2 \right) \bs{\omega}_b\\
&= \mb{I}_b \dot{\bs{\omega}}_b + \widehat{\bs{\omega}}_b \mb{I}_b \bs{\omega}_b.
\end{aligned}
\end{equation}
In the above representation, we introduce a new matrix representation
\begin{align}
\mb{I}_b &= -\sum_{i=1}^n m_i \widehat{\mb{r}}_i^2 \\
 &= -\sum_{i=1}^n m_i \begin{bmatrix}
 0 & -z_i & y_i\\
 z_i & 0 & -x_i\\
 -y_i & x_i & 0\\
 \end{bmatrix} \begin{bmatrix}
 0 & -z_i & y_i\\
 z_i & 0 & -x_i\\
 -y_i & x_i & 0\\
 \end{bmatrix} \\
 &= \begin{bmatrix}
\sum_{i=1}^n m_i(y_i^2+z_i^2) &  -\sum_{i=1}^n m_i x_iy_i &  -\sum_{i=1}^n m_i x_iz_i\\
-\sum_{i=1}^n m_i x_i y_i & \sum_{i=1}^n m_i (x_i^2 + z_i^2) & - \sum_{i=1}^n m_i y_i z_i\\
-\sum_{i=1}^n m_i x_i z_i & -\sum_{i=1}^n m_i y_i z_i & \sum_{i=1}^n m_i (x_i^2 + y_i^2) \\
 \end{bmatrix} \\
 &= \begin{bmatrix}
 I_{xx} & I_{xy} & I_{xz}\\
 I_{xy} & I_{yy} & I_{yz}\\
 I_{zx} & I_{zy} & I_{zz}\\
 \end{bmatrix} \in \mathbbm{R}^{3 \times 3},
\end{align}
which is also named as the rigid body's \textit{rotational inertia matrix}. For a given rigid-body, we know that matrix $\mb{I}_b$ will be a constant and it is also symmetric and positive definite.

The \textit{rotational inertia matrix} also provides a simple representation of the \textit{rotational kinetic energy} of the rigid body. Formally, for all the particles in the rigid body, we can represent the \textit{rotational kinetic energy} of the rigid body as
\begin{align}
K &= \frac{1}{2} \sum_{i=1}^n m_i \mb{v}_i^\top \mb{v}_i \\
&= \frac{1}{2} \sum_{i=1}^n m_i (\bs{\omega} \times \mb{r}_i) \dot (\bs{\omega} \times \mb{r}_i)\\
&= \frac{1}{2} \sum_{i=1}^n m_i \bs{\omega} \cdot \mb{r}_i \times (\bs{\omega} \times \mb{r}_i)\\
&=\frac{1}{2} \bs{\omega} \cdot  \sum_{i=1}^n m_i  \mb{r}_i \times (- \mb{r}_i \times \bs{\omega} )\\
&= \frac{1}{2} \bs{\omega} \cdot  -\sum_{i=1}^n m_i \widehat{\mb{r}_i} (\widehat{\mb{r}_i} \bs{\omega}) \\
&=\frac{1}{2} \bs{\omega}^\top (-\sum_{i=1}^n m_i \widehat{\mb{r}_i}^2) \bs{\omega} \\
&=\frac{1}{2} \bs{\omega}^\top \mb{I}_b \bs{\omega}.
\end{align}

Via the \textit{rotational kinetic energy} value, we can also transform the \textit{rotational inertia matrix} across different coordinate systems. Given two coordinate systems $\Sigma_a$ and $\Sigma_c$, we can denote the rigid body's \textit{angular velocity} vectors in this two vectors as $\bs{\omega}^{[\Sigma_a]}$ and $\bs{\omega}^{[\Sigma_c]}$, where $\bs{\omega}^{[\Sigma_c]} = \mb{R}^{[\Sigma_a \to \Sigma_c]} \bs{\omega}^{[\Sigma_a]}$. Since the rigid body's \textit{rotational kinetic energy} in these two coordinate systems should be equal, we can have
\begin{align}
& K^{[\Sigma_a]} = K^{[\Sigma_c]}\\
\Longleftrightarrow \ \ & \frac{1}{2} (\bs{\omega}^{[\Sigma_a]})^\top \mb{I}_b^{[\Sigma_a]} \bs{\omega}^{[\Sigma_a]} = \frac{1}{2} (\bs{\omega}^{[\Sigma_c]})^\top \mb{I}_b^{[\Sigma_c]} \bs{\omega}^{[\Sigma_c]} \\
\Longleftrightarrow \ \ & \frac{1}{2} (\bs{\omega}^{[\Sigma_a]})^\top \mb{I}_b^{[\Sigma_a]} \bs{\omega}^{[\Sigma_a]} = \frac{1}{2} (\mb{R}^{[\Sigma_a \to \Sigma_c]} \bs{\omega}^{[\Sigma_a]})^\top \mb{I}_b^{[\Sigma_c]} (\mb{R}^{[\Sigma_a \to \Sigma_c]} \bs{\omega}^{[\Sigma_a]})\\
\Longleftrightarrow \ \ & \frac{1}{2} (\bs{\omega}^{[\Sigma_a]})^\top \mb{I}_b^{[\Sigma_a]} \bs{\omega}^{[\Sigma_a]} = \frac{1}{2} (\bs{\omega}^{[\Sigma_a]})^\top \left( (\mb{R}^{[\Sigma_a \to \Sigma_c]})^\top \mb{I}_b^{[\Sigma_c]} \mb{R}^{[\Sigma_a \to \Sigma_c]} \right) \bs{\omega}^{[\Sigma_a]}.
\end{align}
Since the above equation holds for any angular velocity $\bs{\omega}^{[\Sigma_a]}$, so we can get
\begin{equation}
\mb{I}_b^{[\Sigma_a]} = (\mb{R}^{[\Sigma_a \to \Sigma_c]})^\top \mb{I}_b^{[\Sigma_c]} \mb{R}^{[\Sigma_a \to \Sigma_c]}.
\end{equation}
As introduced before, the subscript $b$ of the \textit{rotational inertia matrix} representations above denote it is defined for the rigid body.

In addition to transformation across coordinate systems, the \textit{rotational inertia matrix} can also be transformed within the same coordinate system but at a different point. In the above derivation, matrix $\mb{I}_b$ is defined for the center of the rigid body, {\ie}, the center of the local coordinate. Given another point $\mb{q} = (q_x, q_y, q_z)$ within the coordinate, we can also define the \textit{rotational inertia matrix} with point $\mb{q}$ as the new origin to be
\begin{equation}
\mb{I}_q = \mb{I}_b + m \left(\mb{q}^\top \mb{q} \mb{I} - \mb{q} \mb{q}^\top \right).
\end{equation}
It is also referred to as the \textit{parallel axis theorem}, we will leave its proof as an exercise for the readers.

\subsubsection{Twist-Wrench Formulation}

The representation of \textit{force} and \textit{torque} acting on the rigid body we derive in Equation~\ref{equ:rigid_body_force} and Equation~\ref{equ:rigid_body_torque} can also be organized and represented together as follows:
\begin{align}
\underbrace{\begin{bmatrix}
\bs{\tau}_b\\
\mb{f}_b\\
\end{bmatrix} }_{\mb{w}_b}
&= 
\begin{bmatrix}
\mb{I}_b & \mb{0}\\
\mb{0} & m \mb{I}\\
\end{bmatrix} 
\begin{bmatrix}
\dot{\bs{\omega}}_b\\
\dot{\mb{v}}_b\\
\end{bmatrix} 
+
\begin{bmatrix}
\widehat{\bs{\omega}}_b & \mb{0}\\
\mb{0} & \widehat{\bs{\omega}}_b\\
\end{bmatrix}
\begin{bmatrix}
\mb{I}_b & \mb{0}\\
\mb{0} & m \mb{I}\\
\end{bmatrix}
\begin{bmatrix}
\bs{\omega}_b\\
\mb{v}_b\\
\end{bmatrix} \\
&= \begin{bmatrix}
\mb{I}_b & \mb{0}\\
\mb{0} & m \mb{I}\\
\end{bmatrix} 
\begin{bmatrix}
\dot{\bs{\omega}}_b\\
\dot{\mb{v}}_b\\
\end{bmatrix} 
+ \begin{bmatrix}
\widehat{\bs{\omega}}_b & \mb{0}\\
\mb{0} & \widehat{\bs{\omega}}_b\\
\end{bmatrix}
\begin{bmatrix}
\mb{I}_b & \mb{0}\\
\mb{0} & m \mb{I}\\
\end{bmatrix}
\begin{bmatrix}
\bs{\omega}_b\\
\mb{v}_b\\
\end{bmatrix}\\
&+ \underbrace{\begin{bmatrix}
\mb{0} & \widehat{\mb{v}}_b\\
\mb{0} & \mb{0}\\
\end{bmatrix}
\begin{bmatrix}
\mb{I}_b & \mb{0}\\
\mb{0} & m \mb{I}\\
\end{bmatrix}
\begin{bmatrix}
\bs{\omega}_b\\
\mb{v}_b\\
\end{bmatrix}}_{=\widehat{\mb{v}}_b \mb{v}_b = \mb{v}_b \times \mb{v}_b = \mb{0}}\\
&= \begin{bmatrix}
\mb{I}_b & \mb{0}\\
\mb{0} & m \mb{I}\\
\end{bmatrix} 
\begin{bmatrix}
\dot{\bs{\omega}}_b\\
\dot{\mb{v}}_b\\
\end{bmatrix} 
+ \begin{bmatrix}
\widehat{\bs{\omega}}_b & \widehat{\mb{v}}_b\\
\mb{0} & \widehat{\bs{\omega}}_b\\
\end{bmatrix}
\begin{bmatrix}
\mb{I}_b & \mb{0}\\
\mb{0} & m \mb{I}\\
\end{bmatrix}
\begin{bmatrix}
\bs{\omega}_b\\
\mb{v}_b\\
\end{bmatrix}\\
&= \underbrace{\begin{bmatrix}
\mb{I}_b & \mb{0}\\
\mb{0} & m \mb{I}\\
\end{bmatrix}}_{\mb{G}_b}
\underbrace{\begin{bmatrix}
\dot{\bs{\omega}}_b\\
\dot{\mb{v}}_b\\
\end{bmatrix}}_{\dot{\bs{\nu}}_b}
- \underbrace{{\begin{bmatrix}
\widehat{\bs{\omega}}_b & \mb{0}\\
\widehat{\mb{v}}_b & \widehat{\bs{\omega}}_b\\
\end{bmatrix}^\top}}_{Ad(\bs{\nu}_b)}
\underbrace{\begin{bmatrix}
\mb{I}_b & \mb{0}\\
\mb{0} & m \mb{I}\\
\end{bmatrix}}_{\mb{G}_b}
\underbrace{\begin{bmatrix}
\bs{\omega}_b\\
\mb{v}_b\\
\end{bmatrix}}_{\bs{\nu}_b}.
\end{align}
In the above equation derivation, we use many properties, {\eg}, (1) $\widehat{\mb{v}}_b \mb{v}_b = \mb{v}_b \times \mb{v}_b = \mb{0}$ and (2) $
\begin{bmatrix}
\widehat{\bs{\omega}}_b & \widehat{\mb{v}}_b\\
\mb{0} & \widehat{\bs{\omega}}_b\\
\end{bmatrix} = \begin{bmatrix}
\widehat{\bs{\omega}}_b^\top & \mb{0}\\
\widehat{\mb{v}}_b^\top & \widehat{\bs{\omega}}_b^\top\\
\end{bmatrix}^\top = -\begin{bmatrix}
\widehat{\bs{\omega}}_b & \mb{0}\\
\widehat{\mb{v}}_b & \widehat{\bs{\omega}}_b\\
\end{bmatrix}^\top$. Several notations are also introduced for the terms used in the above equation, which can be further simplified as
\begin{equation}
\mb{w}_b = \mb{G}_b \dot{\bs{\nu}}_b - Ad(\bs{\nu}_b)^\top \mb{G}_b \bs{\nu}_b,
\end{equation}
where $\mb{G}_b = \begin{bmatrix}
\mb{I}_b & \mb{0}\\
\mb{0} & m \mb{I}\\
\end{bmatrix} \in \mathbbm{R}^{6 \times 6}$ is the \textit{spatial inertia matrix} and $Ad(\bs{\nu}_b) = {\begin{bmatrix}
\widehat{\bs{\omega}}_b & \mb{0}\\
\widehat{\mb{v}}_b & \widehat{\bs{\omega}}_b\\
\end{bmatrix}^\top} \in \mathbbm{R}^{6 \times 6}$ is the adjoint matrix representation of vector $\bs{\nu}_b$.

With the $\mb{G}_b$ matrix, we can also represent the overall kinetic energy of the rigid body as
\begin{align}
K &= \frac{1}{2} \bs{\omega}_b^\top \mb{I}_b \bs{\omega}_b + \frac{1}{2}m \mb{v}_b^\top \mb{v}_b\\
&= \frac{1}{2} \begin{bmatrix}
\bs{\omega}_b & \mb{v}_b\\
\end{bmatrix}
\begin{bmatrix}
\mb{I}_b & \mb{0}\\
\mb{0} & m \mb{I}\\
\end{bmatrix}
\begin{bmatrix}
\bs{\omega}_b\\
\mb{v}_b\\
\end{bmatrix}\\
&= \frac{1}{2} \bs{\nu}_b^\top \mb{G}_b \bs{\nu}_b.
\end{align}

Furthermore, the matrix $\mb{G}_b$ define above can also be transformed across coordinate systems. Given two coordinate systems $\Sigma_a$ and $\Sigma_c$, we can represent the \textit{twist velocity} vectors of the rigid body in them as $\bs{\nu}_b^{[\Sigma_a]}$ and $\bs{\nu}_b^{[\Sigma_c]}$, respectively. According to the adjoint representation of the homogeneous transformation matrix introduced in Equation~\ref{equ:adjoint_representation_homogeneous_transformation} in the previous Section~\ref{subsubsec:adjoint_representation_homogeneous_transformation_matrix}, we know that
\begin{equation}
\bs{\nu}_b^{[\Sigma_c]} = Ad(\mb{T}^{[\Sigma_a \to \Sigma_c]}) \bs{\nu}_b^{[\Sigma_a]}.
\end{equation}
Meanwhile, since the kinetic energy of the rigid body is independent of the chosen coordinate system, we can have
\begin{align}
\frac{1}{2} (\bs{\nu}_b^{[\Sigma_a]})^\top \mb{G}_b^{[\Sigma_a]} \bs{\nu}_b^{[\Sigma_a]} &= \frac{1}{2} (\bs{\nu}_b^{[\Sigma_c]})^\top \mb{G}_b^{[\Sigma_c]} \bs{\nu}_b^{[\Sigma_c]}\\
&=\frac{1}{2} \left(Ad(\mb{T}^{[\Sigma_a \to \Sigma_c]}) \bs{\nu}_b^{[\Sigma_a]}\right)^\top \mb{G}_b^{[\Sigma_c]} \left(Ad(\mb{T}^{[\Sigma_a \to \Sigma_c]}) \bs{\nu}_b^{[\Sigma_a]}\right) \\
&= \frac{1}{2} (\bs{\nu}_b^{[\Sigma_a]})^\top \left( Ad(\mb{T}^{[\Sigma_a \to \Sigma_c]})^\top \mb{G}_b^{[\Sigma_c]} Ad(\mb{T}^{[\Sigma_a \to \Sigma_c]}) \right) \bs{\nu}_b^{[\Sigma_a]}.
\end{align}
Since the above equation should hold for any \textit{twist velocity} vector $\bs{\nu}_b^{[\Sigma_a]} \in \mathbbm{R}^6$, then we can get
\begin{equation}
\mb{G}_b^{[\Sigma_a]} = Ad(\mb{T}^{[\Sigma_a \to \Sigma_c]})^\top \mb{G}_b^{[\Sigma_c]} Ad(\mb{T}^{[\Sigma_a \to \Sigma_c]}).
\end{equation}

%
%
%
%

\subsection{A Practical Case Study: Robot Joint Actuator Dynamics}

\begin{figure}[t]
    \centering
    \includegraphics[width=0.8\textwidth]{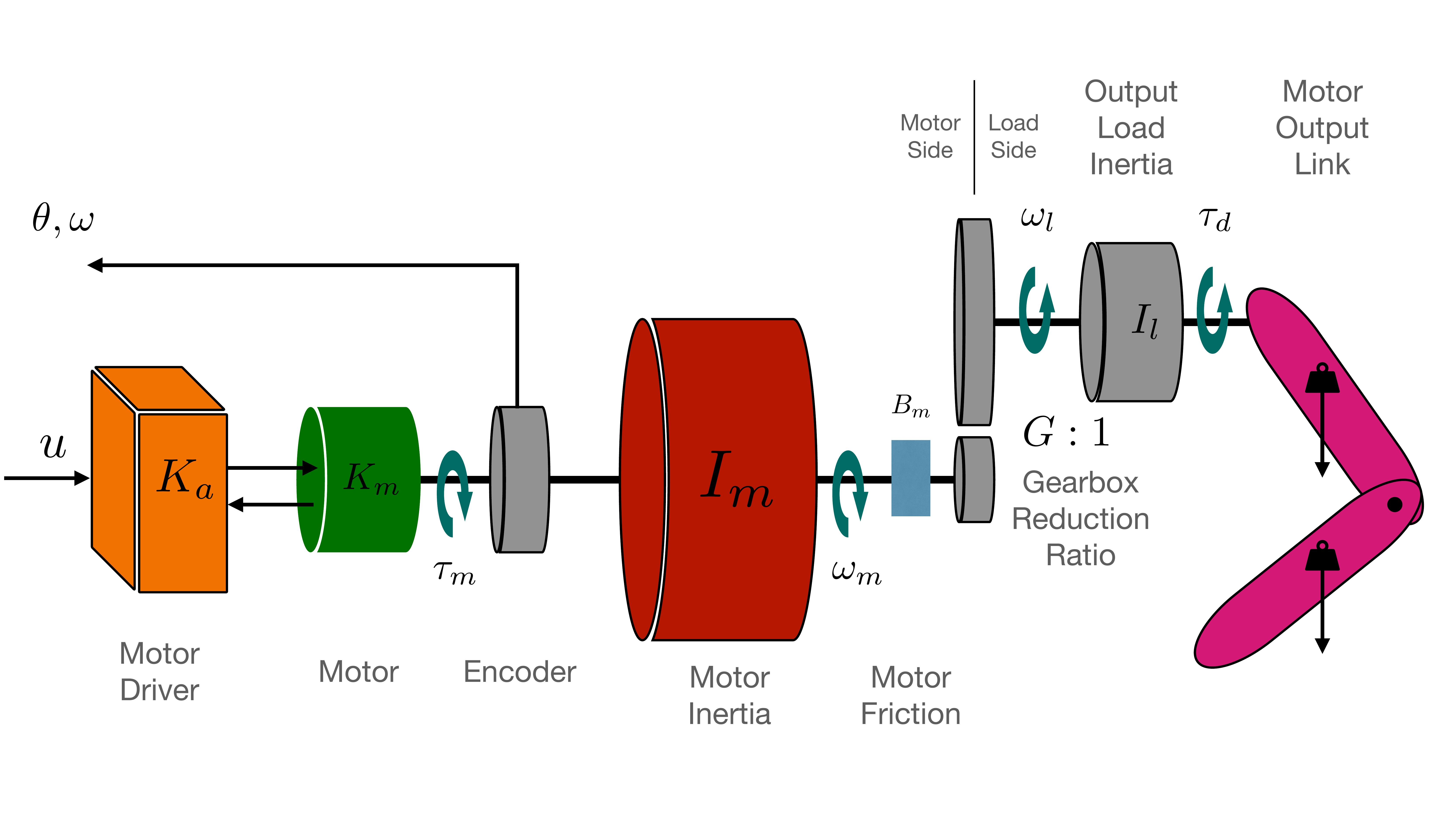}
    \caption{An Example of Robot Arm Joint Components in the DC Motor.}
    \label{fig:robot_arm_joint_components}
\end{figure}

Finally, at the end of this section, we will use a concrete robot arm joint motor to illustrate robot dynamics in practice. As shown in Figure~\ref{fig:robot_arm_joint_components}, we provide the key components of the robot arm joint DC motor, which include \textit{motor driver}, \textit{motor}, \textit{encoder}, \textit{motor inertia}, \textit{motor friction}, \textit{gearbox}, \textit{output load inertia} and \textit{output link}. In the following part, we will introduce these key components one by one for readers. 

\subsubsection{Actuator}

We take the electrical DC motor as an example, whose input is the voltage $u$ from the used power supply. Depending on the power supply and motor used in the robot arm joint, the voltage $u$ value can be different, {\eg}, $5V$, $24V$ or $48V$. Nowadays, the DC motors we use are normally current controlled, {\ie}, the current will control the motor pose, rotational speed and output torque. Via the \textit{motor driver} shown in Figure~\ref{fig:robot_arm_joint_components}, we can calculate the current into the motor as
\begin{equation}
i_m = K_a(u),
\end{equation}
where the mapping $K_a: V \to A$ will calculate the output current corresponding to the input voltage.

Meanwhile, the output torque by the motor is proportional to the current, which can be represented as
\begin{equation}
\bs{\tau}_m = K_m(i_m),
\end{equation}
where the mapping $K_m: A \to Nm$ calculates the output torque for the input current.

\subsubsection{Angular Velocity}

The torque $\bs{\tau}_m$ will drive the motor and accelerate the rotational velocity of the \textit{motor inertia} $\mb{I}_m$ to generate the \textit{angular velocity} $\omega_m$ and the motor current rotation angle $\theta$.

According to the previous Equation~\ref{equ:rigid_body_torque}, we have already illustrated the relationship between the motor output \textit{torque} and its \textit{angular velocity}, which is rewritten for readers below as well:
\begin{equation}
\bs{\tau}_m = \mb{I}_m \dot{\bs{\omega}}_m + \widehat{\bs{\omega}}_m \mb{I}_m \bs{\omega}_m.
\end{equation}

The \textit{angular velocity} $\bs{\tau}_m$ and the current actuator position ({\ie}, the rotated angle) $\theta$ will be outputted to high-level controller via the encoder of the motor.

\subsubsection{Friction}

In the real-world, as the motor actuator rotates, it will be affected by frictions, which can be caused either by the \textit{viscous friction} or by the \textit{Coulomb friction}. The frictions will be related to the \textit{angular velocity} of the motor, and will introduce an offset torque to oppose the motion.

Formally, we can represent the torque created by the \textit{viscous friction} can be represented as
\begin{equation}
\bs{\tau}_v = B (\bs{\omega}).
\end{equation}
Here, mapping $B: rad/s \to Nm$ is also called the \textit{viscous friction} mapping, which is a linear function of the input with coefficient $b$.

Meanwhile, the torque caused due to the \textit{Coulomb friction} depends on the rotation direction and can be represented as
\begin{equation}
\bs{\tau}_c = \begin{cases}
\bs{\tau}_c^+ & \text{, if } \bs{\omega} > 0\\
\mb{0} & \text{, if } \bs{\omega} = 0\\
\bs{\tau}_c^- & \text{, if } \bs{\omega} < 0\\
\end{cases},
\end{equation}
where $\bs{\tau}_c^+$ and $\bs{\tau}_c^-$ are two constants.

In sum, we can represent the sum of torque caused by frictions as
\begin{equation}
\bs{\tau}_f = \bs{\tau}_v + \bs{\tau}_c.
\end{equation}

By taking the frictions into consideration, we can represent the output torque by the \textit{actuator} as
\begin{equation}
\bs{\tau} = \bs{\tau}_m - \bs{\tau}_f.
\end{equation}

\subsubsection{Gearbox}

For the motors used in practice, there usually exist a \textit{gearbox} attached to the motor as the \textit{reducer} to lower down the rotational velocity. The \textit{reducer} is a gear train between the motor and the machinery that is used to reduce the speed with which power is transmitted. It is normally mechanical gadget and its essential use is to duplicate the measure of torque produced by an information power source to expand the measure of usable work.

As shown in Figure~\ref{fig:robot_arm_joint_components}, we assume the \textit{gearbox} reduction ratio to be $G:1$, {\ie}, the motor shaft rotates $G$ rounds, the output shaft will rotate $1$ round. Due to the existence of the \textit{gearbox}, the quantities measured about the motor at the internal motor side and at the output side will be different. 

Let $\bs{\omega}_l$, $\dot{\bs{\omega}}_l$, $\bs{\tau}_l$, $\bs{\tau}_{c,l}$, $b_l$ and $\mb{I}_l$ denote the \textit{angular velocity}, \textit{angular acceleration}, \textit{torque}, \textit{torque due to the Coulomb friction}, \textit{viscous friction torque coefficient} and \textit{inertia matrix} measured at the output load side, respectively. With consideration about the reduction ratio, those quantities measures at the motor side will be very different, which can be denoted as follows:
\begin{align}
\bs{\omega}'_l &= \bs{\omega}_l/G \text{, }\ \ \ \  \dot{\bs{\omega}}'_l = \dot{\bs{\omega}}_l/G,\\
\bs{\tau}'_l &= \bs{\tau}_l G \text{, }\ \ \ \  \bs{\tau}'_{c,l} = \bs{\tau}_{c,l} G,\\
b'_l &= b_l G^2 \text{, }\ \ \ \  \mb{I}'_l = \mb{I}_l G^2.
\end{align}

\subsubsection{Motor Torque Balance}

\begin{figure}[t]
    \centering
    \includegraphics[width=0.9\textwidth]{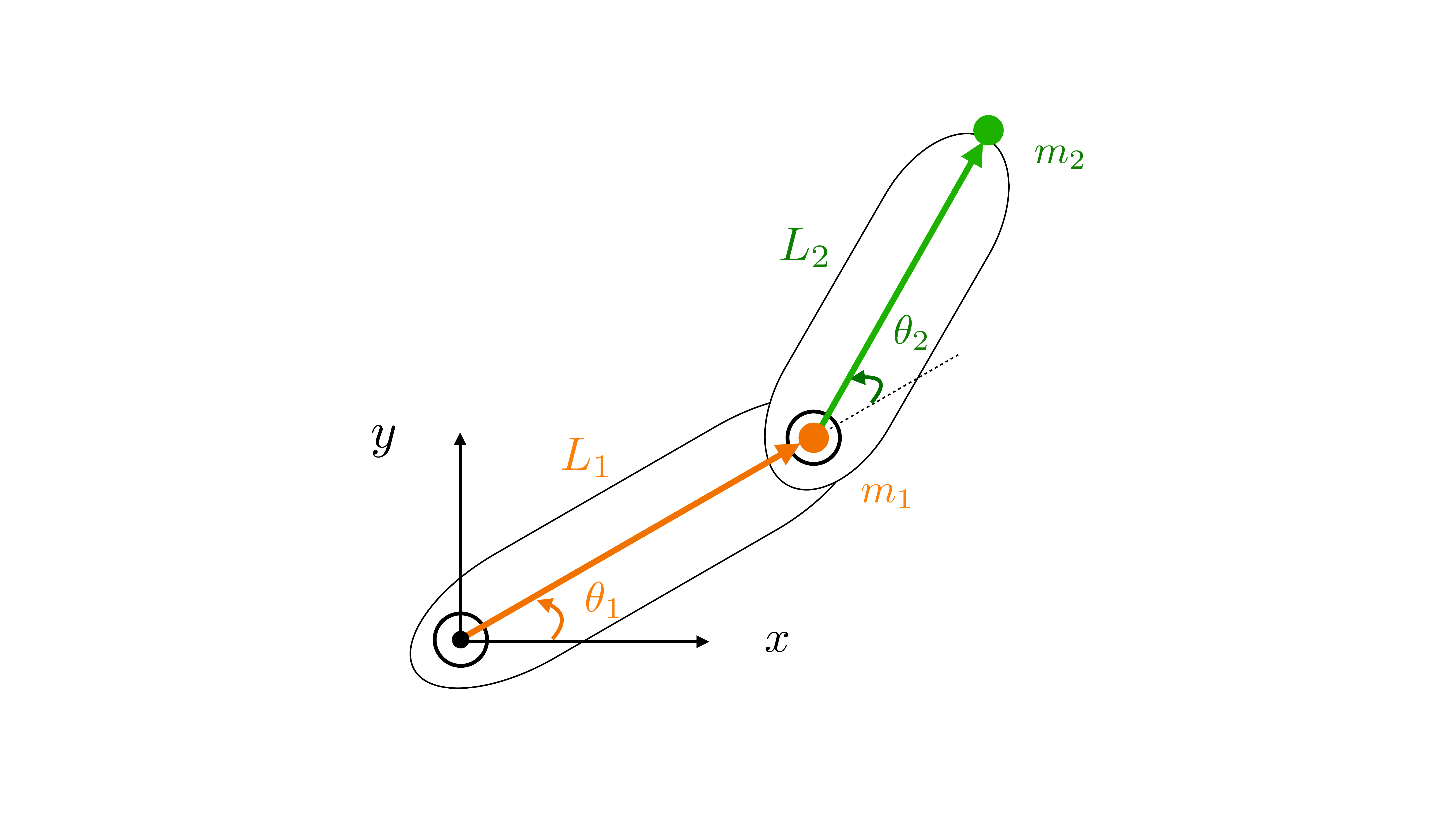}
    \caption{Dynamics of Two-Link Robot Arm.}
    \label{fig:two_link_arm_dynamics2}
\end{figure}

Meanwhile, as to the output link, we have discussed about the dynamics of 2-link robot arm in the previous Section~\ref{subsubsec:2_link_arm_dynamics} already. Furthermore, according to the previous Equation~\ref{equ:2_link_arm_dynamics} we derive before, we can represent the torque at the end point of the 2-link robot arm as
\begin{equation}
\bs{\tau} = \mb{M}(\bs{\theta}) \ddot{\bs{\theta}} + { \mb{c}(\bs{\theta}, \dot{\bs{\theta}}) + \mb{g}(\bs{\theta})},
\end{equation}

Based on the above descriptions, we can summarize all the torques we derive above and get the torque balance on the motor shaft as follows:
\begin{equation}
K_m(K_a(u)) - (B(\bs{\omega}; b_m) + B(\bs{\omega}; b_l/G^2)) - (\bs{\tau}_c + \bs{\tau}_{c,l}/G) - \bs{\tau}_d/G = (\mb{I}_m + \mb{I}_l/G^2)\dot{\bs{\omega}}.
\end{equation}
The dynamics of the motor can be analyzed based on the above torque balance equation. According to the dynamics analysis result, we can further control the robot joint motor to perform the necessary movement according to our task requirements. As to the robot control methods, we will introduce them in the follow-up articles instead.

\section{What's Next?}

By now, we have introduced the \textit{robot motion}, \textit{forward kinematics}, \textit{inverse kinematics}, and \textit{robot dynamics}. In the next articles, we will further talk about several advanced topics about robotics, including \textit{robot control}, \textit{trajectory generation}, \textit{motion planning}, \textit{zero moment point}, \textit{biped walking}, \textit{robot manipulation} and \textit{robot simulation}.

\newpage

\vskip 0.2in
\bibliographystyle{plain}
\bibliography{reference}

\end{document}